\documentclass[a4paper,fleqn]{cas-dc}

\usepackage[dvipsnames]{xcolor}
\usepackage{pifont}
\usepackage{graphicx}
\usepackage{makecell}
\newcommand{\cmark}{\ding{51}}
\newcommand{\xmark}{\ding{55}}
\usepackage[table]{xcolor}
\definecolor{sigGreener}{HTML}{5A7DA6}   
\definecolor{tieColor}{HTML}{EDEAE0}     
\definecolor{sigWorse}{HTML}{B97A66}     

\newcommand{\sigDown}[1]{\cellcolor{sigGreener!55}#1}
\newcommand{\sigTie}[1]{\cellcolor{tieColor}#1}
\newcommand{\sigUp}[1]{\cellcolor{sigWorse!55}#1}

\usepackage[authoryear,longnamesfirst]{natbib}
\usepackage{tabularx}
\usepackage{nicefrac}
\usepackage{float}
\usepackage{svg}
\def\tsc#1{\csdef{#1}{\textsc{\lowercase{#1}}\xspace}}
\tsc{WGM}
\tsc{QE}
\tsc{EP}
\tsc{PMS}
\tsc{BEC}
\tsc{DE}

\usepackage[most]{tcolorbox}
\usepackage{xcolor}
\usepackage{booktabs}
\usepackage{threeparttable}
\usepackage{float}
\usepackage{booktabs}
\usepackage{adjustbox}
\usepackage{rotating} 
\usepackage{subcaption} 

\usepackage{graphicx} 
\usepackage{multirow}
\usepackage{color} 
\usepackage{soul} 
\usepackage{lineno}
\newcommand{\ical}[1]{{\mbox{\usefont{OT1}{pzc}{m}{it}{#1}}}}

\newcommand{\m}[1]{{\mbox{{\fontencoding{T1}\sffamily\slshape{#1\/}}}}}
\renewcommand{\d}[1]{{\mbox{\boldmath$#1$}}}

\begin{document}
\let\WriteBookmarks\relax
\def\floatpagepagefraction{1}
\def\textpagefraction{.001}

\shorttitle{One Trajectory, Many Predictions: Efficient UQ via LoRA Snapshot Ensembles }

\shortauthors{Farag et~al.}

\title [mode = title]{ST-LoRA: Single-Trajectory LoRA Ensemble for Uncertainty-Aware Agricultural Segmentation}                      



%
\author[1]{Mohamed Farag}[type=editor,
                        orcid=0000-0001-7511-2910]

\cormark[1]


\ead{mibrahi2@uni-bonn.de}

\credit{Conceptualization, Methodology, Formal analysis, Ablation-Study ,Software, Validation, Investigation, Writing - original draft,
Writing - review \& editing, Visualization}

\affiliation[1]{organization={Machine Learning in Agriculture Lab, Institute of Geodesy and Geoinformation, University of Bonn},
    city={Bonn},
    country={Germany}}

\affiliation[2]{organization={Lamarr Institute for Machine Learning and Artificial
Intelligence, University of Bonn},
    city={Bonn},
    country={Germany}}

\author[1]{Genc Hoxha}

\credit{Methodology, Investigation, Formal analysis, Visualization, Validation, Writing-review \& editing}

\author[1]{Yahya Maleki}

\credit{Ablation-study, Software, Writing}

\author[3]{Chris McCool}[%
   ]
\credit{Methodology, Investigation, Writing - review \& editing, Visualization}
\author[1,2]{Ribana Roscher}[%
   ]
\cormark[2]

\credit{Methodology, Investigation, Writing - review \& editing, Visualization, Supervision, Project administration,
Funding acquisition}

\affiliation[3]{organization={Commonwealth Scientific and Industrial Research Organisation (CSIRO)},
    country={Australia}}

\cortext[cor1]{Corresponding author}
\cortext[cor2]{Principal corresponding author}



\begin{abstract}
Reliable decision-support in digital agriculture requires not only accurate predictions but also well-calibrated uncertainty estimates, particularly for dense prediction tasks such as semantic segmentation. While ensemble methods provide strong uncertainty quantification, their computational and memory demands limit practical applicability, whereas single-model approximations often compromise uncertainty quality for efficiency. We propose ST-LoRA, a parameter-efficient ensemble framework that constructs diverse ensemble members from a single training trajectory by combining Low-Rank Adaptation (LoRA) with snapshot ensemble approach. Each member shares a frozen pretrained backbone and differs only in its lightweight low-rank adapters, reducing training time compared to deep ensemble, trainable parameters, and storage requirements significantly compared to the full model while preserving ensemble diversity. We conduct experiments across two agricultural datasets -GrowliFlower-L (cauliflower, close-range/open field) and BUP20 (sweet pepper, close-range/glasshouse)- using SegFormer, Mask2Former, and EoMT architectures. Our evaluation spans in-distribution performance, calibration under distribution shift, and near- and far-out-of-distribution detection. Extensive ablations reveal that feed-forward layers are the critical LoRA target for dense prediction, contrary to the attention-only convention established in language models. ST-LoRA emerges as a competitive candidate to full-rank ensembles in segmentation quality, calibration, and out-of-distribution detection, while substantially reducing inference latency, memory footprint, and storage requirements. Compared to the state of the art uncertainty quantification methods, ST-LoRA consistently demonstrates strong performance compared to all approaches on image- and pixel-level out-of-distribution detection, calibration stability under distribution shift, and cross-seed variance, all while requiring substantially fewer parameters and lower computational overhead. These results demonstrate that LoRA-based ensemble adaptation is not merely a coarse approximation to full-rank ensembling, but rather an effective strategy that offers a compelling efficiency/performance trade-off for agricultural vision systems.
\end{abstract}


\begin{highlights}
\item ST-LoRA builds diverse ensembles using less model parameters
\item Feed-forward layers, not attention, are the most important LoRA target for dense prediction
\item ST-LoRA emerges as a competitor to full-rank ensemble at significantly smaller checkpoint size
\item ST-LoRA offers competitive calibration robustness under distribution shift.
\item ST-LoRA delivers competitive performance for near/far OoD detection tasks.
\end{highlights}


\maketitle

\section{Introduction}

Digital agriculture powered by sensor data and Machine Learning (ML) offers promising tools to optimize current practices and explore innovative solutions for future food production. Deep Neural Networks (DNN), in particular, excels at processing large sensor-based datasets to uncover meaningful patterns \cite{Attri2023DeepLearningAgriculture}. This capability has enabled a range of impactful applications, including crop classification \cite{Teixeira2023}, plant growth modeling and forecasting \cite{Drees2022,Drees2024,Quan2019}, disease detection \cite{MobileViTv2PlantDisease2025}, yield estimation \cite{Menon2025CropYieldPrediction}, and building foundation models for enhancing the performance on various downstream tasks by using multi-modal sensory inputs \cite{Attri2025DeepLearningAgricultureAdvances} .

\begin{figure*}[t]
    \centering
    \includegraphics[width=\textwidth]{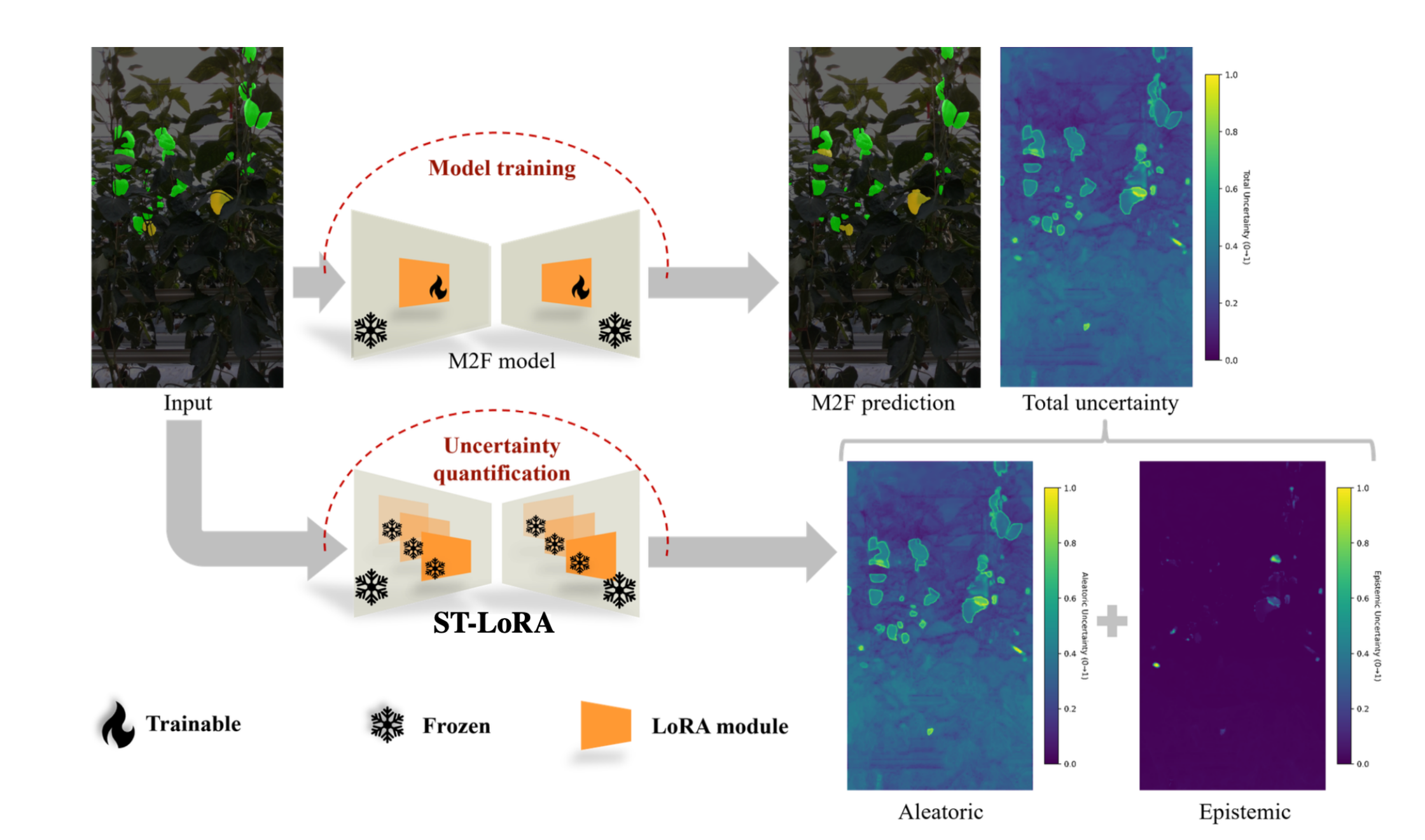}
    \caption{\textbf{ST-LoRA for Efficient Uncertainty Quantification.} The upper part illustrates the fine-tuning process of the semantic segmentation model, where both the input image and ground truth (for simplicity we overlay the ground truth mask over the input image) are provided. Only ST-LoRA modules are fine-tuned, while the rest of the model remains frozen. During inference, an ensemble of ST-LoRA-adapters is applied to the input used to estimate both aleatoric and epistemic uncertainty. The summation denotes the total predictive uncertainty, and yellow/purple represent high/low pixel-wise uncertainty.}
    \label{fig1}
\end{figure*}

As ML models become more integrated into agricultural workflows, enhancing computational efficiency and evaluating the quality of their predictions become essential; not only to establish trust in the predictions, but to know if human oversight is needed and when \textit{abstaining} is safer than a 
confidently wrong decision. Uncertainty quantification (UQ) plays a key role by indicating when predictions are trustworthy unlike point estimates, UQ enables probabilistic predictions that help diagnose issues arising from the data, the model, or both. In supervised learning, models generalize from labelled data to unseen examples, an inductive process that introduces \textit{epistemic}  uncertainty (i.e., model uncertainty) due to limited knowledge about the underlying data-generating process. In contrast, \textit{aleatoric} uncertainty (i.e., data uncertainty) stems from inherent noise and variability in the data. Unlike epistemic uncertainty which can in principle be reduced with more data or improved models, aleatoric uncertainty is irreducible, regardless of model capacity or training set size.~\cite{10.1007/s10462-023-10562-9,Huellermeier2019}.

Among existing UQ methods, Deep Ensembles~\cite{Lakshminarayanan2017} 
are widely regarded as the most commonly used method due to their strong 
calibration, predictive performance, and ability to decompose total 
uncertainty into its epistemic and aleatoric components. However, they are computationally expensive, multiplying both training time 
and memory footprint by the ensemble size, an increasingly critical limitation as modern architectures scale to billions of 
parameters and Vision Transformers~\cite{vaswani2017attention} become the dominant backbone for dense prediction tasks. Snapshot 
ensembling~\cite{huang2017snapshot} addresses the training cost by collecting $M$ checkpoints from a single cosine-annealing trajectory, reducing training time by a factor of $\nicefrac{1}{M}$. However, at deployment time, the snapshot ensembling is affected by large inference latency and memory footprint due to the need of maintaining $M$ full model replicas.

The same scaling pressure that makes deep ensembles prohibitive also 
affects standard fine-tuning. As pretrained vision and language models grow larger, 
adapting them to downstream tasks becomes increasingly costly in both 
compute and storage. Parameter-efficient fine-tuning offers a 
principled solution to this bottleneck. Low-Rank Adaptation 
(LoRA)~\cite{hu2022lora}, originally developed for large language models, 
injects trainable low-rank matrices into the frozen layers of a 
pretrained backbone, dramatically reducing the number of trainable 
parameters and training time. 


This naturally has raised the question of whether LoRA adapters can serve as ensemble members in their own right. Recent work has begun to 
explore this direction for uncertainty quantification~\cite{halbheer2024lora, 
Balabanov2024UncertaintyQI}, freezing the shared backbone and 
constructing diverse members through different low-rank 
initializations. However, existing efforts have been limited to
classification tasks at image level, with narrow ablation studies that leave the effect of key hyperparameters largely 
unexplored. Dense prediction tasks such as semantic segmentation, which impose stricter demands on spatial representations and 
pixel-level calibration, remain entirely unaddressed. This leads to the question at the core of this paper:
\begin{quote}
    \textit{Can uncertainty quantification methods for dense prediction tasks be efficient not only robust ?}
\end{quote}

To fill this gap, we introduce and explore Single Trajectory LoRA (\textit{ST-LoRA}), a lightweight ensemble-based extension of LoRA ensemble method for pixel-wise uncertainty quantification that constructs the members from a shared pretrained vision backbone. Illustrated in Figure~\ref{fig1}, our method emerges as a lightweight competitor to the Full Rank Ensemble (FRE) \footnote{We use this terminology to refer to the snapshot ensemble but by using the full rank instead of lower rank weight matrices.} on calibration and segmentation quality but also with significantly lower computational overhead. 

Our main contributions are:

\begin{tcolorbox}[
  enhanced,                      
  colback=orange!5!white,          
  colframe=orange!75!black,        
  title=\textbf{Contributions},     
  fonttitle=\bfseries,           
  coltitle=black,                
  rounded corners,               
  boxrule=0.4pt,                 
  attach boxed title to top left={yshift=-2mm, xshift=4mm}, 
  boxed title style={
    colback=orange!10!white,
    colframe=orange!75!black,
    boxrule=0.4pt,
    rounded corners
  }
]
   \begin{itemize}
    \item We introduce ST-LoRA, which combines LoRA adapters with 
    snapshot ensembling to reduce training time compared to deep ensemble, number of parameters, and inference latency while being competitive full-rank ensemble's 
    calibration and segmentation quality.

    \item We show that ST-LoRA is less computationally complex by testing it on a real edge hardware used in agricultural context.

    \item We show through extensive ablation that feed-forward layers 
    are important LoRA target in vision transformers for dense 
    prediction, contrary to LLM practice.

    \item We provide a practitioner's guide for configuring ST-LoRA 
    across datasets and architectures, grounded in ablation findings 
    on two crop types and three backbones.

    \item ST-LoRA achieves competitive performance with the low cross-seed variance across 
    in-distribution evaluation, distribution shift, and near/far 
    out-of-distribution detection against Snapshot Ensemble, MC Dropout, and DDU.

\end{itemize}
\end{tcolorbox}

The rest of this article is organized as follows. Section~\ref{sec2} reviews research gaps in uncertainty estimation for agricultural machine learning. Section~\ref{sec3} introduces the mathematical background and presents the proposed ST-LoRA approach. Then, we present in  Section~\ref{sec4} the experimental setup while Section~\ref{sec5} shows the experiment 1 - benchmarking ST-LoRA against FRE in terms of segmentation quality calibration, and computational efficiency- , Section~\ref{sec6} shows an ablation study on the most influential hyperparameters. Followed by Section~\ref{sec7} that presents an assessment of calibration robustness under distribution shift and finally Section~\ref{sec8} shows a comparison with state-of-the art methods covering in-distribution performance, covariate shift, and out-of-distribution (OoD) detection. Section~\ref{fremark} presents the final remarks and Section~\ref{conc} concludes the paper. Additional analysis and implementation details are provided in the Appendix~\ref{appen_a} and Appendix~\ref{appa_b} .

Through extensive evaluation, we demonstrate that ST-LoRA \footnote{Offical code is available at: \url{https://github.com/MohamedFarag21/ST_LoRA}} achieves competitive uncertainty quantification and segmentation quality relative to both full-rank and efficient SOTA uncertainty quantification methods, while offering substantial advantages in computational efficiency, storage footprint, and deployment flexibility.

\section{Related Work}
In this section, We begin with a general overview of uncertainty in machine learning, then follow the 
uncertainty pipeline through its two building blocks — representation, 
and quantification — before reviewing the limitations of 
existing approaches that motivate this work.
\label{sec2}

\noindent\textbf{Uncertainty in ML} can be addressed from two main perspectives. One common approach assumes that predictive uncertainty can be decomposed into \textit{epistemic} (model) and \textit{aleatoric} (data) components, each estimated independently \cite{Shaker2021}. An alternative perspective challenges the validity of this decomposition \cite{Huellermeier2022}, focusing instead on estimating total predictive uncertainty to ensure model reliability \cite{Angelopoulos2021}. As noted by \cite{Gruber2023}, the origin and categorization of uncertainty in ML remain ambiguous, with additional factors such as missing data or deployment in dynamic environments contributing to a more complex uncertainty landscape.

\noindent\textbf{Uncertainty Evaluation}. Evaluating UQ methods is non-trivial, as no ground-truth uncertainty 
labels exist for model predictions. Instead, evaluation relies on 
indirect schemes that probe whether uncertainty estimates are meaningful 
and actionable in practice. Following \cite{Hofman2025UncertaintyQW}, 
four paradigms have been established: \textit{calibration under 
distribution shift} \cite{MORENOTORRES2012521} assesses whether confidence scores remain reliable 
when imaging conditions change — critical in agricultural settings where 
lighting and weather vary; \textit{OoD detection}~\cite{Salehi2021AUS} 
identifies semantically novel inputs, flagging them for human review 
rather than producing overconfident predictions; \textit{selective 
prediction} \cite{xin-etal-2021-art} extends this by allowing abstention on any low-confidence 
sample, preferring a partial prediction over a confident wrong one; and 
\textit{active learning} \cite{bachman2017learning} guides annotation by prioritizing uncertain 
samples, though this is orthogonal to our deployment-oriented focus. 
This work evaluates the first three schemes, as they directly assess 
prediction reliability in real-world agricultural deployment.


\noindent \textbf{Uncertainty Representation} for UQ falls into two primary categories: \textit{stochastic} and \textit{deterministic}. Stochastic representation captures epistemic uncertainty by modeling a second-order distribution over model parameters or outputs \cite{hofman2025uncertainty}. In contrast, deterministic represents uncertainty directly through distributions over latent representations \cite{postels2022practicality}.\footnote{This is not an exhaustive list. Other notable approaches include generative models, credal sets, and Conformal Prediction (CP) \cite{Huellermeier2019}, as well as Laplace Approximation (LA) \cite{laplace1774memoire} and Evidential Deep Learning (EDL) \cite{sensoy2018evidential}, which are increasingly explored in recent literature.}. 

\noindent \textbf{Uncertainty Quantification Methods}. Stochastic methods approximate Bayesian Neural Networks (BNNs), tackling the intractability of posterior inference through techniques like Monte Carlo Dropout (MC-Dropout) \cite{Gal2015DropoutAA}, Ensemble Learning \cite{Lakshminarayanan2017}, and Flipout \cite{Wen2018}. Deterministic methods - sampling free approaches, estimate uncertainty in a single forward pass by treating model weights as fixed. Recent advances in Deterministic Uncertainty Modeling (DUMs) \cite{Mukhoti2023,Amersfoort2020UncertaintyEU,liu2020simple,benkert2024transitional} achieve reliable estimates by regularizing latent representations for smoothness and bi-Lipschitz continuity, while avoiding feature collapse. These approaches offer lower computational overhead however, DUMs can suffer under distribution shifts and may negatively impact predictive accuracy due to the smoothness constraints imposed on the learned representations~\cite{Postels2021}. Hence more efforts focused on enhancing and exploring more efficient ensemble strategies.

\textbf{Efficient Ensembles}. Post-hoc ensemble methods like Checkpoint Ensembles (CE) \cite{chen2017checkpoint} and Snapshot Ensembles (Snap-E) \cite{huang2017snapshot} reduce training cost by saving intermediate model states during a single run. In contrast, integrated ensemble methods, including Packed Ensembles (PE) \cite{laurentpacked}, Multi-Input Multi-Output (MIMO) \cite{havasitraining}, and Batch Ensembles (BE) \cite{wenbatchensemble}, embed ensembling into the architecture itself. For instance, PE uses group convolutions for efficient computation, MIMO extracts sub-networks from a shared backbone, and BE represents weights using a shared matrix modulated by rank-one factors. Finally, MC-Dropout provides a lightweight stochastic alternative by activating dropout at inference time.

 While these methods improve the efficiency of uncertainty estimation, several challenges remain. Post-hoc ensembles reduce training time but do not lower inference cost or memory usage, as each ensemble member must be evaluated independently. Integrated ensemble approaches often require architectural modifications, which can be complex and potentially degrade the base model’s performance; for instance, the group convolutions used in Packed Ensembles (PE) are incompatible with Vision Transformer (ViT) architectures~\cite{dosovitskiyimage}. MC-Dropout tends to be miscalibrated on in-distribution data and overly confident on out-of-distribution inputs, limiting its applicability in safety-critical settings~\cite{Lakshminarayanan2017}. 
 
In the text \cite{mao2025survey} and vision \cite{zhu2024melo,espejogarcia2025foundation} domains, LoRA~\cite{hu2022lora} has been particularly impactful, enabling fine-tuning on downstream tasks at a fraction of the computational cost of full fine-tuning. This efficiency has motivated its extension to ensemble construction: rather than training $M$ independent full models, LoRA adapters can serve as lightweight 
ensemble members over a shared frozen backbone, preserving ensemble 
diversity.

 Closely related to our work, \cite{paul2024parameter} explored adapter-based techniques combined with Bayesian inference for uncertainty estimation in monocular depth estimation, while \cite{onalgaussian} applied a Bayesian framework with LoRA to quantify uncertainty in large language models. Recent efforts to investigate the suitability of LoRA ensembles for uncertainty estimation have been presented in \cite{halbheer2024lora} and \cite{Balabanov2024UncertaintyQI}. However, these studies share several limitations that motivate our work. First, they are restricted to image classification and Natural Language Processing (NLP) benchmarks, leaving open whether LoRA ensemble uncertainty estimation transfers to dense prediction tasks where uncertainty is a spatial field of pixel-wise distributions requiring class-aware calibration metrics, region-level OoD detection — none of which have been evaluated. Second, while \cite{halbheer2024lora} showed that LoRA Ensemble performs well against other uncertainty the investigation of LoRA hyperparameters has been limited which heavily affects the performance. Third, existing work retains M-fold training cost through independent member initialization, treating snapshot ensembling as a weaker alternative rather than a complementary mechanism. While prior layer placement studies compare coarse blocks such as attention versus attention with MLP, fine-grained component-level ablation remains absent, and the optimal configuration in classification settings does not transfer to vision transformer encoders used for segmentation. Finally \cite{FARAG2025110559} have shown that DEs and MC-Dropout are the most commonly used methods for uncertainty estimation in agricultural segmentation, while \cite{espejogarcia2025foundation, 11308188} have shown that LoRA can be used for efficient fine tuning in agriculture without exploring the uncertainty aspect. This reveals a significant gap in exploring alternative methods that offer a better trade-off between computational efficiency and predictive reliability. In conclusion we observe the following gaps:

\begin{tcolorbox}[
  enhanced,
  colback=red!5!white,
  colframe=red!75!black,
  title=\textbf{Research Gaps},
  fonttitle=\bfseries,
  coltitle=black,
  rounded corners,
  boxrule=0.4pt,
  attach boxed title to top left={yshift=-2mm, xshift=4mm},
  boxed title style={
    colback=red!10!white,
    colframe=red!75!black,
    boxrule=0.4pt,
    rounded corners
  }
]
\begin{itemize}
    \item \textbf{LoRA ensembles are untested in dense prediction tasks.} Existing 
    work is confined to image classification and NLP benchmarks, with no 
    evaluation on dense prediction tasks or agricultural datasets.

    \item \textbf{Critical deployment properties are neglected.} No 
    prior LoRA-based UQ method evaluates calibration under distribution 
    shift for dense prediction tasks which is essential for safe 
    real-world deployment.

    \item \textbf{Layer selection is an open question.} All prior work 
    applies LoRA exclusively to attention projections following LLM 
    conventions, without investigating whether feed-forward layers 
    are more effective in vision transformers.

    \item \textbf{Agricultural UQ lacks efficient alternatives.} 
    Uncertainty estimation in agricultural segmentation relies solely 
    on expensive methods~\cite{FARAG2025110559}, with no 
    calibration-aware, resource-efficient solution available.
    
    \item \textbf{LoRA literature in Agriculture overlooks uncertainty quantification.} 
    It has focused only on downstream task without integrating uncertainty.
\end{itemize}
\end{tcolorbox}

\section{Methodology}
\label{sec3}

In this section we begin by establishing 
notation, then briefly review the uncertainty representations used 
in our evaluation, before introducing the ST-LoRA formulation and 
its integration into semantic segmentation architectures.

\subsection*{Problem Formulation: Pixel-Wise Classification}
\label{sec:problem_formulation}

Let $\ical{F}_{\theta}: \mathbb{R}^{H \times W \times 3} \rightarrow 
\mathbb{R}^{H \times W \times L}$ be a pre-trained segmentation model 
mapping an input image $\m{X} \in \mathbb{R}^{H \times W \times 3}$ to 
pixel-wise logits over $L$ classes, where $H$ and $W$ denote image height 
and width. For each pixel $(i,j)$, the predicted class probability vector 
is obtained by applying a softmax:
\begin{equation}
    \ical{P}(\hat{\d y} | \m X)_{ij} = \text{softmax}\bigl(\ical{F}_{\theta}(\m{X})_{ij}
    \bigr) \in \Delta^{L-1},
\end{equation}
where $\Delta^{L-1}$ is the $(L-1)$-dimensional probability simplex. 
The predicted class label is then:
\begin{equation}
    \hat{\d y}_{ij} = \operatorname*{arg\,max}_{l \in \{1,\ldots,L\}} 
    \ical{P}(\hat{\d y} | \m X)_{ij}
\end{equation}
The ground-truth segmentation mask $\m{Y} \in \{1,\ldots,L\}^{H \times W}$ 
assigns a single class label to each pixel. The model is trained by 
minimizing the cross-entropy loss over all pixels:
\begin{equation}
    \ical{L}(\theta) = -\frac{1}{HW}\sum_{i=1}^{H}\sum_{j=1}^{W} 
    \log \ical{P}(\hat{\d y} | \m X)_{ij}\,\d{Y}_{ij}.
\end{equation}
Under this formulation, uncertainty quantification amounts to estimating, 
for each pixel $(i,j)$, how reliable $\hat{\d{y}}_{ij}$ is — 
distinguishing pixels where the model is confidently correct from those 
where the prediction should be treated with caution or abstained from 
entirely.
For clarity, we adopt the following notation throughout this work. Scalars are denoted by lowercase letters 
(e.g., $r$, $\tau$), vectors by bold lowercase letters (e.g., 
$\m{y}$), and matrices by bold uppercase letters (e.g., $\m{X}$). 
Model parameters are denoted by $\theta$, with $\theta_0$ indicating 
frozen pre-trained weights. Probability distribution is represented by $\ical{P}(\cdot)$ and Gaussian Mixture Model (GMM) \cite{477e7e2b-4ded-3369-981e-9b40850a2701} is represented as $\ical{Q}(\cdot)$, and functions or models 
by calligraphic letters (e.g., $\ical{F}$, $\ical{G}$, $\ical{H}$). 
Sets and spaces are denoted by blackboard bold letters 
(e.g., $\mathbb{R}$, $\mathbb{M}$). A summary of the main symbols is provided in Table~\ref{tab:notation}.

\begin{figure*}[t]
    \centering
    \includegraphics[width=0.95\textwidth, trim={1cm 1cm 1cm 1cm}]{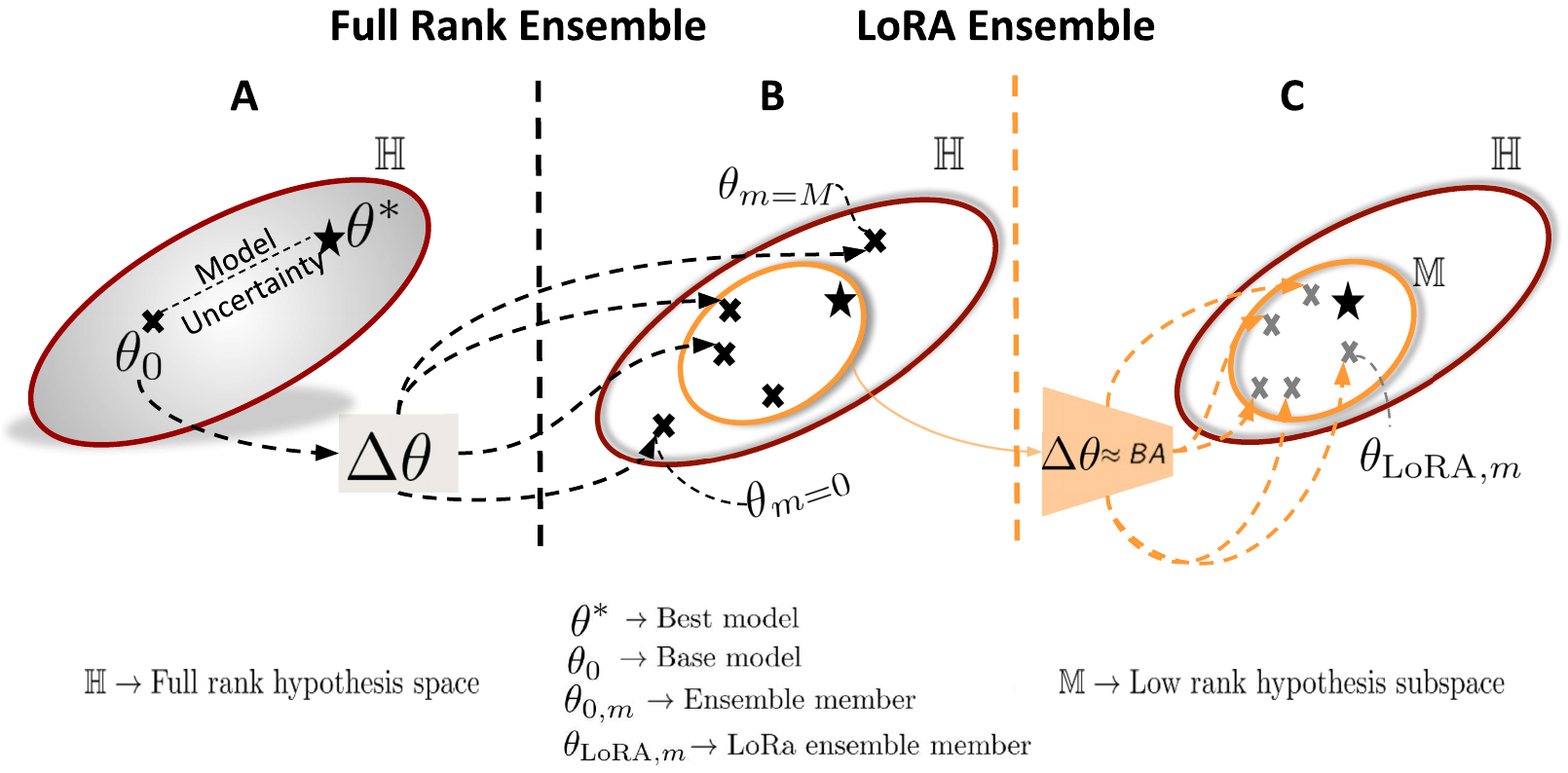}
    \caption{\textbf{LoRA Ensemble Visual Interpretation.} \textit{A)} a DL model once fully trained, the hypothesis space collapses (represented by the gray shade) $\mathbb{H}$ to a single deterministic parameter \(\theta_{0}\) leaving no ability to explore the hypothesis space without retraining. \textit{B)}, the standard ensembling approach is shown, where multiple models are independently updated via full-rank weight shifts \(\Delta \theta\) based on the downstream task. \textit{C)} by sampling or training ensemble members using additive low-rank perturbations, this defines a constrained subspace $\mathbb{M} \subset \mathbb{H}$  within which ensemble members can be efficiently realized. As such, LoRA ensemble allows us to explore the local function space around \(\theta_{0}\) and quantify uncertainty without incurring the cost of full retraining or high-rank diversity. The rank of these perturbations controls the expressive power and the quality of the epistemic uncertainty representation.}
    \label{fig2}
\end{figure*}

\begin{table}[h]
\centering
\caption{\textbf{Key notation used in this work}.}
\label{tab:notation}
\setlength{\tabcolsep}{4pt}
\small
\begin{tabularx}{\columnwidth}{lX}

\textbf{Symbol} & \textbf{Description} \\
\midrule
\multicolumn{2}{l}{\textbf{Data and Labels}} \\
$\m{X}$                               & Input image \\
$\d{y}_l$                             & Label for class $l$ \\
$\m{D}$, $L$, $H$, $W$               & Dataset, classes, height, width \\
\midrule
\multicolumn{2}{l}{\textbf{Model and Parameters}} \\
$\theta_0$                             & Frozen pre-trained weights \\
$\Delta\theta_m$                       & Low-rank update for member $m$ \\
$\m{A}_m$, $\m{B}_m$                  & ST-LoRA decomposition matrices \\
$r$, $\alpha$, $M$                    & Rank, scaling factor, ensemble size \\
$\ical{F}_{\theta}$                    & Segmentation model \\
$\mathbb{M}_r$                        & Low-rank subspace \\
\midrule
\multicolumn{2}{l}{\textbf{Uncertainty}} \\
$\text{TU}$, $\text{EU}$, $\text{AU}$ & Total, epistemic, aleatoric uncertainty \\
$\ical{P}(\m{Y} \mid \m{X})$          & Predictive posterior \\
$\ical{P}(\theta_m \mid \m{D})$        &  Posterior weight for member $m$ \\
$\psi(\m{X})$                &  \\
\midrule
\multicolumn{2}{l}{\textbf{DDU-specific}} \\
$\m{Z}$                               & Latent feature vector \\
$\ical{Q}(\m{Z} \mid l)$              & Class-conditional GMM density \\
$\boldsymbol{\mu}_l$, $\boldsymbol{\Sigma}_l$ & GMM mean and covariance \\
$\Pi(l)$                              & Class prior probability \\
\bottomrule
\end{tabularx}
\end{table}

\subsection{Uncertainty Representation and Quantification}
\label{sec:uncertainty_representation}
In this work, we consider two uncertainty representation frameworks: 1) stochastic representation (e.g. second-Order Distributions) which explicitly model uncertainty over predictions and 2) latent space distributions which represent uncertainty directly through distributions over latent space (i.e., feature space). We briefly describe their mathematical formulation in the following subsections.

\subsubsection{Second-Order Distributions}
Stochastic methods represent epistemic uncertainty by modeling a 
distribution over model parameters. The total predictive uncertainty 
(TU) is decomposed as $\text{TU} \approx \text{AU} + \text{EU}$, 
where AU and EU denote aleatoric and epistemic uncertainty 
respectively. Using Shannon entropy~\cite{hofman2025uncertainty}:
\begin{equation}
\label{eq1}
    \text{TU} \approx -\sum_{l=1}^{L} \ical{P}(\d{y}_l \mid \m{X}_{\text{test}}) 
    \log_2 \ical{P}(\d{y}_l \mid \m{X}_{\text{test}}),
\end{equation}
where $\ical{P}(\d{y}_l \mid \m{X}_{\text{test}})$ is the probability 
of class $l$ under the predictive posterior. The aleatoric component is the 
expected entropy across ensemble members:
\begin{equation}
    \text{AU} \approx -\sum_{m=1}^{M} \ical{P}(\theta_m \mid \m{D})
    \sum_{l=1}^{L} \ical{P}_{\theta_m}(\d{y}_l \mid \m{X}_{\text{test}}) 
    \log_2 \ical{P}_{\theta_m}(\d{y}_l \mid \m{X}_{\text{test}}),
\end{equation}
and epistemic uncertainty follows as $\text{EU} \approx \text{TU} - \text{AU}$.

\subsubsection{Latent Space Distributions}
Deterministic methods such as DDU~\cite{Mukhoti2023} estimate 
uncertainty from the density of the input's latent representation, 
without requiring multiple forward passes. A GMM is fitted per class 
on penultimate-layer features $\m{Z} = \ical{F}_{\theta}(\m{X})$:
\begin{equation}
    \ical{P}(\m{Z}_{\text{test}} \mid l) = 
    \ical{N}(\m{Z};\, \boldsymbol{\mu}_l, \boldsymbol{\Sigma}_l), 
    \qquad \Pi(l) = \frac{n_l}{\sum_{l'} n_{l'}}.
\end{equation}
Epistemic uncertainty is then the negative log-likelihood under the 
class-marginalized mixture:
\begin{equation}
    \text{EU}(\m{X}_{\text{test}}) := 
    -\log \sum_{l=1}^{L} \ical{Q}(\m{Z}_{\text{test}} \mid l)\,\Pi(l),
\end{equation}
where a high value indicates the sample lies far from the training 
manifold. Aleatoric uncertainty is estimated as the categorical 
entropy of the softmax output, consistent with Eq~\ref{eq1}.

\begin{figure*}[t]
    \centering
    \includegraphics[width=0.75\linewidth]{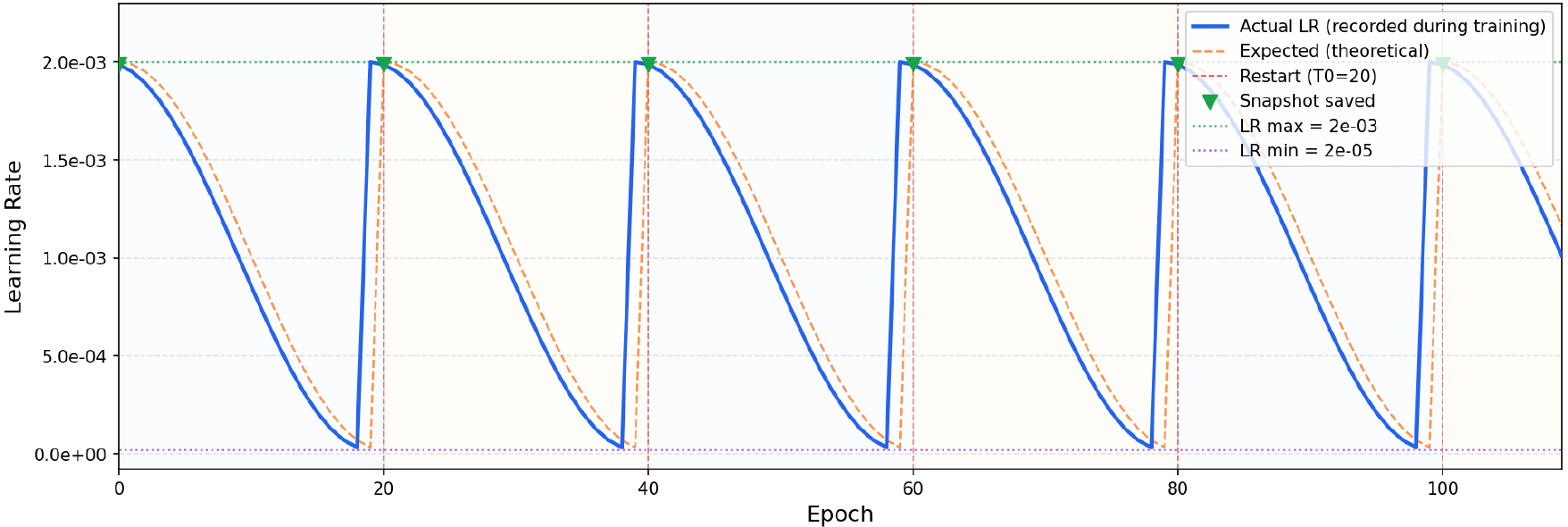}
    \caption{\textbf{Cosine annealing learning rate schedule used for snapshot 
    ensemble construction}. Each cycle drives the model toward a 
    distinct local minimum, and a checkpoint is saved at the end of 
    each cycle (marked by vertical dashed lines). The resulting $M$ 
    snapshots form the ensemble members without requiring $M$ 
    independent training runs.}
    \label{fig4}
\end{figure*}


\subsection{ST-LoRA: Efficient Ensemble Learning via Low-Rank Adaptation}
\label{sec:eST-LoRA}
 We start by injecting LoRA  trainable low-rank matrices into frozen layers. This constrains members' diversity to a learned task-relevant subspace, 
analogous to how Principal Component Analysis (PCA) ~\cite{Shlens2014ATO} identifies low-dimensional 
structure — but unlike PCA, ST-LoRA learns this weight subspace end-to-end 
via gradient descent, aligning it with downstream objectives. For a linear layer with weight 
$\theta_0 \in \mathbb{R}^{d \times k}$, the adapted output is:
\begin{equation}
    \ical{H} = \theta_0\m{X} + \Delta\theta\m{X} 
             = \theta_0\m{X} + \m{B}\m{A}\m{X},
\end{equation}
where $\m{A} \in \mathbb{R}^{r \times k}$ and 
$\m{B} \in \mathbb{R}^{d \times r}$ with $r \ll \min(d,k)$, 
reducing trainable parameters from $dk$ to $r(d+k)$.

This design is theoretically motivated by two observations. First, 
\cite{aghajanyan2021intrinsic} show that pre-trained models have a 
low intrinsic dimension, suggesting low-rank reparameterizations 
can match full-rank fine-tuning. Second, \cite{hu2022lora} demonstrate 
that $\Delta\theta$ is strongly correlated with $\theta_0$, implying 
the low-rank subspace retains sufficient task-relevant information 
for meaningful uncertainty estimation. An illustration is shown in Figure~\ref{fig2}.

Deep Ensembles~\cite{Lakshminarayanan2017, loaiza2025deep} approximate 
Bayesian inference by treating independently trained models as samples 
from the parameter posterior $\ical{P}(\theta \mid \m{D})$. The 
ensemble predictive posterior is:
\begin{equation}
    \ical{P}(\m{Y} \mid \m{X}) = \sum_{m=1}^{M} 
    \ical{P}(\theta_m \mid \m{D})\, 
    \ical{P}_{\theta_m}(\m{Y} \mid \m{X}).
    \label{eq:de_posterior}
\end{equation}
While effective, this requires training and storing $M$ full model 
replicas, making it prohibitive for large vision transformers.

Hence, we use snapshot approach to construct an ensemble from a 
single training trajectory, reducing the effective training cost by a 
factor of $\nicefrac{1}{M}$ compared to training $M$ independent 
models. A cosine annealing learning rate scheduler introduces cyclic 
variation in the learning rate, allowing the model to converge to 
distinct local minima within a single run. At the end of each cycle, 
the current model weights are saved as a snapshot, and the final 
ensemble averages predictions across all $M$ collected 
checkpoints~\cite{huang2017snapshot}. Figure~\ref{fig4} 
illustrates the learning rate schedule and the corresponding snapshot 
collection points.




Each ensemble member $m \in \{1,\ldots,M\}$  in ST-LoRA for Semantic Segmentation
\label{sec:eST-LoRA_seg} is defined by a distinct adapter initialized from a different seed, 
yielding the uniform mixture:
\begin{equation}
    \ical{P}(\m{Y} \mid \m{X}) \approx \frac{1}{M} \sum_{m=1}^{M} 
    \ical{P}_{\theta_0 + \Delta\theta_m}(\m{Y} \mid \m{X}),
\end{equation}
from which TU, AU, and EU are computed pixel-wise using 
Equations~(4) and (5).

\begin{figure*}[t]
    \centering
    \includegraphics[width=1.0\linewidth]{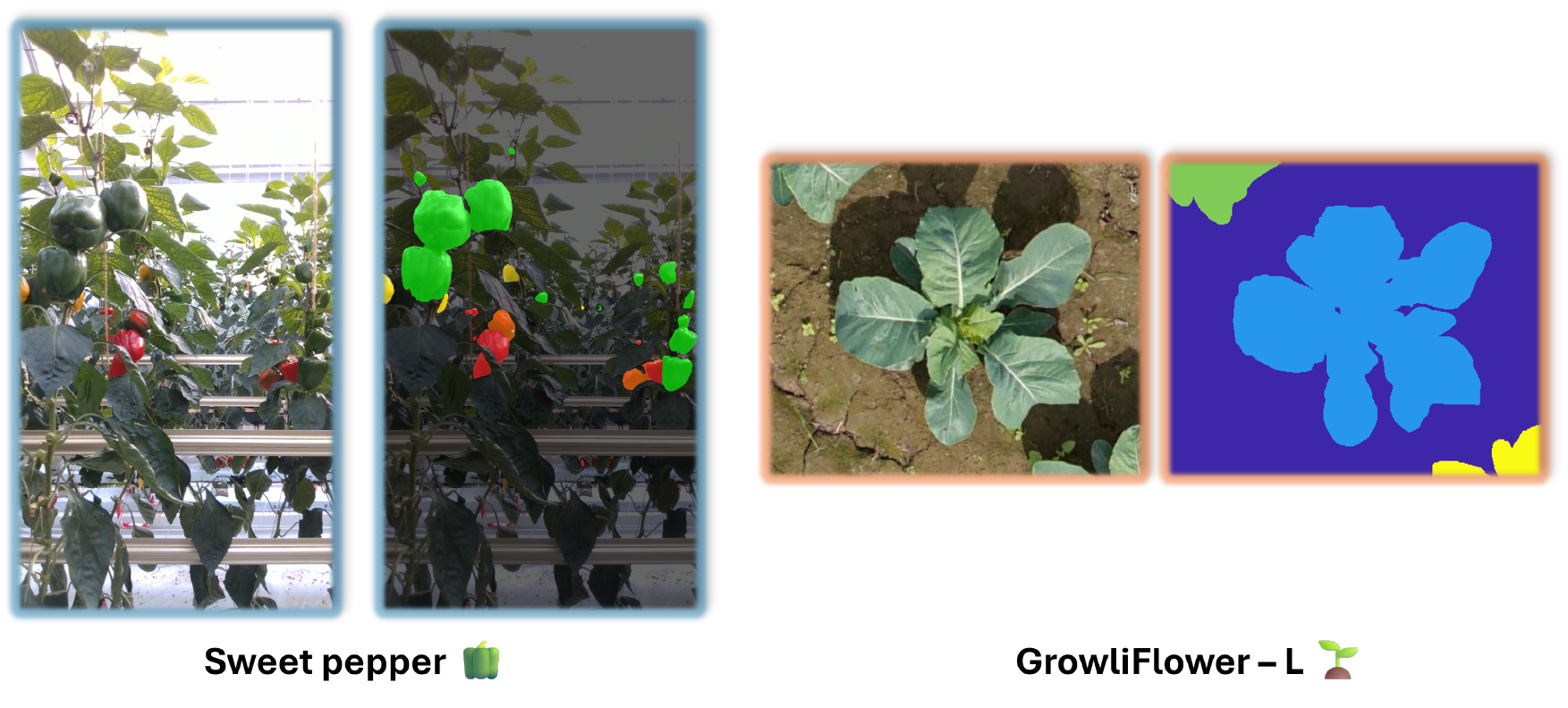}
    \caption{\textbf{Dataset samples.} \textbf{Left:} Sweet pepper (UGV) with semantic segmentation masks for different pepper classes. \textbf{Right:} Cauliflower (UAV) with instance masks; we convert these to binary semantic masks (plant vs. soil) for our experiments.}
    \label{dataset}
\end{figure*}

\section{Experimental Setup}
\label{sec4}

\subsection{Datasets and Pre-Processing}
\textbf{BUP20} \footnote{\url{https://agrobotics.uni-bonn.de/sweet_pepper_dataset/index.html}} sweet pepper phenotyping dataset ~\cite{smitt2021pathobot}, which contains high-resolution (\(1280 \times 720\)) video frames captured by a Unmanned Ground Vehicle (UGV) in a commercial greenhouse. A subset of 279 non-overlapping frames is manually annotated for both semantic and instance segmentation across six classes: background, red, yellow, green, mixed-red, and mixed-yellow peppers. The dataset poses real-world challenges, including occlusions, lighting variation, background clutter, and class overlap, making it a strong benchmark for uncertainty-aware models in agriculture. All images are normalized using the mean $[0.485, 0.456, 0.406]$ and the standard deviation $[0.229, 0.224, 0.225]$, consistent with the ADE20K dataset \cite{zhou2019semantic}.


\textbf{GrowliFlower} \footnote{\url{https://phenoroam.phenorob.de/geonetwork/srv/eng/catalog.search\#/metadata/cb328232-31f5-4b84-a929-8e1ee551d66a}} \cite{Kierdorf2023} is an open-source dataset acquired in 2020 and 2021 by flying a UAV, It contains 14K samples of cauliflower plants with multiple modalities available, including RGB and multi-spectral images. The dataset also includes a comprehensive range of annotations, making it useful for computer vision tasks such as classification and segmentation. All images are resized to $256 \times 256$ \footnote{For images/masks we use bi-linear/nearest interpolations during resizing.} and normalized to be in range $[0,1]$. No augmentation has been used while it helps the ensembles to get more diverse, however we isolate its effect and investigate more its effect at Section~\ref{sec7}. Furthermore, we filter out samples that include only soil at the early growth stages. Samples from both datasets are shown in Figure~\ref{dataset}.



\subsection{LoRA Adaptation for Segmentation Models}
In this work, we focus on Vision Transformer (ViT)--based segmentation models for two reasons. First, our aim is to study the effect of low-rank adaptation in a setting that reflects current practice, and recent advances in semantic segmentation have shifted heavily toward transformer-based architectures, such as SegFormer, DPT, MaskFormer, Mask2Former, EoMT, and SAM~\cite{xie2021segformer,Cheng2021MaskedattentionMT,cheng2021per,ranftl2021vision,kirillov2023segment,kerssies2025your}. Evaluating LoRA in this context therefore aligns our analysis with the predominant direction of modern segmentation research. Second, although LoRA can in principle be applied to convolutional networks, its formulation and most empirical validations have been developed around transformer blocks. 

We adopt SegFormer, Mask2Former and EoMT as our primary models due to their distinct design philosophies. SegFormer follows a per-pixel classification paradigm, combining a ViT-based encoder with a lightweight MLP decoder. In contrast, Mask2Former/EoMT frame segmentation as a mask-classification problem, using a shared transformer-based\footnote{A convolutional encoder can also be used.} encoder and without a decoder or with two decoders: a convolutional decoder for pixel-wise upsampling and a transformer decoder for predicting mask embeddings. Lastly, SegFormer is a task-specific (non-universal) segmentor tailored primarily for semantic segmentation, whereas Mask2Former/EoMT are universal segmentors capable of handling semantic, instance, and panoptic segmentation within a unified framework.

To enable efficient and scalable uncertainty quantification, we integrate LoRA into selected modules of both architectures. We first choose the target components (e.g., MLP layers or Multi-Head Attention blocks) and specify a low rank (e.g.,~ r=8). The remaining model parameters can either remain frozen or be partially fine-tuned alongside the LoRA adapters, depending on the configuration. During inference, only the LoRA adapter and partly fine tuned weights are loaded, enabling faster execution and reduced memory usage. This design supports the construction of lightweight ensembles with significantly lower computational cost, making the approach well suited for resource-constrained agricultural robotics.


\subsection{Evaluation Schemes and Metrics}
We organize our evaluation metrics into three categories aligned with the evaluation schemes described above: predictive performance, computational efficiency, and uncertainty quantification.

\subsubsection{Predictive Performance and Computational Efficiency}
We assess segmentation quality using mean Intersection over Union (mIoU), which measures the average overlap between predicted and ground-truth segments across all classes. We evaluate computational cost across multiple dimensions: the number of trainable parameters; deployment I/O overhead, measured as checkpoint loading latency for each ensemble strategy; training duration, tracked via wall-clock time and seconds per epoch; and energy consumption, measured as average GPU power, GPU energy, and total energy (GPU, CPU, and RAM) on a single \textit{NVIDIA A100-SXM4-80GB} GPU (400\,W maximum thermal power), from which we derive CO$_2$ emissions.\footnote{Energy and emissions are monitored using \url{https://github.com/mlco2/codecarbon}. Wall-clock time is reported in minutes, average GPU power in Watts (W), energy in Watt-hours (Wh), and emissions are obtained by multiplying kWh by 
the German electricity grid carbon intensity for 2022 (380.95\,g\,CO$_2$/kWh).} Together, these metrics provide a comprehensive picture of deployment feasibility on resource-constrained agricultural platforms.

\subsubsection{Uncertainty Method Quality Assessment}

\textbf{Robustness under Distribution Shift}. 
Dataset shift occurs when the joint distribution of training and test data differ, formally expressed as $\ical P_{\text{train}}(\m X, \d y) \neq \ical P_{\text{test}}(\m X, \d y)$ \cite{MORENOTORRES2012521}. Multiple shift types exist, including prior probability shift, concept shift, and covariate shift. This work focuses on \textit{covariate shift}, defined as:
\begin{equation}
   \ical P_{\text{train}}(\d y|\m X) = \ical P_{\text{test}}(\d y|\m X) \quad \text{but} \quad \ical P_{\text{train}}(\m X) \neq \ical P_{\text{test}}(\m X),
\label{eq:covariate_shift}
\end{equation}
where the conditional relationship between inputs $\m X$ and labels $\d y$ remains invariant, but the marginal input distribution shifts between training and test conditions. This scenario is particularly relevant for agricultural applications, where imaging conditions vary across seasons, weather, and lighting.

To evaluate calibration robustness, we adopt the concept of \textit{breakdown point} from robust statistics \cite{robst}. The breakdown point quantifies the maximum proportion of corrupted observations an estimator can tolerate before producing arbitrarily incorrect results. In our context, we assess how calibration quality (measured by ECE) \cite{guo2017calibration} computed with both $\ell_1$ and $\ell_\infty$ norms to capture average and worst-case miscalibration respectively and segmentation performance (measured by mIoU) degrade as the severity of distribution shift increases. A robust UQ method should maintain relatively stable calibration and performance across varying shift intensities.

\textbf{Out of Distribution (OoD) detection}
  An essential method for evaluating the quality of an uncertainty quantification technique is to observe its response to these OoD samples. This response reflects the model's ignorance expressed as high uncertainty about these points.

The problem of OoD detection can be formulated as follows:
\begin{equation}
    \begin{cases} 
    \text{ID}, & \text{if } \psi(\m X_{n+1}) \geq \tau \\
    \text{OoD}, & \text{if } \psi(\m X_{n+1}) < \tau
\end{cases}
\end{equation}

When $\m X_{n+1} \text{ is declared }$, the OoD detector assigns a score to the sample by using the $\psi$ which is the scoring function, if it is greater than or equals $\tau$ it is considered ID, otherwise it is rejected and classified as OoD. For scoring we can use confidence, entropy, or any function under the condition that it is able to differentiate between the classes or to rank them differently. Hence, OoD detection can be treated as a binary classification problem, the only difference is that we lack the reference values for the OoD samples. For out-of-distribution detection, we measure discrimination performance using the area under the receiver operating characteristic curve (AUCROC), which quantifies the method's ability to separate in-distribution from out-of-distribution samples across all possible confidence thresholds.

\subsection{Design rationale}
 The experimental stages are deliberately ordered so that each is motivated by the findings of the  previous one, building toward a complete picture of ST-LoRA's deployment readiness.

We begin with \textit{in-distribution performance} 
(Section~\ref{sec5}), addressing the most fundamental question: can 
low-rank adapters substitute full model replicas without meaningful loss in accuracy or calibration? Establishing this is a prerequisite 
for everything that follows; if ST-LoRA fails here, its efficiency advantages are irrelevant.

Having confirmed in-distribution reliability, we turn to the \textit{ablation study} (Section~\ref{sec6}), which asks how ST-LoRA 
should be configured. Since prior work applies LoRA exclusively to attention projections~\cite{hu2022lora,muhlematter2024lora}, we systematically investigate target components, rank, scaling factor, and dropout to explore the hyperparameter landscape for vision 
transformer segmentation. All subsequent experiments use the configuration identified here.

With a validated and well-configured model, we assess its \textit{robustness under distribution shift }(Section~\ref{sec7}). 
Agricultural imaging conditions vary across seasons, weather, and 
lighting - a model that is well calibrated in-distribution but 
degrades rapidly under shift offers limited practical value. This 
stage examines whether ST-LoRA's calibration advantage is stable or 
fragile, and identifies which hyperparameter choices offers better robustness.

Finally, \textit{comparison against efficient baselines} 
(Section~\ref{sec8}) situates ST-LoRA within the broader landscape 
of uncertainty quantification and post-hoc calibration methods, asking whether it matches or outperforms established SOTA alternatives - Snapshot Ensemble, MC 
Dropout, and DDU \footnote{To our knowledge this is the first time to introduce snapshot ensemble and DDU in digital agriculture context.} - across In distribution performance, robustness under distribution shift , and out-of-distribution detection. Together, these four stages 
demonstrate that ST-LoRA's uncertainty estimates are not only accurate 
in controlled conditions but practically reliable under the 
challenging and variable conditions of real agricultural deployments.

\section{Experiment 1: In Distribution (ID) Performance and Computational Efficiency}
\label{sec5}

We evaluate four model architectures (SegFormer-B2, SegFormer-B4, and Mask2Former Swin-Base \footnote{On BUP20, we target the query, value, key, and MLP for both models. For GrowliFlower-L we target the query, value, and key only.}, and EoMT) with three uncertainty quantification approaches (Softmax as a baseline), Full-Rank Ensemble (referred to in literature as SnapShot ensemble , and ST-LoRA)\footnote{Training settings in details are shown in Table~\ref{tblsec5.1}.}. Predictive performance is assessed via mIoU, calibration via ECE, and computational efficiency via FLOPs and parameter count. Additionally, we quantify the deployment I/O overhead of each ensemble strategy by measuring checkpoint loading latency on both GPU available by Google Colab and edge hardware (Raspberry Pi 4B) \footnote{NVIDIA T4 GPU has been used on the google colab, and Raspberry Pi 4B (Cortex-A72, 1.8 GHz, 8 GB RAM) as a representative of mid-range edge deployment hardware}, comparing per-switch disk reads against in-memory adapter switching. , and finally we measure the energy dissipated and CO$_2$ emissions.

\begin{figure*}[t]
    \centering
    \includegraphics[width=0.85\linewidth]{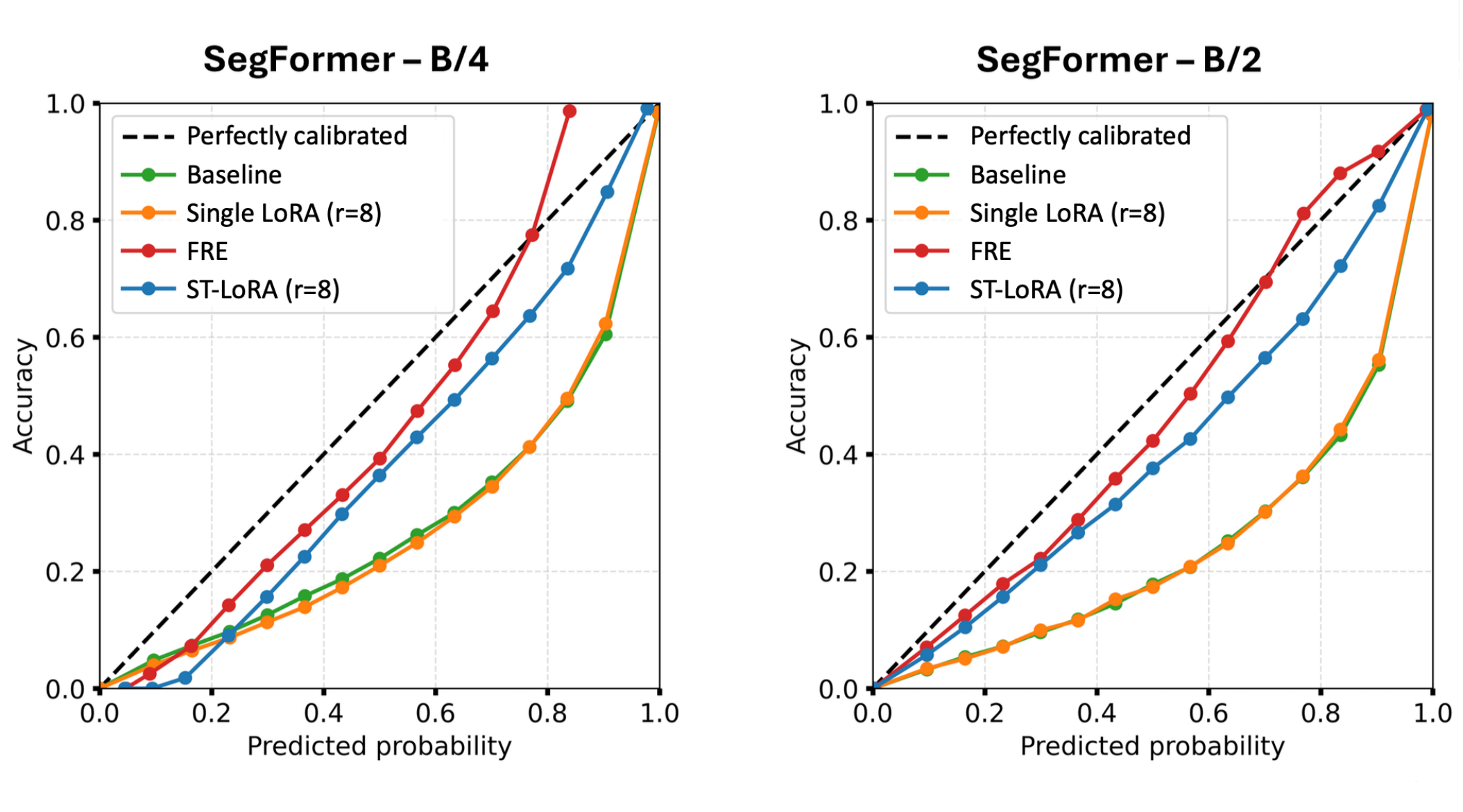}
    \caption{\textbf{GrowliFlower-L reliability diagram for ID calibration assessment.} As shown in the figure, we compare four different settings for both model (e.g., FRE, ST-LoRA, single LoRA adapter, and Baseline model). The ensemble of ST-LoRA enhances calibration by uplifting the plot upwards which reduces the gap between confidence and empirical accuracy.}
    \label{fig5}
\end{figure*}

\subsection{Segmentation Quality and Calibration}
\label{sec5.1}

\begin{table*}[t]
\centering
\caption{\textbf{Segmentation accuracy and calibration across datasets, backbones, and ensemble methods.} FRE refers to a snapshot ensemble built on top of a fully fine-tuned model, $\uparrow$: higher is better; $\downarrow$: lower is better, best result per metric is in \textbf{bold}. $\Delta$ is the relative change over the baseline; \textcolor{ForestGreen}{green}: improvement, \textcolor{Plum}{purple}: degradation.}
\label{in_dist_results}
\setlength{\tabcolsep}{1pt}
{\footnotesize{\scriptsize}
\adjustbox{max width=\textwidth}{%
\begin{tabular}{l|cc|cccc|>{\columncolor{gray!8}}c>{\columncolor{gray!8}}c>{\columncolor{gray!8}}c>{\columncolor{gray!8}}c}

\multirow{2}{*}{\textbf{Model}} 
& \multicolumn{2}{c|}{\textbf{Baseline}} 
& \multicolumn{4}{c|}{\textbf{FRE}} 
& \multicolumn{4}{c}{\textbf{ST-LoRA} (ours)}\\
\cmidrule(lr){2-3} \cmidrule(lr){4-7} \cmidrule(lr){8-11}
 & mIoU(\%) $\uparrow$ & ECE $\downarrow$ 
 & mIoU(\%) $\uparrow$ & $\Delta$mIoU(\%) & ECE $\downarrow$ & $\Delta$ECE(\%)
 & mIoU(\%) $\uparrow$ & $\Delta$mIoU(\%) & ECE $\downarrow$ & $\Delta$ECE(\%) \\
\midrule
\multicolumn{11}{l}{\textit{GrowliFlower-L (Cauliflower)}} \\
\midrule
SgF-B4 & \textbf{88.6} & 0.0052 
       & 83.9 & \textcolor{Plum}{$-$5.3} & 0.0039 & \textcolor{ForestGreen}{$-$25.0}
       & 84.5 & \textcolor{Plum}{$-$4.6} & \textbf{0.0027} & \textcolor{ForestGreen}{$-$48.1} \\
SgF-B2 & 87.8 & 0.0061 
       & 83.9 & \textcolor{Plum}{$-$4.4} & \textbf{0.0034} & \textcolor{ForestGreen}{$-$44.3}
       & 84.7 & \textcolor{Plum}{$-$3.5} & 0.0052 & \textcolor{ForestGreen}{$-$14.8} \\
\midrule
\multicolumn{11}{l}{\textit{BUP20 (Sweet Pepper)}} \\
\midrule
M2F    & 67.8 & 0.0112 
       & 67.1 & \textcolor{Plum}{$-$1.0}  & \textbf{0.0030} & \textcolor{ForestGreen}{$-$73.2}
       & \textbf{68.1} & \textcolor{ForestGreen}{$+$0.4} & 0.0043 & \textcolor{ForestGreen}{$-$61.6} \\
\midrule
SgF-B2 & 66.1 & 0.0028 
       & 70.0 & \textcolor{ForestGreen}{$+$6.0} & 0.0048 & \textcolor{Plum}{$+$71.4}
       & 69.3 & \textcolor{ForestGreen}{$+$5} & 0.0021 & \textcolor{ForestGreen}{$+$25} \\
\midrule
EoMT & 67.2 & 0.0073 
       & 67.4 & \textcolor{ForestGreen}{$+$0.30} & 0.0033 & \textcolor{ForestGreen}{$-$54.8}
       & \textbf{67.5} & \textcolor{ForestGreen}{$+$45} & 0.0032 & \textcolor{ForestGreen}{$-$56.2} \\       
\bottomrule
\end{tabular}%
}
}
\end{table*}


Table~\ref{in_dist_results} reveals a consistent pattern across both datasets: 
ensembling reduces mIoU on GrowliFlower-L but improves it on BUP20 dataset while improving calibration on both reflected by lower ECE compared to single-model counterparts. This trade-off is expected, as ensemble averaging smooths overconfident predictions and can be seen for all models in our comparative analysis.

Comparing ST-LoRA directly to FRE, ST-LoRA achieves competitive performance with less trainable parameters across both datasets and all architectures. On BUP20, 
ST-LoRA matches or exceeds the FRE mIoU among all methods with similar ECE performance on average across models and datasets. On GrowliFlower-L with SegFormer-B4, ST-LoRA achieves the best ECE while also outperforming FRE in segmentation 
accuracy, demonstrating that low-rank adapters can effectively substitute full model replicas without significant performance degradation.

Furthermore, we use reliability diagrams \cite{Dimitriadis2021StableReliabilityDiagrams} to assess model calibration by plotting prediction confidence on the x-axis against empirical accuracy on the y-axis. A perfectly calibrated model follows the diagonal line $y=x$, where accuracy equals confidence at all prediction levels. Deviations below the diagonal indicate overconfidence (predictions are more confident than accurate), while deviations above indicate underconfidence. Figure~\ref{fig5} reveals the substantial impact of ensembling on calibration. Both the baseline and single ST-LoRA models exhibit severe overconfidence, with their reliability curves falling significantly below the diagonal across most confidence bins. Ensembling significantly improves calibration for both approaches: FRE and ST-LoRA curves shift closer to the ideal diagonal, reducing the gap between predicted confidence and actual accuracy. FRE demonstrates marginally better calibration than ST-LoRA across most confidence bins, particularly in the 0.4--0.8 range. However, FRE exhibit underconfidence in the highest confidence bin (0.8--1.0), where the curves rise above the diagonal. This suggests that when the ensembles are most correct, they are actually too conservative in their predictions. This underconfidence at high confidence levels represents a different calibration failure mode that may require targeted post-processing techniques as mentioned in \cite{wang2023calibrating}..

\subsection{Computational Efficiency, Energy Consumption, and $\text{CO}_2$ Emissions}
\label{sec5.2}

Beyond predictive quality, ST-LoRA is substantially more computationally 
efficient than FRE in terms of number of trainable parameters, memory footprint and I/O latency as shown in Table~\ref{tbl:efficiency} and Table~\ref{tab:io_bottleneck}. However, we observe ST-LoRA increases FLOPs more compared to FRE which is a cost shared by both methods due to the ensemble averaging step\footnote{LoRA adds more FLOPS as it introduces more mathematical operations dependent on the chosen rank. Hence for four members we get (4 $\times$ [base model weights + LoRA newly added matrices]).}. 

\begin{table}[t]
\centering
\caption{\textbf{Ensemble I/O latency (ms/model).}
Forward pass excluded.
Best in \textbf{bold}.}
\label{tab:io_bottleneck}
\setlength{\tabcolsep}{4pt}
\small
\begin{tabular}{ll|r>{\columncolor{gray!15}}r}

\textbf{Device}& \textbf{Strategy} & \textbf{FRE} & \textbf{ST-LoRA} (ours) \\
\hline
\multirow{4}{*}{Colab}
  & Sequential (ms)        & 1704.7          & \textbf{172.2}  \\
\cmidrule{2-4}
  & Setup$^\dagger$ (ms)   & 6402.0          & \textbf{2237.4} \\
  & Per switch (ms)        & \textbf{0.001}  & 8.0             \\
\hline
\multirow{4}{*}{Rasp. Pi}
  & Sequential (ms)        & 1017.0          & \textbf{8.0}    \\
\cmidrule{2-4}
  & Setup$^\dagger$ (ms)   & 4080.0          & \textbf{25.0}   \\
  & Per switch (ms)        & \textbf{0.001}  & \textbf{0.001}  \\
\hline
\end{tabular}
\begin{tablenotes}
\small
\item $\dagger$ One-time preloading cost
\item Sizes: FRE $4{\times}431.8$\,MB,
      ST-LoRA $4{\times}5.5$\,MB ($78{\times}$ smaller).
\end{tablenotes}
\end{table}

\begin{table}[t]
\caption{\textbf{Computational efficiency comparison for forward propagation.}}
\label{tbl:efficiency}
\centering
\adjustbox{max width=\textwidth}{%
\begin{tabular}{l|cc}

\textbf{Approach} & \textbf{FLOPs} (GMAC) & \textbf{Train Params.} (M) \\
\midrule
Sgf-B4 (base)  & 12.54  & 61.37   \\
FRE            & 62.70  & 306.84  \\
\rowcolor{gray!15}
ST-LoRA (ours)         & 64.78  & 70.84   \\
\midrule
Sgf-B2 (base)  & 5.34   & 24.72   \\
FRE            & 26.64  & 123.61  \\
\rowcolor{gray!15}
ST-LoRA (ours)         & 27.64  & 29.85   \\
\midrule
M2F (base)     & 229.72 & 108.77  \\
FRE         & 1148   & 534     \\
\rowcolor{gray!15}
ST-LoRA (ours)        & 1170   & 116     \\

\bottomrule

\end{tabular}%
}
\end{table}

The latency reduction stems from the size asymmetry between LoRA adapters and full model replicas: since each adapter is significantly smaller than a full checkpoint, the I/O overhead that typically bottlenecks sequential model loading becomes negligible. To verify this, we benchmark two loading strategies for Mask2Former checkpoints under both FRE and ST-LoRA configurations, on Google colab (network-mounted filesystem) and an edge device (Raspberry Pi 4B, local ext4, 449 MB/s). The first strategy loads each checkpoint from disk before every forward pass, incurring full I/O cost per ensemble member. The second strategy preloads all checkpoints into RAM once at startup, then switches between them in memory during inference, eliminating repeated disk reads entirely.  

Table~\ref{tab:io_bottleneck} reports ensemble switching latency under 
sequential and preloaded strategies on server and edge hardware. 
In the sequential setting, ST-LoRA reduces per-model loading time by 
$10\times$ on the server and $127\times$ on the edge device, a direct 
consequence of its $78\times$ smaller checkpoint size. Under the preloaded 
strategy, ST-LoRA's one-time setup cost is $2.9\times$ lower on the server 
and $163\times$ lower on the edge, reducing startup overhead from 4080\,ms 
to just 25\,ms on the Raspberry Pi~4B. Once preloaded, per-switch latency 
is negligible on both methods and both devices with an advantage for ST-LoRA as the memory is much smaller. 

We have shown that the efficiency of our approach is reflected in reducing training time compared to deep ensemble, fewer trainable parameters, reduced storage requirements ,and lower I/O latency compared to the FRE method. However, all of these metrics have been measured after training. To assess the training-time energy implications of ST-LoRA, we train FRE and ST-LoRA for 40 epochs and monitor their energy footprint. To isolate the training loop from confounding factors, we exclude 
the first epoch - corresponding to GPU warm-up - and omit the validation loop from monitoring.

\begin{table*}[t]
\centering
\caption{\textbf{Statistical comparison of energy consumption and 
emissions, FRE vs.\ ST-LoRA.} Paired, $n{=}5$ seeds 
(42, 123, 456, 789, 1337). $\%\Delta = 100 \cdot 
(\text{ST-LoRA}-\text{FRE})/\text{FRE}$ (negative $\Rightarrow$ 
ST-LoRA lower). Lower is better for every metric. Significance: 
paired $t$-test $p{<}0.05$ \textbf{and} 95\% bootstrap CI excludes 
0 (Wilcoxon two-sided $p$ floors at $0.0625$ at $n{=}5$). Colour 
scale: 
\colorbox{sigGreener!55}{ST-LoRA significantly greener}\,; 
\colorbox{tieColor}{no significant difference}\,; 
\colorbox{sigWorse!55}{ST-LoRA significantly worse}.}
\label{tab:energy_stats_full}
\resizebox{\textwidth}{!}{
\begin{tabular}{llccccccc}
\toprule
\textbf{Arch} & \textbf{Metric} & \textbf{FRE} & \textbf{ST-LoRA} & \textbf{\%$\Delta$} & \textbf{95\% CI} & \textbf{$t$ $p$} & \textbf{Wilcoxon $p$} & \textbf{$d_z$} \\
\midrule
\multicolumn{9}{c}{\cellcolor{gray!8}\textit{Headline --- whole training run (incl.\ validation + warmup)}} \\
\midrule
Mask2Former & CO$_2$ (kg)              & 0.08749\,$\pm$\,0.00049 & \sigDown{0.08193\,$\pm$\,0.0026} & $-6.3\%$  & $[-8.0,-3.5]$   & 0.011 & 0.062 & $+2.02$ \\
Mask2Former & Total energy (kWh)       & 0.2297\,$\pm$\,0.0013   & \sigDown{0.2151\,$\pm$\,0.0069}  & $-6.3\%$  & $[-8.0,-3.5]$   & 0.011 & 0.062 & $+2.02$ \\
Mask2Former & GPU energy (kWh)         & 0.1795\,$\pm$\,0.0015   & \sigDown{0.1654\,$\pm$\,0.0024}  & $-7.8\%$  & $[-8.8,-6.8]$   & 0.000 & 0.062 & $+6.01$ \\
Mask2Former & Wall-clock (s)           & 2154\,$\pm$\,40         & \sigDown{1980\,$\pm$\,24}        & $-8.1\%$  & $[-9.2,-7.2]$   & 0.000 & 0.062 & $+5.55$ \\
Mask2Former & GPU power (W)            & 298.7\,$\pm$\,7.4       & \sigTie{299.3\,$\pm$\,7.3}       & $+0.2\%$  & $[-1.0,+2.3]$   & 0.844 & 0.625 & $-0.09$ \\
\addlinespace
SegFormer & CO$_2$ (kg)                & 0.01658\,$\pm$\,0.00037 & \sigTie{0.01677\,$\pm$\,0.00056} & $+1.1\%$  & $[-1.2,+2.6]$   & 0.387 & 0.625 & $-0.43$ \\
SegFormer & Total energy (kWh)         & 0.04353\,$\pm$\,0.00097 & \sigTie{0.04402\,$\pm$\,0.0015}  & $+1.1\%$  & $[-1.2,+2.6]$   & 0.387 & 0.625 & $-0.43$ \\
SegFormer & GPU energy (kWh)           & 0.03325\,$\pm$\,0.0011  & \sigTie{0.03335\,$\pm$\,0.0011}  & $+0.3\%$  & $[-5.0,+5.1]$   & 0.926 & 1.000 & $-0.04$ \\
SegFormer & Wall-clock (s)             & 444.3\,$\pm$\,6.4       & \sigTie{427.7\,$\pm$\,11}        & $-3.7\%$  & $[-6.5,-0.7]$   & 0.100 & 0.188 & $+0.95$ \\
SegFormer & GPU power (W)              & 261.8\,$\pm$\,12        & \sigTie{273.5\,$\pm$\,18}        & $+4.5\%$  & $[-4.4,+13.3]$  & 0.423 & 0.438 & $-0.40$ \\
\addlinespace
EoMT & CO$_2$ (kg)                     & 0.02035\,$\pm$\,0.00094 & \sigTie{0.02258\,$\pm$\,0.0023}  & $+10.9\%$ & $[+2.6,+21.5]$  & 0.107 & 0.062 & $-0.92$ \\
EoMT & Total energy (kWh)              & 0.05342\,$\pm$\,0.0025  & \sigTie{0.05926\,$\pm$\,0.0061}  & $+10.9\%$ & $[+2.6,+21.5]$  & 0.107 & 0.062 & $-0.92$ \\
EoMT & GPU energy (kWh)                & 0.04035\,$\pm$\,0.0013  & \sigUp{0.04299\,$\pm$\,0.0022}   & $+6.5\%$  & $[+2.7,+11.1]$  & 0.049 & 0.062 & $-1.25$ \\
EoMT & Wall-clock (s)                  & 566.5\,$\pm$\,58        & \sigTie{604.4\,$\pm$\,71}        & $+6.7\%$  & $[+0.1,+14.2]$  & 0.166 & 0.312 & $-0.76$ \\
EoMT & GPU power (W)                   & 253.1\,$\pm$\,18        & \sigTie{254.8\,$\pm$\,16}        & $+0.7\%$  & $[-3.3,+4.1]$   & 0.767 & 0.625 & $-0.14$ \\
\midrule
\multicolumn{9}{c}{\cellcolor{gray!8}\textit{Supporting --- training loop only (validation excluded, warmup epoch trimmed)}} \\
\midrule
Mask2Former & GPU energy (train-only)  & 0.1384\,$\pm$\,0.0012   & \sigDown{0.1214\,$\pm$\,0.0027}  & $-12.3\%$ & $[-13.5,-10.6]$ & 0.000 & 0.062 & $+6.61$ \\
Mask2Former & Wall-clock (train-only)  & 1683\,$\pm$\,17         & \sigDown{1498\,$\pm$\,10}        & $-11.0\%$ & $[-11.6,-10.4]$ & 0.000 & 0.062 & $+12.04$ \\
\addlinespace
SegFormer & GPU energy (train-only)    & 0.02018\,$\pm$\,0.00043 & \sigTie{0.01969\,$\pm$\,0.0005}  & $-2.4\%$  & $[-5.1,+0.2]$   & 0.196 & 0.312 & $+0.69$ \\
SegFormer & Wall-clock (train-only)    & 331.5\,$\pm$\,6.5       & \sigTie{315.9\,$\pm$\,7.9}       & $-4.7\%$  & $[-7.8,-1.4]$   & 0.072 & 0.188 & $+1.09$ \\
\addlinespace
EoMT & GPU energy (train-only)         & 0.02055\,$\pm$\,0.00047 & \sigUp{0.02279\,$\pm$\,0.00067}  & $+10.9\%$ & $[+9.1,+13.4]$  & 0.001 & 0.062 & $-3.87$ \\
EoMT & Wall-clock (train-only)         & 357.6\,$\pm$\,0.92      & \sigUp{406.5\,$\pm$\,29}         & $+13.7\%$ & $[+8.3,+20.8]$  & 0.020 & 0.062 & $-1.68$ \\
\bottomrule
\end{tabular}}
\end{table*}

 As shown in Table~\ref{tab:energy_stats_full}, ST-LoRA behaves differently across the three architectures. For Mask2Former, 
ST-LoRA is \textit{greener} than FRE, even though both draw a 
comparable average GPU power; the reduction comes instead from a 
shorter training duration, which lowers total energy and emissions. 
For SegFormer-B2, ST-LoRA is essentially with lower training time but with higher power compared to FRE: while it still reduces the storage footprint, it consumes a similar amount 
of energy and produces similar emissions. For EoMT, however, 
ST-LoRA performs \textit{worse} than FRE, with higher values across 
all monitored metrics on average. This pattern holds consistently 
whether we consider the training loop alone or the full run 
including validation and warm-up.

We hypothesize that this architecture-dependent behaviour is driven 
by the fact that the frozen base-model weights still contribute 
substantially to total emissions, since forward and backward passes 
over the full-rank backbone remain energy-intensive regardless of 
how few parameters are actually updated. The proportion of 
fine-tuned parameters relative to the total model size also appears 
to play a role. As an example in Mask2Former, ST-LoRA fine-tunes roughly 1 million of 108 million parameters ($\approx 0.9\%$); for SegFormer-B2, roughly 3.5 million of 30{,}775{,}504 ($\approx 11.4\%$); and for EoMT, roughly 12 million of 320 million ($\approx 3.8\%$). Notably, Mask2Former - the architecture with 
the smallest fine-tuned fraction - shows the strongest efficiency gain, while EoMT - with the largest absolute number of fine-tuned parameters and a comparatively larger fine-tuned share than Mask2Former - shows the weakest outcome. This suggests that as 
the fine-tuned fraction grows, the relative savings from adapting only a small subset of parameters are increasingly offset by the fixed energy cost of the full-rank forward and backward passes 
through the frozen backbone.

This aligns with observations made in the original LoRA paper~\cite{hu2022lora}. The authors emphasize that LoRA's primary benefit lies in reducing the number of trainable parameters and the GPU memory required during fine-tuning, rather than eliminating the computational overhead of the adapters themselves. They explicitly note that the extra computation introduced by the low-rank matrices cannot be bypassed - it can only be minimized by constraining the adapter rank. Critically, the full pre-trained weights must still be loaded at deployment time, meaning LoRA does not reduce the model's footprint during inference so much as it adds a smaller number of parameters on top of it. Their reported efficiency gains are concentrated in memory storage and I/O, rather than raw compute: the only training-time improvement they document is a $25\%$ reduction for GPT-3~\cite{brown2020language}, and no FLOPs figures are provided for LoRA itself, beyond the general recommendation to keep the rank small in order to limit additional compute. 

This is further corroborated by \cite{ko2025lora}, who show that LoRA does not universally accelerate training across all model architectures and training configurations. In particular, they demonstrate that LoRA can be slower than full fine-tuning on  an NVIDIA A100 GPU\footnote{The same GPU model used throughout our training.}, with the effect depending on batch size and sequence length. They attribute this slowdown to how computation is scheduled on the GPU: since GPUs are optimized to execute a single kernel at a time, the additional low-rank adapter layers introduced by LoRA must be processed sequentially rather than fused into the base computation, creating bottleneck that can offset the training efficiency gains LoRA is otherwise expected to provide.

\cite{cherniuk-etal-2025-run} similarly observe that existing LoRA implementations rely on a fixed computation graph regardless of the underlying layer configuration, making training unnecessarily inefficient. They note that computational cost has been largely overlooked in the low-rank adaptation literature, with prior work focusing almost exclusively on parameter efficiency. To address 
this, they propose selecting the optimal forward and backward computation graph based on the linear layer's dimensions, the input dimensionality, and the LoRA rank, guided by FLOPs and runtime estimates; achieving faster training without sacrificing accuracy.

Finally for the LoRA ensemble \cite{halbheer2024lora}, the authors have also mentioned that their approach achieved the same training time compared to the \textit{Explicit ensemble}. Hence, Our efficiency results are therefore consistent with findings in the literature that LoRA's advantage is parameter, storage efficiency, and I/O efficiency, not a clear reduction in training-time computation.

\begin{table*}[t]
\centering
\caption{\textbf{Statistical comparison of segmentation quality and 
calibration, FRE vs.\ ST-LoRA, across three architectures.} 
Paired, $n{=}5$ seeds (42, 123, 456, 789, 1337). 
$\%\Delta = 100\cdot(\text{ST-LoRA}-\text{FRE})/\text{FRE}$; mIoU 
higher-better; ECE lower-better. $p$ columns report paired-$t$ / 
Wilcoxon (two-sided; Wilcoxon floors at $0.0625$ at $n{=}5$). 
Significance: $t$-test $p{<}0.05$ \textbf{and} 95\% bootstrap CI 
excludes 0. Colour 
scale: 
\colorbox{sigGreener!55}{ST-LoRA significantly greener}\,; 
\colorbox{tieColor}{no significant difference}\,; 
\colorbox{sigWorse!55}{ST-LoRA significantly worse}.}
\label{tab:calib_stats_full}
\setlength{\tabcolsep}{4pt}
\resizebox{\textwidth}{!}{
\begin{tabular}{llccccccc}
\toprule
\textbf{Arch} & \textbf{Metric} & \textbf{Dir} & \textbf{FRE} & \textbf{ST-LoRA} & \textbf{\%$\Delta$} & \textbf{95\% CI} & \textbf{$p$ ($t$/$W$)} & \textbf{$d_z$} \\
\midrule
Mask2Former & mIoU & $\uparrow$   & 0.6882\,$\pm$\,0.02      & \sigTie{0.6731\,$\pm$\,0.0072}    & $-2.2\%$  & $[-0.0293,-0.0009]$ & 0.144/0.188 & $-0.81$ \\
Mask2Former & ECE  & $\downarrow$ & 0.00316\,$\pm$\,0.0012   & \sigTie{0.003349\,$\pm$\,0.00035} & $+6.0\%$  & $[-0.0006,+0.0010]$ & 0.708/1.000 & $+0.18$ \\
\midrule
SegFormer & mIoU & $\uparrow$   & 0.6717\,$\pm$\,0.004    & \sigDown{0.6784\,$\pm$\,0.002}    & $+1.0\%$  & $[+0.0029,+0.0097]$ & 0.025/0.125 & $+1.57$ \\
SegFormer & ECE  & $\downarrow$ & 0.004362\,$\pm$\,0.00015 & \sigDown{0.003005\,$\pm$\,0.00045} & $-31.1\%$ & $[-0.0018,-0.0010]$ & 0.003/0.062 & $-2.95$ \\
\midrule
EoMT & mIoU & $\uparrow$   & 0.6873\,$\pm$\,0.0025 & \sigUp{0.6753\,$\pm$\,0.0024}    & $-1.7\%$ & $[-0.0146,-0.0098]$ & 0.001/0.062 & $-3.86$ \\
EoMT & ECE  & $\downarrow$ & 0.003285\,$\pm$\,0.0012  & \sigTie{0.003195\,$\pm$\,0.00067} & $-2.7\%$ & $[-0.0009,+0.0006]$ & 0.842/1.000 & $-0.10$ \\
\bottomrule
\end{tabular}}
\end{table*}

\subsection{Statistical Significance and Results Discussion}
\label{sec5.3}

We have compared the average performance of ST-LoRA and 
FRE across multiple perspectives. To strengthen our findings, we additionally perform paired statistical 
significance tests comparing both approaches seed-to-seed. We report 
the paired $t$-test~\cite{student1908probable}, a bootstrap 
confidence interval (CI)~\cite{efron1979bootstrap}, the Wilcoxon 
signed-rank test~\cite{wilcoxon1945individual} as a non-parametric 
alternative, and Cohen's $d_z$ as a paired effect size 
measure~\cite{cohen1988statistical}\footnote{ We use Crameri \textit{vik} diverging colormap from \url{https://pratiman-91.github.io/colormaps/}}. We use 5 seeds to represent the number of samples (4 degrees of freedom for t-student) and two-tailed for t-test. 

As shown in Table~\ref{tab:energy_stats_full}, the efficiency gains 
favour ST-LoRA for Mask2Former, are on par for SegFormer-B2, and 
favour FRE for EoMT. Overall, ST-LoRA consistently reduces the 
number of fine-tuned parameters, inference time, and storage 
requirements; however, for training time, energy, and emissions, 
further analysis is needed to investigate the relationship between 
the proportion of fine-tuned parameters relative to the total model 
size and the role of the frozen base model in driving these costs.

Table~\ref{tab:calib_stats_full} shows a similarly mixed picture: 
ST-LoRA is broadly on par with FRE for calibration (ECE), while for mIoU it oscillates between outperforming FRE (SegFormer-B2), matching it (Mask2Former), and underperforming it (EoMT). Hence, our approach emerges as a strong competitor to the FRE approach.

While ensembling  enhances calibration, it comes at the cost of reduced segmentation accuracy. Both effects stem from the same underlying mechanism, which is directly 
visible in the reliability diagrams of Figure~\ref{fig5}. 
All single models exhibit overconfidence predicted probabilities consistently exceed empirical accuracy because individual models tend to concentrate predictions in high-confidence bins regardless of whether 
those predictions are correct \cite{guo2017calibration}. Ensemble averaging softens this behavior; since the averaged probability across members is almost always lower than any individual member's maximum, predictions are redistributed from overconfident high-probability bins into more moderate bins where confidence and accuracy are better aligned. This redistribution moves the reliability curve upward toward the perfectly calibrated diagonal, 
reducing ECE. The same softening, however, occasionally overrides confident correct predictions, causing the marginal accuracy drop observed in mIoU.

We attribute the limited magnitude of both effects to insufficient ensemble diversity as members share architecture, initialization, and augmentation strategy, with only learning rate scheduling as a source of variation. This restricts inter-member disagreement, limiting both the calibration gains and the accuracy cost \footnote{More insights for member diversity analysis are in appendix~\ref{appen_a}.}. As ST-LoRA's rank constraint implicitly restricts optimization to a low-dimensional subspace where good solutions are more densely concentrated, which is also supported by \cite{halbheer2024lora}. This subspace retains sufficient capacity. Within it, learning rate variation may be enough to drive members toward solutions that are both competitive and complementary. In contrast, full-rank optimization spans a much larger space where the same perturbation may have proportionally less influence on the training trajectory, potentially causing members to converge to similar solutions. We leave empirical/theoretical validation of this hypothesis to future work. 

As conclusion:
\begin{tcolorbox}[
  enhanced,
  colback=blue!5!white,
  colframe=blue!75!black,
  title=\textbf{Smaller Everywhere, Faster Only Sometimes} ,
  fonttitle=\bfseries,
  coltitle=black,
  rounded corners,
  boxrule=0.4pt,
  attach boxed title to top left={yshift=-2mm, xshift=4mm},
  boxed title style={
    colback=blue!10!white,
    colframe=blue!75!black,
    boxrule=0.4pt,
    rounded corners
  }
]
Overall, ST-LoRA emerges as a strong competitor to FRE, consistently reducing storage requirements, the number of trainable parameters, 
and I/O bottlenecks across all three architectures. However, this 
efficiency advantage does not extend uniformly to training time: 
ST-LoRA's advantages over FRE are architecture-dependent, clearly 
outperforming on Mask2Former, performing on par on SegFormer-B2, and 
underperforming on EoMT. This indicates that LoRA hyperparameters 
and the target architecture jointly determine training efficiency, 
and that LoRA cannot guarantee faster training by design; 
identifying which architectural properties predict this behaviour 
is an important direction for future work.
\end{tcolorbox}

Having established that ST-LoRA is a viable and efficient alternative 
to full-rank ensemble, a natural question follows: \textit{is the baseline 
configuration optimal, or can targeted hyperparameter choices unlock 
further gains?} We investigate this in the next Experiment.

\section{Experiment 2: Ablation Study of ST-LoRA's Hyperparameters}
\label{sec6}
We conduct an extensive ablation study to identify the hyperparameters 
that most critically govern performance, stability, and parameter 
efficiency under ST-LoRA-based ensemble fine-tuning. Given the 
architectural heterogeneity of our benchmark  SgF-B2 on 
GrowliFlower-L versus M2F-Swin/B on BUP20, we treat each 
dataset/architecture pair independently, as hyperparameter sensitivity 
may differ substantially across them \footnote{We have unified the analysis in appendix~\ref{appen_a} for SegFormer and Mask2Former on the BUP20 dataset, in addition to adding a new architecture (EoMT) to further validate the generalizability of our ablation.}. Our analysis focuses on three 
factors: rank $r$, target components, and scaling factor $\alpha$, 
which together govern the capacity, scope, and magnitude of the 
low-rank adaptation \footnote{For better readability we have decided to leave the results for the appendix.}. Each experiment varies a single factor while 
holding all remaining settings fixed at the baseline configuration 
as mentioned in Table~\ref{tab:ablation_config}, allowing the effect of each choice to be isolated cleanly. The baseline is deliberately chosen as a 
reproducible standard reference rather than a pre-optimized 
configuration, so that improvements over it reflect genuine sensitivity 
rather than tuning artifacts. Interactions between hyperparameters are 
not explored, as exhaustive combinatorial search falls outside the 
scope of this study.

\subsection{LoRA Rank (r)}

Across both architectures, we observe a consistent trend: increasing the 
LoRA rank improves performance in terms of both segmentation accuracy and 
calibration as shown in Table~\ref{table15} and Table~\ref{table16}. For SegFormer-B2, rank $r=32$ emerges as a sweet spot between 
parameter efficiency and robust performance; higher ranks yield marginal 
further gains at the cost of a proportionally larger number of trainable 
parameters. Mask2Former-Swin/B exhibits a qualitatively similar trend but 
with a notably different sensitivity profile even at rank $r=2$, the model 
achieves reasonable segmentation performance with mIoU reaching $65.8\%$, 
suggesting that a low-dimensional subspace already captures much of the 
task-relevant variation. We hypothesize that this difference is not solely 
attributable to rank, but is also shaped by the substantially different 
adaptation strategies employed across the two models. SegFormer-B2 applies 
LoRA to both the self-attention projections and the position-wise 
feed-forward layers of the encoder, while the decoder is fully fine-tuned. 
In contrast, Mask2Former-Swin/B forgoes full fine-tuning entirely: the 
encoder feed-forward layers are adapted via LoRA, and the decoder is 
handled through low-rank injection into the query and key projections along 
with the class predictor. This contrasts full decoder fine-tuning versus 
LoRA-only decoder adaptation introduces a confounding factor beyond 
rank alone, and may explain why the two models exhibit different calibration 
sensitivity profiles as rank increases. We next investigate the effect of 
the targeted components directly.

\subsection{LoRA Target Components }

We investigate whether adapting the decoder alone is sufficient for 
segmentation. Keeping the encoder completely frozen, we inject LoRA
modules exclusively into the transformer decoder of Mask2Former-Swin/B and 
vary the rank. Across all rank values, no reasonable performance is observed 
and the overall results are substantially worse than those obtained when the 
encoder is also adapted as shown in Table~\ref{table17}. This is expected: the encoder is responsible for 
extracting task-relevant feature representations, and without adapting it 
to the target domain, the decoder operates on generic pre-trained embeddings 
that are ill-suited for fine-grained plant segmentation. Under these 
conditions, the decoder merely attempts to upsample frozen representations 
into a segmentation mask, offering little capacity to recover the necessary 
discriminative structure.

We then ask the complementary question: can the decoder be frozen 
entirely while adapting only the encoder? For Mask2Former-Swin/B, 
we freeze the decoder completely and evaluate all component 
configurations previously tested with decoder adaptation. The results 
reveal a consistent asymmetry: while a subset of configurations yields 
marginal mIoU improvements, nearly all exhibit severe calibration 
degradation across ECE, ACE, and MACE as illustrated in Table~\ref{table18}. This is attributable to the 
frozen decoder applying a fixed mapping optimized for the source 
domain, which is misaligned with the target domain's feature 
distribution regardless of how well the encoder is adapted. The 
decoder therefore acts as a calibration bottleneck; accurate 
enough to segment, but insufficiently adapted to produce reliable 
confidence estimates. These findings motivate the hybrid strategy 
adopted in our final configuration: LoRA-only adaptation of selected 
encoder components, with the decoder adapted via LoRA rather than 
frozen or fully fine-tuned.

We then follow the conventional LoRA adaptation strategy of targeting 
only the self-attention projections~\cite{hu2022lora},  injecting 
low-rank modules into $\m{Q}$, $\m{K}$, and $\m{V}$. Contrary to the gains 
reported in LLM settings, this configuration yields worse 
performance compared to the respective baselines for both architectures, as 
reflected by both mIoU and ECE. This suggests that attention-only adaptation, 
while effective for generic image classification and language tasks \cite{muhlematter2024lora,Balabanov2024UncertaintyQI}, is insufficient for dense prediction in agricultural imagery, where spatial 
detail and pixel-level calibration impose stricter representational demands 
on the adapted layers.

We now turn to an intensive component ablation, examining how each 
encoder component contributes to segmentation quality and uncertainty 
calibration. Our analysis focuses on two families of components: the 
self-attention projections and the position-wise Multi-Layer Perception (MLP) 
layers, as these constitute the two primary sites of learnable 
computation within a transformer encoder block.

Adapting the query projection alone yields mixed results: for 
Mask2Former-Swin/B, performance remains comparable to the baseline 
despite the added parameters, whereas for SegFormer-B2 it surpasses 
both the baseline and the conventional $(q, k, v)$ configuration. 
Extending adaptation to the key and value projections jointly improves 
over query-only across both architectures, again outperforming the 
baseline and the conventional attention-only setting. However, the 
most pronounced gains are observed when the MLP layers are included. 
Across both models and all component combinations tested, incorporating 
MLP layers consistently yields the strongest improvements in both mIoU 
and calibration metrics while having the smaller number of parameters, indicating that these layers carry the dominant 
share of task-relevant adaptability in the context of agricultural 
semantic segmentation.

This observation is further supported by a converging body of evidence from both the text and vision literature. In the text domain, \cite{dong2021attention} showed that transformer networks without skip connections or MLP layers converge doubly exponentially to a rank-1 matrix, rendering them unable to learn, establishing MLP layers as structurally necessary rather than merely beneficial. \cite{geva2021transformer} further demonstrated that MLP layers act as key-value memories that store and retrieve factual knowledge, with deeper layers encoding increasingly rich semantic content. From an adaptation perspective, \cite{hetowards} found that under a restricted parameter budget, MLP layers yield better adaptation than attention projections, and \cite{zhangadaptive} confirmed through extensive experimentation that MLP layers consistently outperform attention at all ranks, matching full fine-tuning at higher ranks.

In the vision domain, the evidence is equally compelling. \cite{melas2021you} constructed a vision transformer using only MLP layers and achieved performance comparable to the full model, while the attention-only counterpart degraded severely. \cite{zhang2024you} showed that attention is not needed at every layer, with selective removal reducing computational cost while improving performance. \cite{wang2022shift} replaced self-attention entirely with a mean-shift operation and obtained competitive results, and \cite{venkataramananskip} demonstrated that skipping attention modules via skip connections incurs minimal performance loss. Complementing this, \cite{zhang2022moefication} revealed that MLP layers exhibit a Mixture-Of-Experts (MOE) behaviour, with only a sparse subset of neurons activating per input, suggesting that their representational capacity is both rich and highly structured.

\begin{table*}[t]
\centering
\caption{\textbf{Paired significance comparison: FFN-only vs.\ 
attention + FFN target modules.} $d = \text{FFN-only} - (\text{attn+FFN})$; 
$d{>}0$ indicates FFN-only is higher. Paired, $n{=}5$ seeds. 95\% CI 
via 20k-resample bootstrap. $p$ columns report paired-$t$ / Wilcoxon 
(two-sided; Wilcoxon floors at $0.0625$ at $n{=}5$). Verdict is oriented so that \textit{Win} favours 
FFN-only (better/cheaper). Colour 
scale: 
\colorbox{sigGreener!55}{ST-LoRA significantly greener}\,; 
\colorbox{tieColor}{no significant difference}\,; 
\colorbox{sigWorse!55}{ST-LoRA significantly worse}}
\label{tab:ffn_vs_attn_stats}
\setlength{\tabcolsep}{15pt}
\resizebox{\textwidth}{!}{
\begin{tabular}{llccccc}
\toprule
\textbf{Arch} & \textbf{Metric} & \textbf{Mean $d$} & \textbf{95\% CI} & \textbf{$p$ ($t$/$W$)} & \textbf{$d_z$} \\
\midrule
SegFormer & mIoU $\uparrow$          & \sigTie{$+0.0010$} & $[-0.0019,+0.0047]$ & 0.614/0.625 & $+0.24$  \\
SegFormer & ECE $\downarrow$         & \sigDown{$-0.0072$} & $[-0.0118,-0.0017]$ & 0.065/0.125 & $-1.13$  \\
SegFormer & GPU energy (kWh) $\downarrow$ & \sigDown{$-0.0022$} & $[-0.0030,-0.0014]$ & 0.010/0.062 & $-2.04$  \\
SegFormer & CO$_2$e (kg) $\downarrow$     & \sigDown{$-0.0013$} & $[-0.0017,-0.0008]$ & 0.009/0.062 & $-2.14$  \\
SegFormer & Wall-clock (min) $\downarrow$ & \sigDown{$-0.7841$} & $[-1.1030,-0.4652]$ & 0.014/0.062 & $-1.86$  \\
\midrule
EoMT & mIoU $\uparrow$          & \sigUp{$-0.0030$} & $[-0.0047,-0.0015]$ & 0.032/0.062 & $-1.44$  \\
EoMT & ECE $\downarrow$         & \sigTie{$+0.0007$} & $[-0.0003,+0.0020]$ & 0.354/0.438 & $+0.47$   \\
EoMT & GPU energy (kWh) $\downarrow$ & \sigDown{$-0.0054$} & $[-0.0087,-0.0022]$ & 0.043/0.062 & $-1.31$ \\
EoMT & CO$_2$e (kg) $\downarrow$     & \sigTie{$-0.0009$} & $[-0.0032,+0.0014]$ & 0.533/0.625 & $-0.30$  \\
EoMT & Wall-clock (min) $\downarrow$ & \sigTie{$-1.2636$} & $[-3.0206,+0.2225]$ & 0.241/0.625 & $-0.62$  \\
\midrule
Mask2Former & mIoU $\uparrow$          & \sigTie{$-0.0036$} & $[-0.0111,+0.0044]$ & 0.471/0.438 & $-0.36$  \\
Mask2Former & ECE $\downarrow$         & \sigTie{$+0.0005$} & $[-0.0001,+0.0011]$ & 0.252/0.312 & $+0.60$  \\
Mask2Former & GPU energy (kWh) $\downarrow$ & \sigTie{$-0.0007$} & $[-0.0077,+0.0076]$ & 0.871/0.625 & $-0.08$ \\
Mask2Former & CO$_2$e (kg) $\downarrow$     & \sigTie{$-0.0056$} & $[-0.0135,+0.0007]$ & 0.252/0.312 & $-0.60$  \\
\bottomrule
\end{tabular}}
\end{table*}

To verify empirically and align with the theoretical and empirical findings in the literature, we train all three architectures using two configurations: adapting the feed-forward layers only, and adapting both the feed-forward layers and the attention mechanism jointly.\footnote{Both configurations use rank $r{=}8$.} As shown in Table~\ref{tab:ffn_vs_attn_stats}, removing attention adaptation leaves Mask2Former largely unaffected, with only marginal and non-significant changes in mIoU and ECE, while computational efficiency remains essentially unchanged while having a lower number of parameters. For SegFormer-B2, we observe a small, non-significant drop in mIoU; however, FFN-only adaptation brings significant improvements in both ECE and computational efficiency, leading to reduced GPU energy, CO$_2$ emissions, and wall-clock time. 

EoMT, however, stands out as the exception. FFN-only adaptation results in a significant drop in mIoU, no meaningful change in ECE, and notably, a significant reduction in GPU energy despite comparable wall-clock time. This indicates that adaptation feed forward layers only can be beneficial for computational efficiency while affecting the predictive power slightly. 

Taken together, this points to the same conclusion: MLP layers are the dominant site of task-relevant computation and adaptation in transformer architectures, in both text and vision settings. Our ablation results and significance align with and extend this finding to the dense agricultural prediction regime, where attention-only adaptation consistently underperforms MLP-inclusive configurations across the three architectures studied with more computational power needed.

\subsection{LoRA Scaling Factor ($\alpha$)}

Having established rank and target components as the two most 
influential hyperparameters, we turn to the scaling factor $\alpha$. 
Recalling the LoRA adaptation equation but with the previously omitted $\delta_{r} = \nicefrac{\alpha}{r}$ for simplicity,
\begin{equation}
    \ical{H} = \theta_0\m{X} + \Delta\theta\m{X} 
             = \theta_0\m{X} + \delta_{\text{r}}\m{B}\m{A}\m{X}.
\end{equation}
The effective magnitude of the low-rank update is governed by 
$\delta_{r} = \nicefrac{\alpha}{r}$, the ratio of the scaling factor 
to the rank. Rather than treating $\alpha$ in isolation, this 
formulation makes clear that it is $\delta_{r}$ that controls how 
strongly the adapter output is blended into the pre-trained 
representation. We therefore ablate across $\alpha$ values at fixed 
rank to characterize the sensitivity of both segmentation accuracy 
and calibration to this ratio.

For SegFormer-B2, we observe a clear degradation in performance as 
$\alpha$ increases, manifesting in both mIoU and calibration metrics. 
Since the effective update magnitude is governed by $\delta_r = 
\nicefrac{\alpha}{r}$, a large $\alpha$ at fixed rank amplifies the 
low-rank update $\Delta\theta = \delta_r \m {B}\m {A}$ to a 
degree that distorts the pre-trained feature representations, 
effectively overriding the visual prior encoded during pre-training 
rather than refining it as shown in Table~\ref{table15}. Mask2Former-Swin/B exhibits the opposite trend: performance improves as $\alpha$ increases, suggesting that a 
stronger update signal is beneficial in this setting. Nevertheless, 
this gain saturates and reverses once the ratio $\delta_r$ grows large 
enough that $\Delta\theta \gg \theta_0$, at which point the adapter 
update dominates the frozen weights and degrades learning, 
consistent with the representation distortion observed for SegFormer-B2. Notably, Mask2Former-Swin/B appears considerably less sensitive to $\alpha$ overall, implying that the scaling factor does not act in isolation but interacts with rank and the choice of adapted components as shown in Table~\ref{table16}. 

The original LoRA formulation~\cite{hu2022lora} acknowledges this 
sensitivity and recommends treating $\alpha$ as a regularization-like 
hyperparameter, noting that setting $\alpha = r$ provides a neutral 
starting point where the adapter contributes without overwhelming the 
frozen weights. For high ranks $r>>128$ \cite{Biderman2024LoRALL} suggest using $\alpha=2r$ but this is in text domain and needs more verification for dense prediction in vision, this reinforces the broader observation that LoRA 
hyperparameters should not be tuned independently, and that the 
relative magnitude of the adapter update with respect to the 
pre-trained weights is a more principled diagnostic than any single 
hyperparameter alone \cite{hu2022lora, buyukakyuz2024olora}.

\begin{figure*}[t]
    \centering
    \includegraphics[width=1\linewidth]{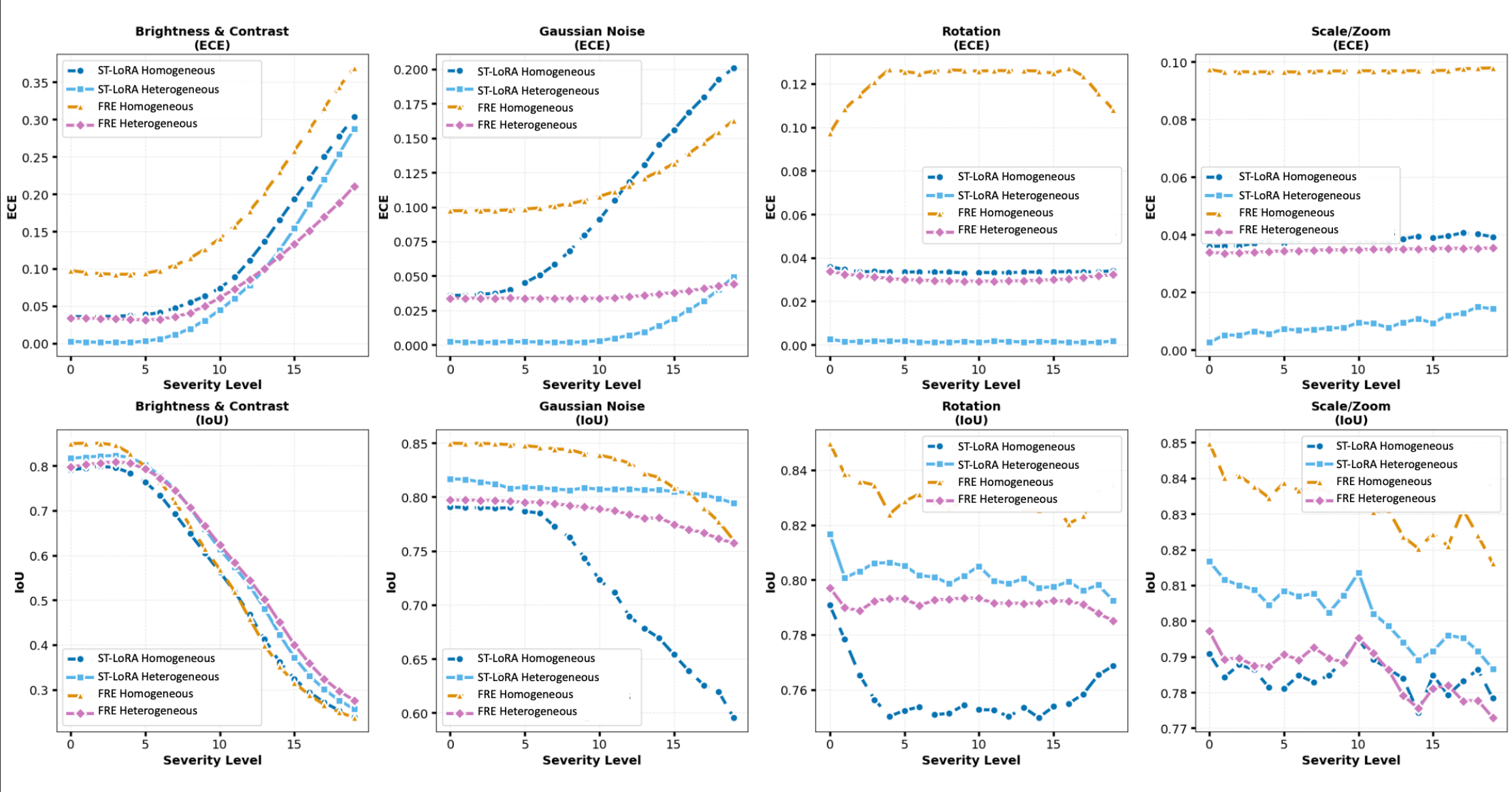}
    \caption{\textbf{Ensemble robustness under covariate shift.} 
    Homogeneous vs.\ heterogeneous ensemble variants for ST-LoRA and 
    FRE on GrowliFlower-L across 80 shift conditions.}
    \label{fig6}
\end{figure*}


Finally, The relative insensitivity of Mask2Former-Swin/B to $\alpha$ likely 
reflects two compounding factors. First, unlike SegFormer-B2 where 
the fully fine-tuned decoder creates a rigid dependency on the 
quality of the adapted encoder representations, Mask2Former-Swin/B 
distributes adaptation entirely through LoRA modules across both 
encoder and decoder. Second, the adapted components in Mask2Former-Swin/B 
are predominantly MLP layers, which may be intrinsically less sensitive to 
scaling perturbations than the lower-dimensional attention projections 
targeted in SegFormer-B2. We note that this remains a hypothesis, as 
disentangling the individual contributions of architecture, component 
choice, and scaling interactions would require controlled experiments 
beyond the scope of this ablation. Our results confirm this intuition in the vision segmentation regime and suggest that conservative scaling is 
particularly important when adapting pre-trained encoders to 
fine-grained, data-limited agricultural datasets. Hence we provide a practitioner's guide according to the previous experiments for ST-LoRA as a starting point:

\begin{tcolorbox}[
  enhanced,
  colback=blue!5!white,
  colframe=blue!75!black,
  title=\textbf{ST-LoRA Hyperparameters Starting Point},
  fonttitle=\bfseries,
  coltitle=black,
  rounded corners,
  boxrule=0.4pt,
  attach boxed title to top left={yshift=-2mm, xshift=4mm},
  boxed title style={
    colback=blue!10!white,
    colframe=blue!75!black,
    boxrule=0.4pt,
    rounded corners
  }
]
\sloppy
\textbf{Always adapt the encoder }; freezing it entirely produces near-random 
performance regardless of rank. \textbf{Target feed-forward layers 
first}: contrary to standard LoRA practice, FF layers 
consistently outperform attention-only adaptation for dense prediction, 
with fewer parameters. \textbf{Do not freeze the decoder entirely}, as it acts 
as a calibration bottleneck regardless of encoder quality; adapt it 
via LoRA instead. For \textbf{rank} $r{\in}[8,32]$ is recommended. Finally, 
keep $\delta_r{=}\nicefrac{\alpha}{r}$ conservative: setting 
$\alpha{=}r$ is a safe default, and large $\delta_r$ degrades both 
accuracy and calibration.
\end{tcolorbox}

However, a well-configured model is 
a necessary but not sufficient condition for reliable deployment as
agricultural imaging conditions vary substantially across seasons, 
lighting, and sensing conditions, making robustness under distribution 
shift equally critical. We therefore turn to two complementary 
questions: \textit{does ensemble member diversity affect calibration stability 
when the input distribution shifts, and do the component choices 
identified above retain their advantage beyond the training 
distribution}?

\section{Experiment 3: Ensemble Diversity and Architectural Sensitivity under Distribution Shift: }
\label{sec7}
We evaluate robustness under covariate shift through two 
complementary analyses. The first examines how ensemble member diversity 
affects robustness, using SegFormer-B2 on GrowliFlower-L. The second 
investigates the sensitivity of different LoRA component configurations 
to distribution shift, using Mask2Former-Swin/B on BUP20. In both cases, 
covariate shift is simulated by applying controlled corruptions to the 
test set, the full experimental setup for both 
analyses is summarized in Table~\ref{tab:dist_shift_setup}.

\subsection{Ensemble diversity under distribution shift.}
We compare two ensemble construction strategies for both FRE and ST-LoRA 
to assess the impact of member diversity on calibration robustness. 
\textit{Homogeneous ensembles} rely solely on learning rate scheduling 
as a source of diversity \footnote{More insights for member diversity analysis are in appendix~\ref{appen_a}.}, with no data augmentation applied during training; members therefore differ only in their optimization 
trajectories. \textit{Heterogeneous ensembles} additionally introduce 
per-member data augmentation policies, incorporating random rotation 
(uniformly sampled from $\{0^{\circ}, 25^{\circ}\}$) and horizontal flipping 
(probability $0.5$), applied consistently to both images and 
segmentation masks. This introduces 
diversity at both the optimization and data levels, which
leads to more robust ensemble predictions and better-calibrated 
uncertainty estimates under distribution shift.

Across all distribution shifts as shown in Figure~\ref{fig6}, ST-LoRA achieves 
performance competitive with FRE, while exhibiting greater robustness 
to increasing shift severity maintaining lower ECE values with only 
a $3$--$5\%$ reduction in mIoU. For FRE, homogeneous ensembles achieve 
higher mIoU but substantially worse calibration compared to their 
heterogeneous counterparts. The heterogeneous strategy is therefore 
critical for calibration quality in full-rank ensembles. ST-LoRA, by 
contrast, benefits from heterogeneity in both metrics simultaneously: 
heterogeneous ST-LoRA outperforms its homogeneous counterpart in mIoU 
and ECE alike, achieving the best calibration across all shift 
conditions while maintaining segmentation quality competitive with FRE variants.

We attribute this asymmetry to the reduced parameter space 
of LoRA adapters: the constrained adaptation subspace acts as an 
implicit regularizer, allowing ST-LoRA members to leverage 
augmentation-induced diversity more efficiently than full-rank models 
operating under the same limited training budget. Furthermore, we observe that SgF-B2 performance is deteriorating under high brightness/contrast shift compared to other shifts. This is due to high severity which completely distorts the input image to the model, however around this shift level of $10$, the model still achieves $64\%$ mIoU, and this robustness is due to the design of SegFormer which has been mentioned in \cite{xie2021segformer}.

\subsection{Component Sensitivity under Distribution Shift.}
We evaluate how ST-LoRA hyperparameter choices - specifically rank, 
target components, scaling factor, and dropout - affect calibration 
robustness when the input distribution shifts. For each hyperparameter, 
we compare how varying a single setting relative to the baseline 
configuration affects robustness across the 19 corruption conditions 
described in Table~\ref{tab:dist_shift_setup}.

\begin{figure*}[t]
    \centering
    \includegraphics[width=1\linewidth]{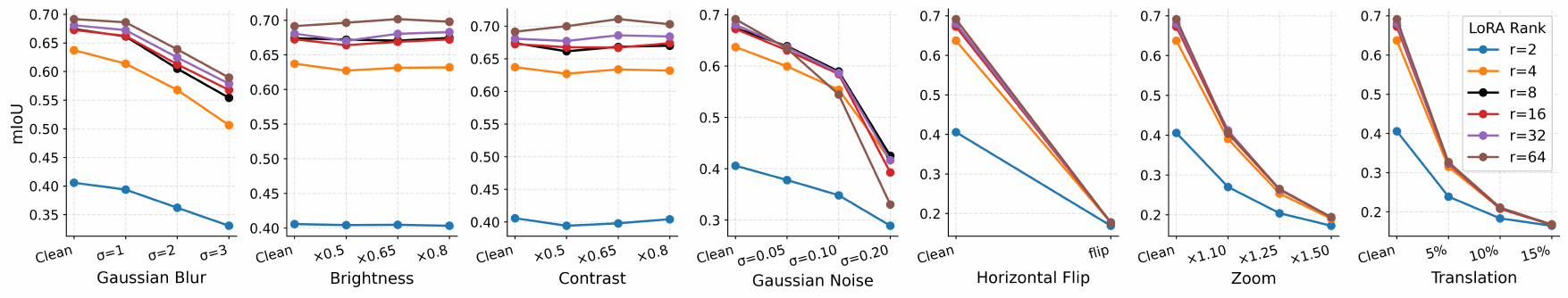}
    \caption{\textbf{mIoU under distribution shift as a function of LoRA 
rank $r$, evaluated across seven corruption types at three severity 
levels. Each curve corresponds to a distinct shift type}.}
    \label{miourank}
\end{figure*}

\begin{figure*}[t]
    \centering
    \includegraphics[width=1\linewidth]{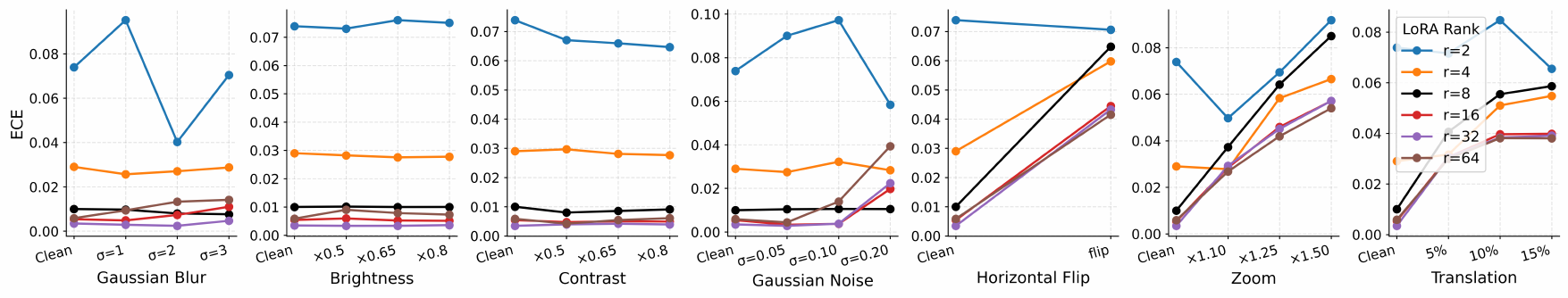}
    \caption{\textbf{ECE under distribution shift as a function of LoRA 
rank $r$, evaluated across seven corruption types at three severity 
levels. Each curve corresponds to a distinct shift type}.}
    \label{ecerank}
\end{figure*}

\textbf{Rank.} Across all corruption types and severity levels, a 
consistent pattern emerges: low-rank configurations are brittle under 
distribution shift, exhibiting substantially worse calibration alongside 
degraded segmentation quality as shown in Figure~\ref{miourank}. This instability is most pronounced at 
$r=2$ as shown in Figure~\ref{ecerank}, where calibration metrics deteriorate sharply even under mild 
photometric perturbations, suggesting that a low-dimensional adaptation 
subspace lacks sufficient capacity to maintain well-calibrated predictions 
beyond the training distribution. Increasing the rank to $r=16$ yields a 
notable improvement in both robustness and calibration stability across 
shift types, establishing it as a practical lower bound for reliable 
deployment under covariate shift. However, continuing to increase the 
rank beyond this point introduces diminishing returns: while segmentation 
accuracy may marginally improve, calibration under distribution shift 
does not benefit proportionally and can even degrade at higher ranks as shown in Figure~\ref{ecerank} which has been also shown by \cite{halbheer2024lora}. The additional computational cost of high-rank adaptation therefore does not translate into a robustness advantage, reinforcing 
$r \in [8, 16]$ as the recommended operating range for ST-LoRA on 
Mask2Former-Swin/B under distribution shift.

\textbf{Target components.} Ranking configurations by mIoU and ECE 
robustness under distribution shift reveals a persistent tension between 
segmentation accuracy and calibration as shown in Figure~\ref{miouablat} and Figure~\ref{eceablat}: the configurations that best 
preserve accuracy under shift are not the same as those that best 
preserve calibration. Specifically, $\text{base} + k_e$\footnote{$q,k,\text{and} \space v$ refer to the Query, Key, and Value in Transformer model, and the subscript $e$ refers to the encoder.} ranks second 
on mIoU but drops to sixth on calibration, while $\text{base} + q_e$ 
ranks third on mIoU yet last on calibration. The configuration achieving 
the strongest mIoU robustness overall is $\text{base} + k_e + v_e$, 
whereas $\text{base} + q_e + k_e$ leads on calibration robustness. 
The baseline configuration (FF layers, Class predictor) occupies 
a rank of approximately five on both objectives, placing it consistently 
in an intermediate position that avoids the extremes of either. This 
trade-off mirrors the accuracy/calibration tension observed in the in-distribution ablation and suggests that no single component 
configuration simultaneously dominates both objectives under covariate 
shift.

Furthermore, Cross-referencing with the in-distribution ablation results 
reveals that the baseline and the $\text{base} + q_e + k_e$ 
configuration yield comparable performance across both mIoU and ECE 
relative to the best-performing configurations for both SegFormer-B2 
and Mask2Former-Swin/B. This consistency across architectures, datasets, 
and evaluation regimes - in-distribution and under shift - suggests 
that adapting query and key projections alongside the encoder 
MLP layers provides a robust and transferable operating point.
$\text{baseline} + q_e + k_e$ therefore represents the most principled 
choice, offering a stable balance between segmentation quality and 
uncertainty reliability without requiring architecture-specific tuning.

\textbf{Scaling factor $\alpha$.} The effect of $\alpha$ on robustness 
under distribution shift exhibits a clear asymmetry between segmentation 
accuracy and calibration. For mIoU, increasing $\alpha$ initially 
improves robustness, suggesting that a stronger adapter signal helps the 
model maintain discriminative representations under input perturbations. 
However, this benefit saturates and reverses beyond a certain threshold, 
where the amplified low-rank update begins to distort the pre-trained 
prior rather than refine it as shown in Figure~\ref{miouablat} and Figure~\ref{eceablat}, consistent with the in-distribution 
degradation observed at high $\delta_r = \nicefrac{\alpha}{r}$ values. 
For calibration under distribution shift, the relationship is 
qualitatively different: increasing $\alpha$ consistently yields better 
calibrated predictions, with ECE improving monotonically across shift 
types and severity levels. This suggests that a stronger adapter update enhance robustness when the input deviates from the training distribution. The two objectives therefore pull in opposite directions as $\alpha$ increases, reinforcing the broader observation 
that accuracy robustness and calibration robustness are not jointly 
optimizable through a single hyperparameter, and that the choice of 
$\alpha$ should be guided by the deployment priority - higher $\alpha$ 
for better-calibrated uncertainty, lower to moderate $\alpha$ for 
stronger segmentation accuracy under shift but we leave the empirical validation for future work.

\textbf{LoRA Dropout} $p$. The effect of LoRA dropout on robustness 
under distribution shift follows a non-monotonic pattern for 
segmentation accuracy: both the lowest ($p=0.1$) and highest ($p=0.3$) 
dropout values yield degraded mIoU under shift, while intermediate 
values $p \in [0.2, 0.25]$ achieve the strongest robustness as shown in Figure~\ref{miouablat} and Figure~\ref{eceablat}. Small dropout allows the adapter to overfit to 
in-distribution features that do not generalize under shift, while 
too much dropout disrupts the adapter's capacity to maintain 
discriminative representations altogether. A moderate dropout rate 
therefore strikes the best operating point between preventing overfitting 
and preserving representational capacity.

\begin{figure*}[t]
    \centering
    \includegraphics[width=1\linewidth]{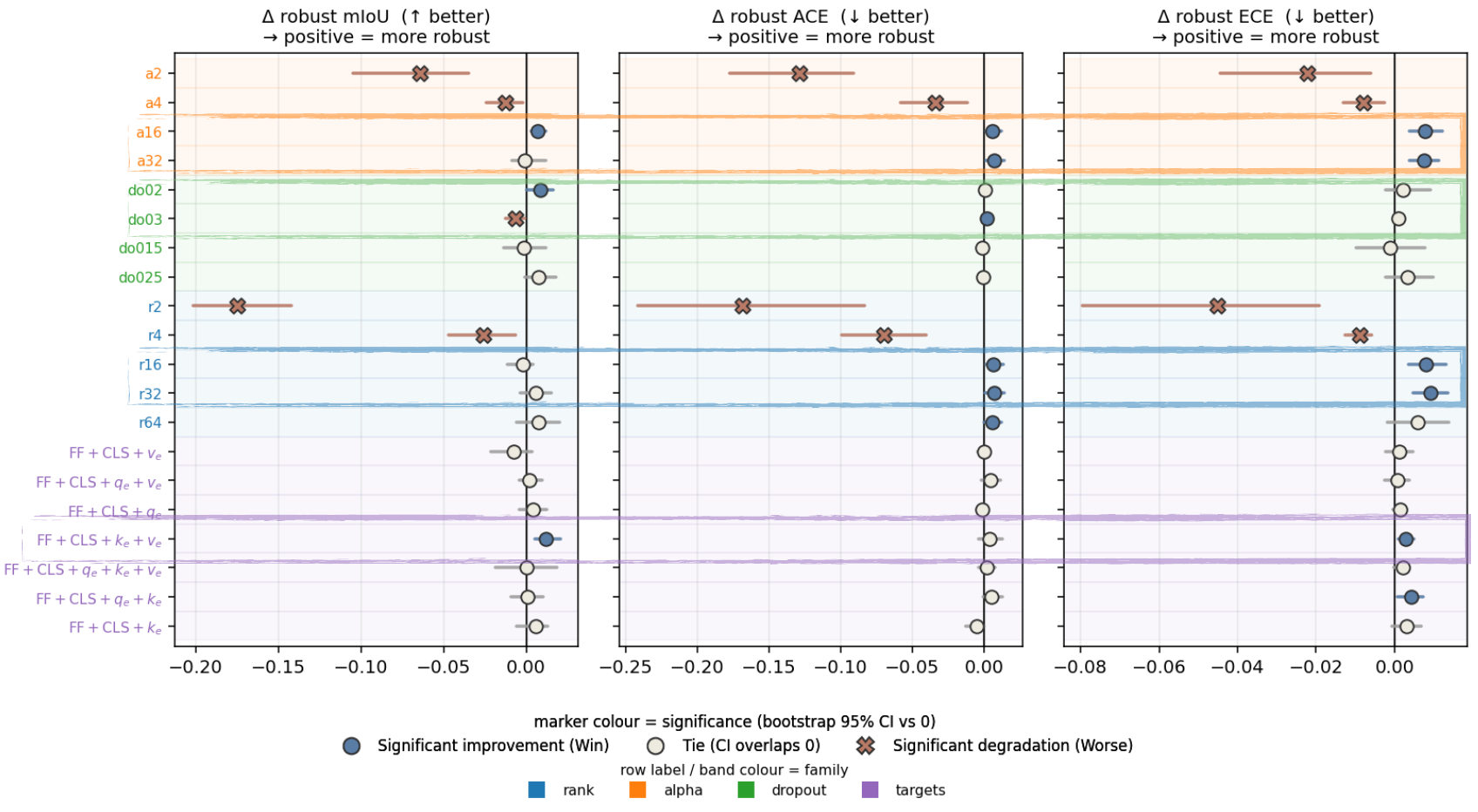}
    \caption{\textbf{Statistical comparison of segmentation quality and calibration of
ST-LoRA hyperparameter configurations under covariate shift.}
Paired, $n{=}5$ seeds (42, 123, 456, 789, 1337). Each panel plots the \emph{sign-oriented improvement} of a robust metric for every configuration relative to the baseline: $\mathrm{sign}\cdot(\text{variant}-\text{base})$, with $\mathrm{sign}{=}{+}1$ for the higher is better metric (mIoU) and $\mathrm{sign}{=}{-}1$ for the lower is better calibration metrics (ACE, ECE). A positive value therefore
\emph{always} denotes a more robust configuration -higher mIoU, or lower
ACE/ECE- regardless of the metric's native direction.
Dots are the mean $\Delta$ over seeds; whiskers are the $95\%$ percentile
bootstrap CI. Configurations are grouped by family (rank, alpha, dropout, target modules); $\mathrm{FF}{+}\mathrm{CLS}$
denotes the feed-forward $+$ class-predictor baseline and $q_e,k_e,v_e$ the
added encoder query, key, and value of self attention layer. }
    \label{lora_parameters_robust}
\end{figure*}

For calibration, the picture is more nuanced and shift-type selective. 
Under photometric corruptions, $p=0.3$ yields the best calibration 
robustness, suggesting that aggressive dropout encourages the model 
to be robust when pixel statistics deviate from the training distribution. Under geometric corruptions, however, 
$p=0.3$ underperforms relative to lower dropout values, which instead 
produce more stable calibration. This shift-type selectivity of dropout echoes the broader finding that photometric and geometric corruptions engage 
different aspects of the model's learned representations, and 
reinforces the recommendation to treat calibration robustness under 
the two shift families as distinct design objectives.

For statistical significance, as shown in Figure~\ref{lora_parameters_robust}, the low rank/alpha values ($r \leq 4$, $\alpha \leq 4$ ) are hurting the segmentation quality and calibration. While for larger values the gains are significant. For the targeted components we first notice that as an overview there are no big differences regarding the attention components being adapted with the FF layers. However we observe that $\text{MLP}{+}q_e{+}k_e$ or $\text{MLP}{+}k_e{+}v_e$ enhances the calibration with respect to other configurations but for less fine tuned parameters we could start by $\text{MLP}{+}q_e$. 

Finally, \cite{guo2017calibration} highlight an inherent tension between minimizing classification error and achieving well-calibrated predictions. We observe a similar trade-off in ST-LoRA: accuracy and calibration emerge as competing objectives, and we hypothesize that this competition has a 
low-rank geometric interpretation. Drawing on the notion of intrinsic 
dimensionality~\cite{aghajanyan2021intrinsic}, the low-rank subspace, 
sufficient for optimizing the segmentation objective, may not be 
expressive enough to simultaneously ensure reliable calibration; 
particularly under distribution shift, where the adapter must 
generalize beyond the fine-tuning data manifold. This suggests that 
accuracy and calibration place structurally different demands on the 
adaptation subspace, and that jointly optimizing both may require 
explicit calibration-aware training objective.

According to the previous observations, we give a recommendation for hyperparameters selection under covariate shift:

\begin{tcolorbox}[
  enhanced,
  colback=blue!5!white,
  colframe=blue!75!black,
  title=\textbf{Hyperparameters under covariate Shift},
  fonttitle=\bfseries,
  coltitle=black,
  rounded corners,
  boxrule=0.4pt,
  attach boxed title to top left={yshift=-2mm, xshift=4mm},
  boxed title style={
    colback=blue!10!white,
    colframe=blue!75!black,
    boxrule=0.4pt,
    rounded corners
  }
]
\sloppy
\textbf{Start with} $r{\in}[8,16]$: low ranks ($r{\leq}4$) are brittle under all 
shift types; higher ranks add cost without calibration benefit. 
\textbf{Prefer} $\text{MLP}{+}q_e{+}k_e$ for the most consistent 
accuracy--calibration balance across architectures and shift types. 
Tune $\delta_r$: lower $\alpha$ when calibration is 
the priority, raise it when accuracy matters more; $\alpha{=}r$ \textbf{is 
the safe default}. For dropout, $p{\in}[0.2,0.25]$ maximizes accuracy 
robustness; $p{=}0.3$ is better when photometric shift dominates. 
Accuracy and calibration under shift are competing objectives; 
no single configuration wins on both.
\end{tcolorbox}

Having established ST-LoRA's advantages over full-rank ensemble and 
characterized its behaviour under distribution shift, we turn to a 
broader question: \textit{how does ST-LoRA compare against 
established efficient SOTA uncertainty quantification methods?} We 
benchmark against three approaches - Snapshot 
Ensemble, MC Dropout, and DDU - spanning deterministic, and stochastic uncertainty estimation paradigms, and six post-hoc calibration methods.

\section{Experiment 4: Benchmarking ST-LoRA Against Efficient 
Uncertainty Methods}
\label{sec8}




We compare ST-LoRA against three efficient uncertainty estimation 
baselines whose 
configurations are detailed in Table~\ref{tab:method_configs}. 
Integrated ensemble methods are excluded due to architectural and 
compatibility constraints discussed in Section~\ref{sec2}. To ensure 
a fair comparison, all methods share an identical training setup: the 
same cosine annealing learning rate schedule and the same heterogeneous 
augmentation pipeline, isolating the effect of the uncertainty estimation strategy itself.

The augmentation pipeline consists of two families of transformations.
\textit{Geometric} augmentations, applied jointly to image and mask,
include random horizontal flipping ($p{=}0.5$) and lateral translation
(up to $10\%$ of image width, $p{=}0.5$). \textit{Photometric}
augmentations, applied to the image only, include colour jitter
(brightness $\pm0.3$, contrast $\pm0.3$, saturation $\pm0.1$),
Gaussian blur ($\sigma{\in}[0.5,1.5]$, $p{=}0.3$), Gaussian noise
($\sigma{=}0.02$, $p{=}0.2$), spatially non-uniform depth-of-field
blur ($p{=}0.3$) simulating centre-sharp optics via a Gaussian spatial
weight mask, and random erasing of up to three patches per image
($p{=}0.3$, patch area $\in[2\%,15\%]$ of image) to simulate leaf
occlusion and canopy overlap. Additionally, CutMix \cite{yun2019cutmix} ($p{=}0.5$) is
applied at the dataset level, randomly replacing a rectangular region
of one image and its corresponding mask labels with those from another
training sample, encouraging the model to learn from partially
occluded scenes. \footnote{Mask interpolation uses nearest-neighbour throughout
to preserve integer label integrity; image interpolation uses bilinear}. 

\begin{table*}[t]
\centering
\caption{\textbf{In-distribution (ID) performance of ST-LoRA against efficient 
uncertainty quantification baselines and post-hoc calibrators on Mask2Former-Swin/B 
(BUP20, mean\,$\pm$\,std over 5 seeds)}. FRE is a snapshot ensemble built on top 
of a fully fine-tuned model. MC Dropout uses $p=0.5$. 
DDU: Deep Deterministic Uncertainty \cite{Mukhoti2023}. 
$\uparrow$: higher is better; $\downarrow$: lower is better. 
Best result per metric in \textbf{bold}.}
\label{tab:id_performance_m2f}
\resizebox{1\textwidth}{!}{
\begin{tabular}{lccccccc}
\toprule
\textbf{Method} & \textbf{mIoU} $\uparrow$ & \textbf{ECE} $\downarrow$ & \textbf{MECE} $\downarrow$ & \textbf{ACE} $\downarrow$ & \textbf{MACE} $\downarrow$ & \textbf{Brier} $\downarrow$ & \textbf{NLL} $\downarrow$ \\
\midrule
\multicolumn{8}{l}{\textit{Uncertainty estimation methods}} \\
\midrule
FRE  
    & \textbf{0.6990 $\pm$ 0.0259} 
    & 0.0068 $\pm$ 0.0068 
    & 0.1792 $\pm$ 0.1613 
    & 0.0047 $\pm$ 0.0037
    & 0.0209 $\pm$ 0.0242 
    & \textbf{0.0171 $\pm$ 0.0003 }
    & \textbf{0.0389 $\pm$ 0.0024} \\
MC Dropout       
    & 0.6505 $\pm$ 0.0391 
    & 0.0049 $\pm$ 0.0049 
    & \textbf{0.0686 $\pm$ 0.0302} 
    & 0.0062 $\pm$ 0.0090 
    & 0.0177 $\pm$ 0.0126 
    & 0.0193 $\pm$ 0.0014 
    & 0.0440 $\pm$ 0.0062 \\
DDU                        
    & 0.6553 $\pm$ 0.0197
    & \textbf{0.0019 $\pm$ 0.0010} 
    & 0.1171 $\pm$ 0.1069
    & \textbf{0.0016 $\pm$ 0.0008}
    & 0.0106 $\pm$ 0.0062 
    & 0.0197 $\pm$ 0.0004 
    & 0.0419 $\pm$ 0.0007 \\
\rowcolor{gray!15}
ST-LoRA (ours)              
    & 0.6920 $\pm$ 0.0147 
    & 0.0026 $\pm$ 0.0018 
    & 0.0901 $\pm$ 0.0311 
    & 0.0027 $\pm$ 0.0011 
    & \textbf{0.0077 $\pm$ 0.0059} 
    & 0.0186 $\pm$ 0.0004 
    & 0.0518 $\pm$ 0.0044 \\
\midrule
\multicolumn{8}{l}{\textit{Post-hoc calibrators (applied to single model)}} \\
\midrule
Uncalibrated      
    & 0.6692 $\pm$ 0.0289 
    & 0.0049 $\pm$ 0.0026 
    & 0.1794 ± 0.0463 & 0.0058 ± 0.0019 & 0.0369 ± 0.0127 
    & 0.0192 $\pm$ 0.0021 
    & 0.0476 $\pm$ 0.0069 \\
TS                
    & 0.6692 $\pm$ 0.0289 
    & 0.0052 $\pm$ 0.0039 
    &0.1572 ± 0.0581 & 0.0055 ± 0.0023 & 0.0364 ± 0.0174 
    & 0.0193 $\pm$ 0.0023 
    & 0.0467 $\pm$ 0.0077 \\
Logistic          
    & 0.6717 $\pm$ 0.0276 
    & 0.0042 $\pm$ 0.0020 
    & 0.1414 ± 0.0588 & 0.0048 ± 0.0019 & 0.0302 ± 0.0152 
    & 0.0189 $\pm$ 0.0016 
    & 0.0454 $\pm$ 0.0059 \\
Dirichlet         
    & 0.6432 $\pm$ 0.0196 
    & 0.0023 $\pm$ 0.0008 
    & 0.1073 ± 0.0763 & 0.0043 ± 0.0015 & 0.0313 ± 0.0146 
    & 0.0203 $\pm$ 0.0008 
    & 0.0443 $\pm$ 0.0031 \\
LTS               
    & 0.6692 $\pm$ 0.0289 
    & 0.0044 $\pm$ 0.0011 
    &0.1761 ± 0.0707 & 0.0043 ± 0.0010 & 0.0346 ± 0.0111 
    & 0.0185 $\pm$ 0.0010 
    & 0.0410 $\pm$ 0.0022 \\
Meta-Cal          
    & 0.3921 $\pm$ 0.0694 
    & 0.0284 $\pm$ 0.0131 
    & 0.3510 ± 0.1015 & 0.0201 ± 0.0073 & 0.1794 ± 0.0721 
    & 0.0520 $\pm$ 0.0108 
    & 0.1302 $\pm$ 0.0266 \\
Selective         
    & 0.4156 $\pm$ 0.1006 
    & 0.0718 $\pm$ 0.1233 
    & 0.3820 ± 0.2166 & 0.0810 ± 0.1485 & 0.1971 ± 0.2980
    & 0.0967 $\pm$ 0.1309 
    & 0.2328 $\pm$ 0.3098 \\
\bottomrule
\end{tabular}}
\end{table*}

\begin{table}[t]
\caption{\textbf{Computational efficiency comparison across approaches for forward propagation.}}
\label{tbl:computational_complexity}
\centering
\begin{tabular}{l|cc}

\textbf{Approach} & \textbf{Flops} (GMAC) & \textbf{Train Params.} (M) \\
\midrule
M2F (base)                     & 229,72 & 108.77   \\
\midrule
FRE               & 1148            & 534   \\

MC-Dropout & 1148           & 108.77  \\
DDU                            & 320            & 108.8    \\
\rowcolor{gray!15}
ST-LoRA (ours)                   & 1170   & 116  \\
\bottomrule
\end{tabular}
\end{table}

For ensemble methods, five snapshots are collected per training 
trajectory for both FRE and ST-LoRA by saving the model weights 
blindly at the end of each cosine annealing cycle, without any 
validation-based selection. For MC Dropout and DDU, a single 
checkpoint is saved at the end of training. This blind collection 
strategy is applied uniformly across all methods: tracking the best 
validation loss would introduce a selection advantage that is 
incompatible with the snapshot ensembling design, where checkpoints 
are taken at fixed cycle endpoints regardless of their individual 
validation performance. Keeping this consistent ensures that 
observed differences in performance reflect the uncertainty 
estimation strategy rather than checkpoint selection policy.

Finally, to position ST-LoRA within the broader calibration literature, we 
additionally compare it against six established post-hoc calibration 
methods: Temperature Scaling (TS)~\cite{guo2017calibration}, Logistic 
scaling~\cite{platt1999probabilistic}, Dirichlet 
scaling~\cite{kull2019beyond}, Meta-Cal~\cite{ma2021meta}, Local 
Temperature Scaling (LTS) ~\cite{ding2021local}, and the Selective scaling method of \cite{wang2023calibrating} for semantic segmentation 
calibration\footnote{we use the same code from \url{https://github.com/dwang181/selectivecal}}. 
To calibrate these methods, we randomly split the original validation 
data (63 samples) into two subsets: subset A, the new validation set 
(33 samples), and subset B, the calibration set (30 samples).

We additionally report 
Negative Log-Likelihood (NLL) and the Brier 
Score~\cite{brier1950verification} alongside ECE/ACE. As, ECE alone has several known limitations: a low ECE does not necessarily correspond to high accuracy, its binning procedure introduces a bias/variance trade-off, and it considers only the maximum predicted probability 
rather than the full predictive distribution~\cite{pavlovic2025understanding}.\footnote{We additionally observed that ECE is unstable across numerical precisions for semantic segmentation - \textit{float32} and \textit{float64} encodings can yield different results. We use \textit{float32} throughout and leave a systematic investigation of 
this precision sensitivity to future work.}

\subsection{In-Distribution Performance and Robustness Under Distribution Shift}
\label{sec8.1}

We follow the same two-stage evaluation protocol as previous 
experiments: in-distribution performance followed by robustness 
under distribution shift.

\textbf{In-distribution performance.} Table~\ref{tab:id_performance_m2f} shows that ST-LoRA achieves mIoU competitive with the best-performing method (FRE) and remains highly competitive in calibration, ranking among the top three on most metrics. Specifically, on ECE, DDU performs best, followed by Dirichlet scaling, with ST-LoRA ranking third. On Brier score, FRE leads, followed by LTS, with ST-LoRA again placing third. On NLL, however, most methods outperform ST-LoRA, with the exception of Meta-Cal and Selective Scaling. For the remaining calibration metrics, ST-LoRA ranks second on ACE and MACE, while achieving the best performance on MECE.

Among all methods, DDU offers the lowest computational cost and calibration gains (see Table~\ref{tbl:computational_complexity}), as uncertainty is estimated in a single forward pass without sampling. However, this efficiency comes at the cost of segmentation accuracy. The observed mIoU drop for DDU is attributed to the widespread application of spectral normalization, which constrains the feature space and reduces discriminative capacity, consistent with the accuracy/robustness trade-off reported in~\cite{huang2021training, gogianu2021spectral}.

\begin{figure*}[t]
    \centering
    \includegraphics[width=0.90\linewidth]{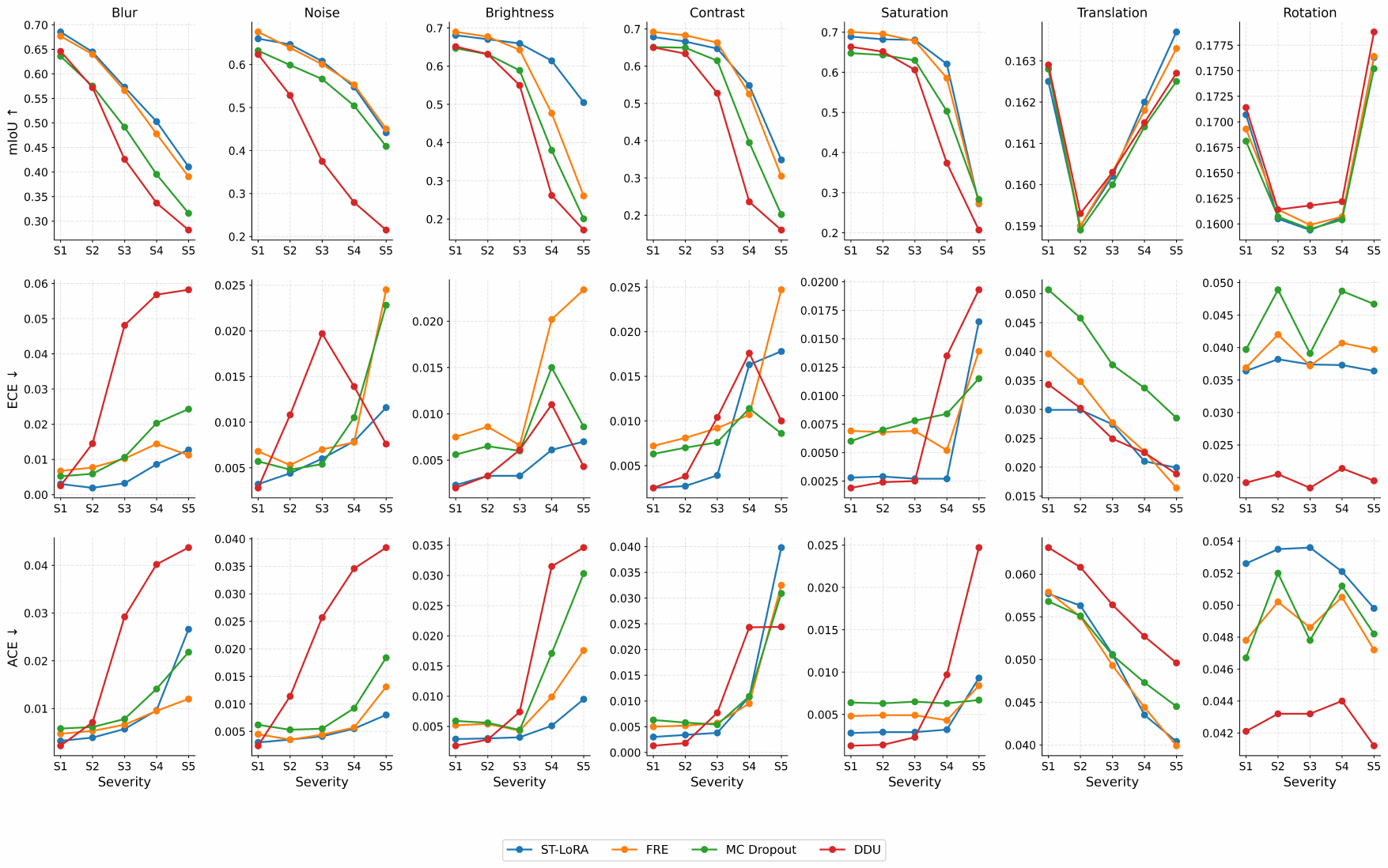}
    \caption{\textbf{Robustness under distribution shift comparison}. ST-LoRA demonstrates the best performance against different kinds of shifts. $\uparrow$: higher is better; $\downarrow$: lower is better. Represented the mean of five different seeds.}
    \label{compare_rob}
\end{figure*}

Furthermore, applying SN may inherently limit ViT performance. As \cite{dong2021attention} demonstrate, MLP layers play a critical role in preserving representational capacity by counteracting the rank collapse induced by self-attention and increases the Lipschitz constant of the network. SN directly opposes this dynamic by constraining the Lipschitz constant to a fixed upper bound, creating a counter force between both. Consequently, how to enforce a meaningful Lipschitz bound in transformers without sacrificing representational capacity remains an open question, with promising directions explored in \cite{newhouse2025training}.
Finally, MC Dropout, performance sensitivity likely reflects the joint effect of the targeted dropout layer and the number of stochastic forward passes in addition to other hyperparameters as mentioned in \cite{JMLR:v15:srivastava14a}.

ST-LoRA exhibits the lowest standard deviation across all metrics among all methods, indicating stable performance regardless 
of the random seed a practically important property for reliable 
deployment, in addition to reducing the need for architectural changes like in the DDU case or the need for more sample like in the case of the calibration methods. Taken together, these results position ST-LoRA as a strong candidate that 
balances calibration quality, segmentation accuracy, and 
computational efficiency - consistently competitive without being 
the top performer on any single metric in isolation. \footnote{Further analysis on the mIoU/calibration per class is demonstrated in the appendix (Table~\ref{tab:per_class_iou}, and Table~\ref{tab:nll_brier_comparison})}
          
However, in the next section we show that 
the results obtained for in-distribution experiment does not necessarily transfer to distribution-shifted settings.

\textbf{Robustness under distribution shift.} We evaluate all methods 
on BUP20 under a systematic corruption grid of seven corruption types 
at five severity levels each, yielding 35 shift conditions. All 
severity levels are deliberately chosen to lie outside the training 
augmentation range, ensuring exposure to unseen conditions . 
Photometric corruptions, applied to the image only, include Gaussian 
blur ($\sigma \in \{1.5, 3, 5, 7, 10\}$\,px, versus training 
$\sigma \in [0.5, 1.5]$\,px), additive Gaussian noise ($\sigma \in 
\{0.04, 0.08, 0.12, 0.18, 0.25\}$, versus training $\sigma = 0.02$), 
brightness reduction (factor $\in \{0.60, 0.40, 0.25, 0.10, 0.05\}$, 
versus training $[0.7, 1.3]$), contrast reduction (same factor 
range), and saturation reduction (factor $\in \{0.80, 0.60, 0.40, 
0.15, 0.00\}$, versus training $[0.9, 1.1]$). Geometric corruptions 
include lateral translation ($\{20, 30, 40, 50, 60\}\%$ of image 
width, versus training $\pm10\%$) and rotation (vertical flip, 
$90^\circ$, $180^\circ$, $270^\circ$, and $90^\circ{+}$vertical 
flip, versus training horizontal flip only).

\begin{figure*}[t]
    \centering
    \includegraphics[width=1.0\linewidth]{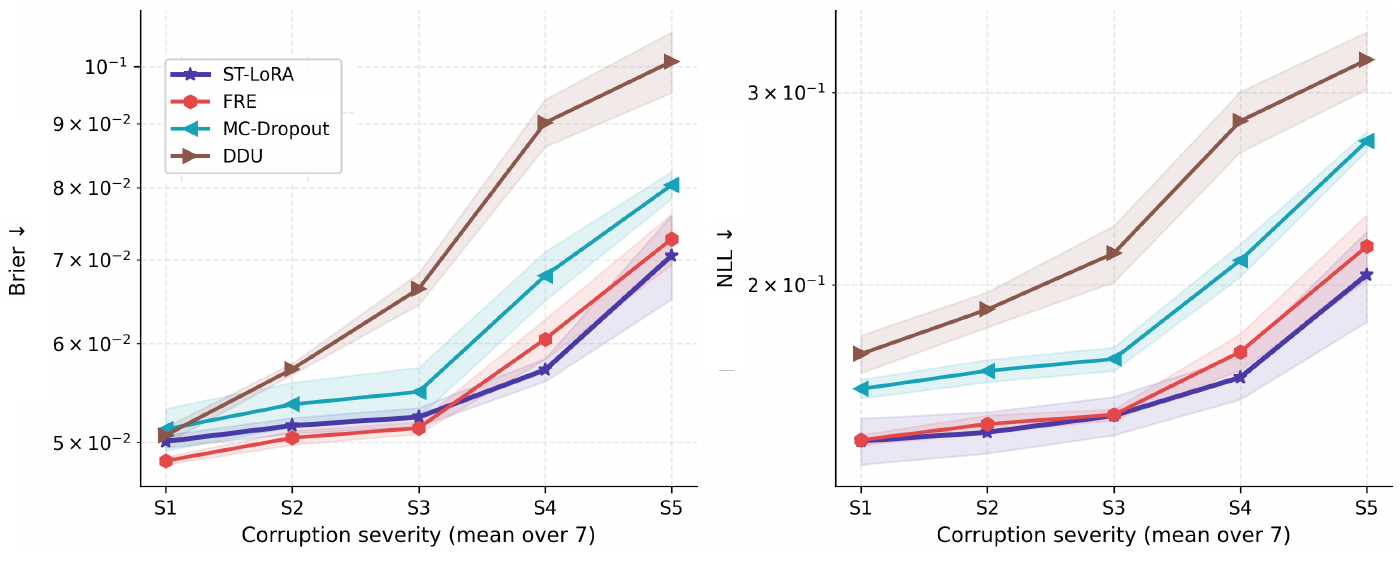}
    \caption{\textbf{Uncertainty methods robustness under covariate shift. We measure the performance at each tick on the x-axis by averaging over the shift types and seeds.}}
    \label{nll_brier_compare}
\end{figure*}

\begin{figure*}[t]
    \centering
    \includegraphics[width=1.0\linewidth]{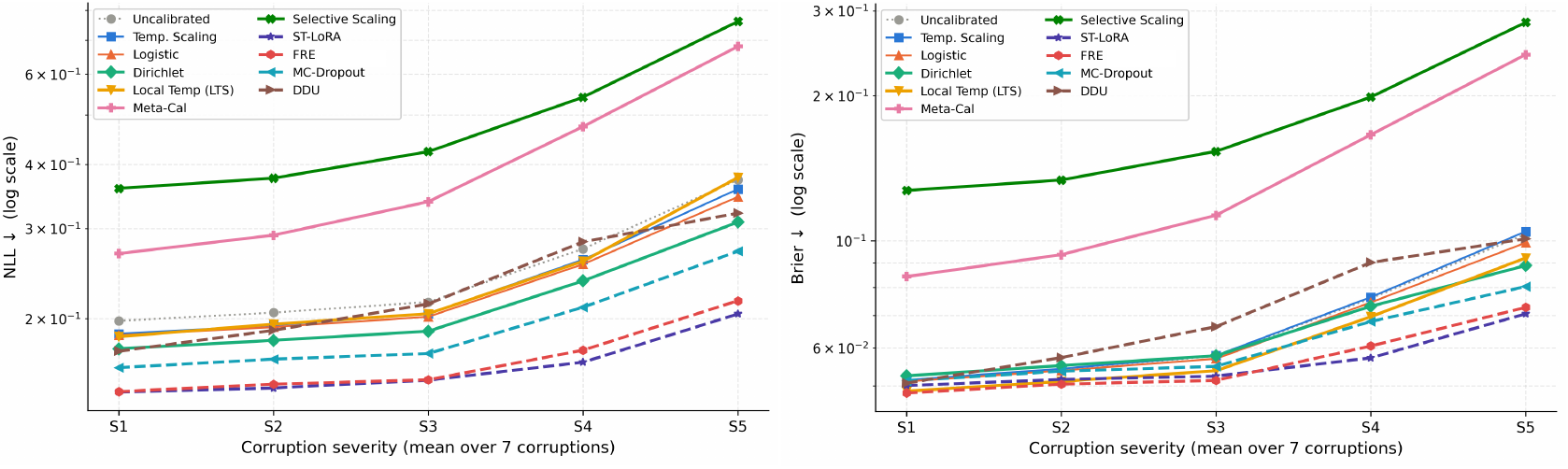}
    \caption{\textbf{Uncertainty methods and post hoc calibrators robustness under covariate shift. We measure the performance at each tick on the x-axis by averaging over the shift types and seeds.}}
    \label{nll_brier_all}
\end{figure*}

As shown in Figure~\ref{compare_rob}, across nearly all shift types and severity levels, ST-LoRA achieves the strongest calibration and segmentation robustness among all methods. FRE follows as the second most robust, ahead of MC Dropout, while DDU degrades most severely under distribution shift. Furthermore, We compare the four methods across the seven corruption types introduced in Experiment 3, reporting performance 
averaged across corruption type and severity level, with mean and 
standard deviation over five seeds. As shown in 
Figure~\ref{nll_brier_compare}, ST-LoRA maintains strong calibration 
robustness under distribution shift, consistent with our earlier 
findings in Figure~\ref{compare_rob}. At low shift severity, FRE 
performs slightly better; however, ST-LoRA becomes increasingly 
competitive as shift severity increases, ultimately matching or 
exceeding FRE under the most severe corruptions. Finally, we extend this comparison to include the post-hoc calibration baselines, providing a complete picture of how all 
evaluated methods degrade under shift. Figure~\ref{nll_brier_all} demonstrates that the Bayesian  approaches (MC Dropout, FRE, ST-LoRA) generally outperform post-hoc calibration 
methods under distribution shift

Although DDU is the most computationally efficient method at inference time, carries the smallest memory footprint, and nearly best calibrated at in-distribution analysis, its calibration collapses rapidly under shift which is a well-documented limitation of deterministic uncertainty methods, not exclusive to DDU \cite{postels2022practicality}. The poor robustness of MC-Dropout is consistent with our earlier hypothesis that both the choice of dropout layer and the sampling rate critically affect performance under shift, and needs further investigation to identify configurations that generalize more reliably beyond the training distribution. For post-hoc calibration methods our findings align with \cite{ovadia2019can}, who show that strong in-distribution calibration does not guarantee calibration or robustness under distributional shift. Consistent with our earlier observation, FRE and ST-LoRA remain the two best-performing methods overall: FRE is stronger at low shift severity, while ST-LoRA outperforms all methods - including FRE - under higher shift severity.

\subsection{Out of Distribution (OoD) detection}

\begin{figure*}[t]
    \centering
    \includegraphics[width=1.0\linewidth]{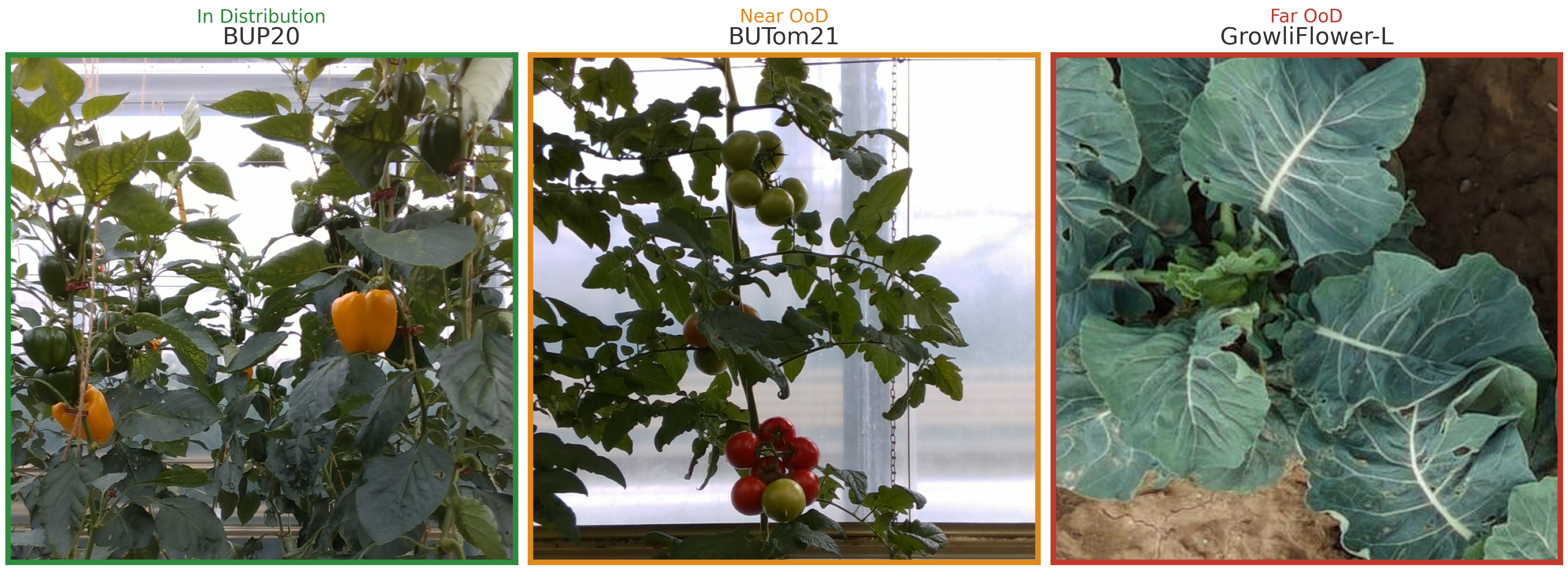}
    \caption{\textbf{The in-distribution → out-of-distribution spectrum used in our evaluation}. From left to right: an in-distribution sample from BUP20 (glasshouse sweet pepper, the training domain); a near-OoD sample from BUTom21 (glasshouse tomato — same acquisition setup and imaging conditions, but a novel crop with different fruit/foliage appearance); and a far-OoD sample from GrowliFlower-L (open-field cauliflower — a different crop, scene layout, background, and illumination). Border colour encodes the shift severity (green → amber → red). This spectrum lets us probe model calibration and pixel-level anomaly detection under increasing semantic and covariate shift.}
    \label{id_near_far_ood}
\end{figure*}

We evaluate our uncertainty quantification methods under two 
distribution-shift regimes: a far-OoD condition and a near-OoD 
condition, illustrated in Figure~\ref{id_near_far_ood}.

\textbf{Far-OoD.} We use GrowliFlower-L, which represents markedly 
different capture conditions - open-field, UAV-acquired imagery- relative to BUP20, the dataset on which our models are trained. From BUP20 we use 93 images (the full test set), while from GrowliFlower-L we use 1{,}500 samples from its validation set, after 
removing soil-only images corresponding to the early growth stage, all cauliflower plant's pixels are grouped into a single foreground (anomalous) class, with the background assigned as the normal class.

\textbf{Near-OoD.} We select the \textit{BUTom21} tomato 
dataset~\cite{halstead2026still}\footnote{\url{https://bonndata.uni-bonn.de/dataset.xhtml?persistentId=doi:10.60507/FK2/DHTEH1}} 
as a well-suited candidate, captured in the University of Bonn's commercial-style glasshouse at \textit{Campus Klein-Altendorf} (CKA). Although the glasshouse hosts multiple crops across separate cells, 
we use only the \textit{Sweeterno} tomato cell, which contains no other fruit or vegetable species. This dataset forms a natural near-OoD counterpart to BUP20, sharing the same glasshouse environment and camera setup, with tomato classes that are visually similar in color to the sweet pepper classes in BUP20. We evaluate on the dataset's validation split, comprising approximately 4{,}500 images at $1280 \times 720$ resolution and spanning five ripeness classes: green, orange, mixed-orange, mixed-red, and red. For our purposes, all tomato classes are grouped into a single foreground (anomalous) class, with the background assigned as the normal class.

\textbf{Image-level OoD detection.} We assess each method's 
ability to separate ID and OoD samples at the image level, without 
exploiting local spatial structure. Uncertainty scores are computed 
per pixel and aggregated by mean across each image. For ensemble and 
MC Dropout methods, we use predictive entropy and mutual information \cite{Shaker2021} as scoring functions; for DDU, we use the GMM log-density\footnote{More details on each method is available at Table~\ref{tab:method_configs}.}. Finally, we evaluate ID/OoD separability qualitatively via score histograms and quantitatively via the Area Under the Receiver Operating Characteristic curve (AUROC) \cite{hendrycks2016baseline} and False Positive Rate@95$\%$ (FPR@95).

\textbf{Pixel-level anomaly segmentation} We are the first to apply such evaluation scheme - following the protocol of \cite{chan2021segmentmeifyoucan,sodano2024arxiv} - to close-range glasshouse crop segmentation, assessing the sensitivity of uncertainty estimation methods to cross-crop OoD objects at the pixel level. Pixel-level anomaly segmentation using \textit{ranking} metrics( Area Under the Precision-Recall Curve (AUPR), False Positive Rate at $95\%$ True Positive Rate (FPR95)), and \textit{localization} metrics (segment-wise Intersection over Union (sIoU), Positive Predictive Value (PPV), and mean F1 score (MeanF1)) has not, to our knowledge, been evaluated in this setting.

Table~\ref{tab:ood_results} summarizes OoD detection performance across all methods. On the image level, ST-LoRA achieves the best AUROC/FPR@95 between ID and OoD images on both datasets; followed by FRE, MC Dropout, and DDU. Critically, ST-LoRA also exhibits the lowest standard deviation across both scoring functions among all methods, confirming that its OoD sensitivity is stable and seed-independent, a practically important property for deployment.

Across all bayesian approximation methods, mutual information consistently underperforms predictive entropy as an OoD scoring function on GrowliFlower-L and vice versa on the BUTom21 dataset except for MC-Dropout approach where the observation is the opposite \footnote{More insights are in the appendix}. This is not a failure of the ensemble approach: \cite{Huellermeier2022} have shown that very low MI values reflect a fundamental limitation of the measure itself under OoD conditions. For the mismatch of how mutual information behaves on different\footnote{Further analysis are in Appendix~\ref{appen_a}} datasets, \cite{Huellermeier2022} mention that the additivity of uncertainties in DL models remain subtle. This questions the assumption itself and remains a research hotspot. Furthremore, \cite{fellaji2024epistemic} explores the \textit{epistemic uncertainty hole} which is an epistemic uncertainty collapse phenomenon that occurs when we train on small dataset or having large models. DDU shows the highest standard deviation across both of its scoring functions, making it the least reliable method for OoD detection on Mask2Former, consistent with the architectural incompatibility concerns discussed in the previous experiment \footnote{Robustness under covariate shift experiment.}.

\begin{figure*}[t]
    \centering
    \begin{subfigure}{\linewidth}
        \centering
        \includegraphics[width=0.65\linewidth, height=0.28\textheight, keepaspectratio]{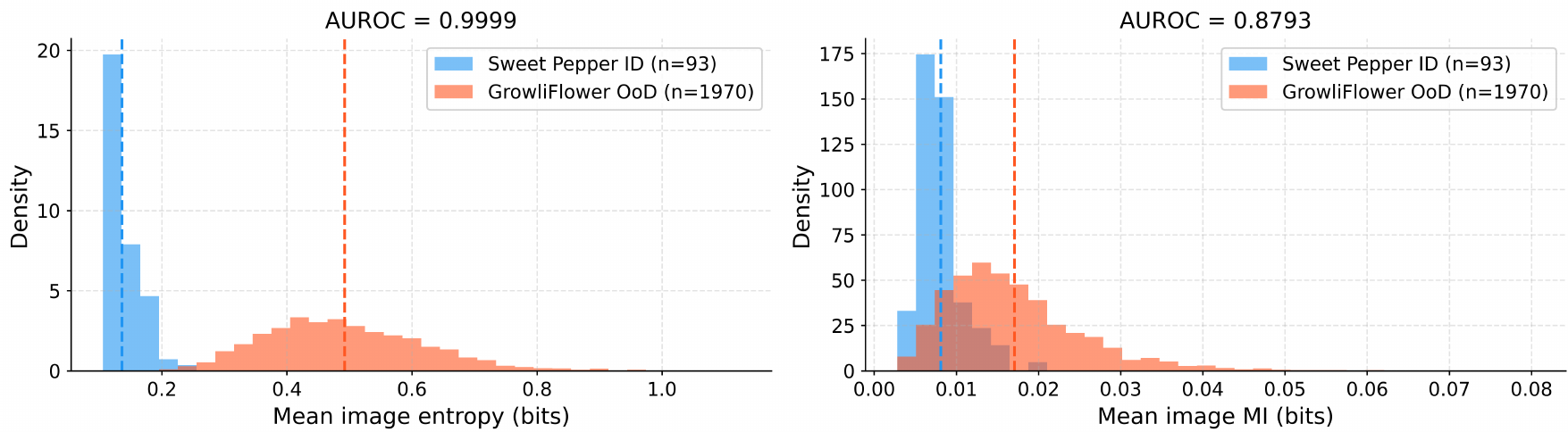}
        \caption{ST-LoRA (\textit{seed 42})}
        \label{fig:hist_lora}
    \end{subfigure}
    \begin{subfigure}{\linewidth}
        \centering
        \includegraphics[width=0.65\linewidth, height=0.28\textheight, keepaspectratio]{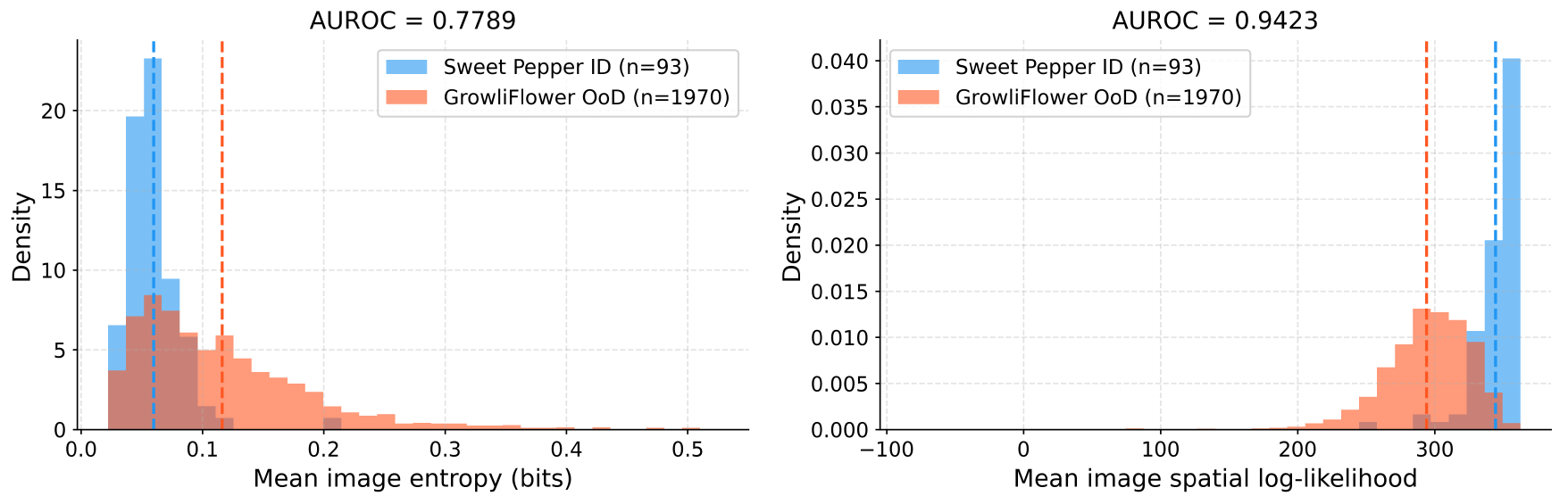}
        \caption{DDU (\textit{seed 123})}
        \label{fig:hist_ddu}
    \end{subfigure}
    \begin{subfigure}{\linewidth}
        \centering
        \includegraphics[width=0.65\linewidth, height=0.28\textheight, keepaspectratio]{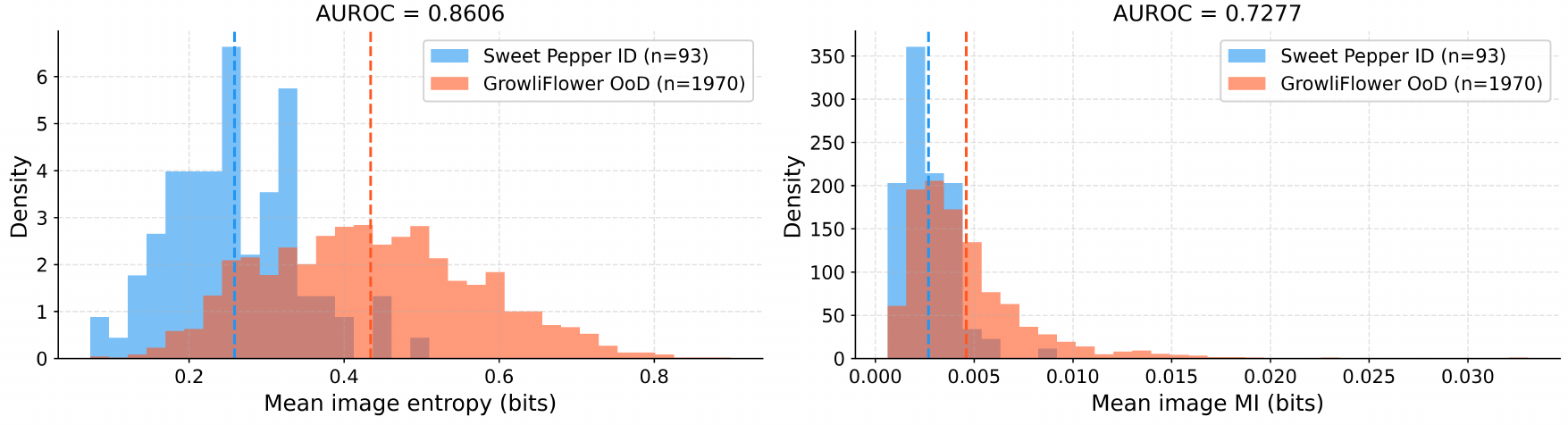}
        \caption{MC Dropout (\textit{seed 456})}
        \label{fig:hist_mcdropout}
    \end{subfigure}
    \begin{subfigure}{\linewidth}
        \centering
        \includegraphics[width=0.65\linewidth, height=0.28\textheight, keepaspectratio]{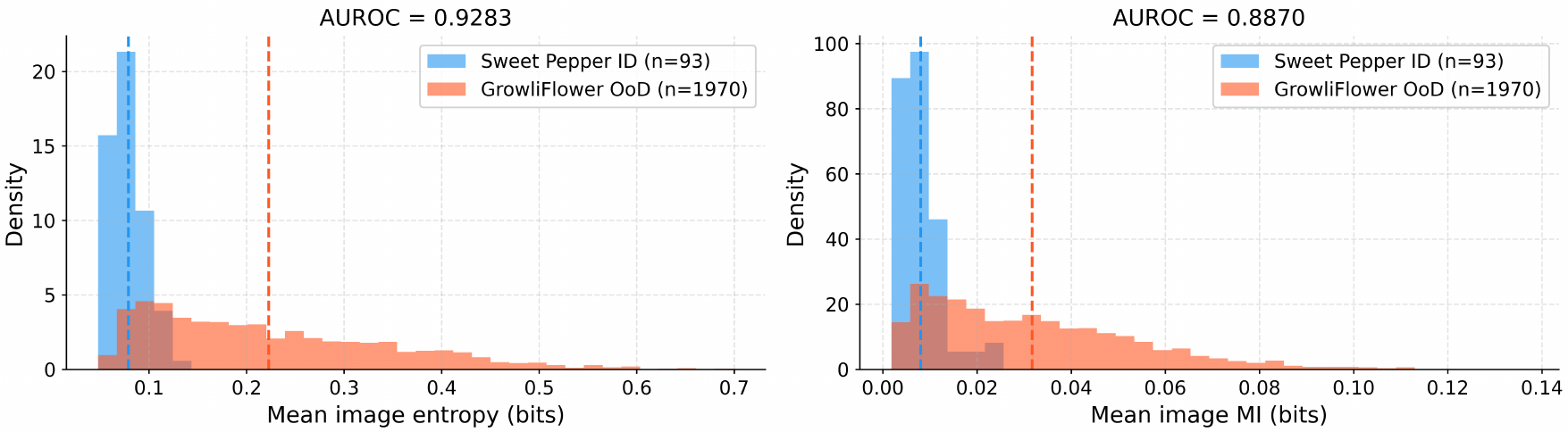}
        \caption{FRE (\textit{seed 1337})}
        \label{fig:hist_fullft}
    \end{subfigure}
    \caption{\textbf{Image-level OoD score distributions across methods.} 
    Each panel shows the density of predictive entropy (left) and mutual 
    information/Density (right), dashed vertical lines indicate the mean of each distribution. We decide to present the best performance for each method according to only one seed.}
    \label{fig:auroc_comparison}
\end{figure*}

Figure~\ref{fig:auroc_comparison} shows that ST-LoRA achieves the clearest separation between ID and OoD distributions under entropy scoring, exhibiting the largest gap between distribution means. Across all methods, OoD samples consistently produce \textit{higher} entropy/lower density than ID samples; the expected and desirable behaviour, confirming that uncertainty estimates are well-aligned with the novelty of the input.

Among the baselines, DDU achieves strong separation using GMM density scores but is highly unstable across seeds, making it unreliable in practice. MC Dropout collapses a large fraction of OoD samples into the ID score range a known failure mode under far OoD conditions~\cite{Lakshminarayanan2017}, where the method 
produces overconfident, low-variance predictions on inputs far from the training distribution. FRE ranks second overall, offering reliable but less consistent separation than ST-LoRA while performing better on mutual information.

\begin{table*}[t]
\centering
\caption{\textbf{OoD detection results for image-level and pixel-level 
anomaly segmentation across two shift regimes}: 
\textit{Far}-OoD (ID: Sweet Pepper (BUP20); OoD: 
GrowliFlower-L) and \textit{Near}-OoD (ID: Sweet 
Pepper (BUP20); OoD: Tomato (BUTom21)). Image-level scores are 
aggregated per image via mean pooling. Pixel-level metrics follow 
the \textsc{SegmentMeIfYouCan} protocol~\cite{chan2021segmentmeifyoucan}. 
Scorer: Ent.\ = predictive entropy; MI = mutual information; 
Den = Density. $\uparrow$: higher is better; $\downarrow$: lower is 
better. All results report mean\,$\pm$\,std over 5 seeds. Best 
result per metric in \textbf{bold}.}
\label{tab:ood_results}
\resizebox{\textwidth}{!}{
\begin{tabular}{ll cc cccccc}
\toprule
& & \multicolumn{2}{c}{\textbf{Image-level}} 
  & \multicolumn{6}{c}{\textbf{Pixel-level}} \\
\cmidrule(lr){3-4} \cmidrule(lr){5-10}
\textbf{Method} & \textbf{Scorer} 
    & \textbf{AUROC} $\uparrow$
    & \textbf{FPR95} $\downarrow$
    & \textbf{AUROC} $\uparrow$ 
    & \textbf{AUPR} $\uparrow$ 
    & \textbf{FPR95} $\downarrow$ 
    & \textbf{sIoU} $\uparrow$ 
    & \textbf{PPV} $\uparrow$ 
    & \textbf{MeanF1} $\uparrow$ \\

\midrule
\multicolumn{10}{c}{\cellcolor{red!8}\textit{\textbf{Far-OoD (GrowliFlower-L)}}} \\
\midrule

\multirow{2}{*}{FRE}
    & Ent. & 0.9282\,$\pm$\,0.0497 & 0.4129±0.2723 & 0.5828\,$\pm$\,0.0896 & 0.9021\,$\pm$\,0.0266 & 0.9185\,$\pm$\,0.0436 & 0.4121\,$\pm$\,0.0184 & 0.1986\,$\pm$\,0.0848 & 0.2074\,$\pm$\,0.0856 \\
    & MI   & 0.7886\,$\pm$\,0.1241 & 0.7505±0.1989 & 0.5845\,$\pm$\,0.0767 & 0.9016\,$\pm$\,0.0242 & 0.9256\,$\pm$\,0.0370 & \textbf{0.4133\,$\pm$\,0.0225} & 0.1820\,$\pm$\,0.0718 & 0.1927\,$\pm$\,0.0734 \\
\midrule
\multirow{2}{*}{MC Dropout}        
    & Ent. & 0.5874\,$\pm$\,0.1549 &  0.9161±0.1579 & \textbf{0.5901\,$\pm$\,0.0606} & \textbf{0.9058\,$\pm$\,0.0165} & \textbf{0.9094\,$\pm$\,0.0451} & 0.4029\,$\pm$\,0.0250 & 0.2364\,$\pm$\,0.0894 & 0.2425\,$\pm$\,0.0844 \\
    & MI   & 0.6914\,$\pm$\,0.0486 & 0.8323±0.1052 & 0.5274\,$\pm$\,0.0238 & 0.8830\,$\pm$\,0.0073 & 0.9436\,$\pm$\,0.0189 & 0.4121\,$\pm$\,0.0124 & 0.2322\,$\pm$\,0.0462 & 0.2455\,$\pm$\,0.0432 \\
\midrule
\multirow{1}{*}{DDU}     
    & Den. & 0.3860\,$\pm$\,0.3335 & 0.8220 ± 0.3930 & 0.5475\,$\pm$\,0.0458 & 0.8910\,$\pm$\,0.0155 & 0.9426\,$\pm$\,0.0190 & 0.3405\,$\pm$\,0.0192 & 0.0036\,$\pm$\,0.0001 & 0.0065\,$\pm$\,0.0008 \\
\midrule
\multirow{2}{*}{ST-LoRA (ours)} 
    & \cellcolor{gray!15}Ent. & \cellcolor{gray!15}\textbf{0.9981\,$\pm$\,0.0025} & \cellcolor{gray!15}\textbf{0.0065±0.0096} & \cellcolor{gray!15}0.4069\,$\pm$\,0.1043 & \cellcolor{gray!15}0.8309\,$\pm$\,0.0357 & \cellcolor{gray!15}0.9719\,$\pm$\,0.0230 & \cellcolor{gray!15}0.4022\,$\pm$\,0.0338 & \cellcolor{gray!15}\textbf{0.3013\,$\pm$\,0.0792} & \cellcolor{gray!15}\textbf{0.3088\,$\pm$\,0.0765} \\
    & \cellcolor{gray!15}MI   & \cellcolor{gray!15}0.5877\,$\pm$\,0.2110 & \cellcolor{gray!15}0.8903±0.1001 & \cellcolor{gray!15}0.4044\,$\pm$\,0.0862 & \cellcolor{gray!15}0.8329\,$\pm$\,0.0278 & \cellcolor{gray!15}0.9794\,$\pm$\,0.0127 & \cellcolor{gray!15}0.3938\,$\pm$\,0.0325 & \cellcolor{gray!15}\textbf{0.2746\,$\pm$\,0.0727} & \cellcolor{gray!15}\textbf{0.2845\,$\pm$\,0.0707} \\

\midrule
\multicolumn{10}{c}{\cellcolor{orange!8}\textit{\textbf{Near-OoD (BUTom21)}}} \\
\midrule
\multirow{2}{*}{FRE} 
    & Ent. & 0.8340±0.1381 & 0.5441±0.1821& 0.9734\,$\pm$\,0.0053 & 0.5826\,$\pm$\,0.0310 & 0.0675\,$\pm$\,0.0165 & \textbf{0.5745\,$\pm$\,0.0096} & 0.2514\,$\pm$\,0.0117 & 0.3157\,$\pm$\,0.0119 \\
    & MI   & 0.8884±0.1424 & 0.3183±0.2546 & 0.9709\,$\pm$\,0.0065 & \textbf{0.6179\,$\pm$\,0.0575} & 0.0801\,$\pm$\,0.0244 & 0.5715\,$\pm$\,0.0097 & 0.1591\,$\pm$\,0.0173 & 0.2312\,$\pm$\,0.0199 \\
\midrule
\multirow{2}{*}{MC Dropout} 
    & Ent. & 0.7177±0.1050 & 0.8258±0.1111 & 0.9474\,$\pm$\,0.0185 & 0.4525\,$\pm$\,0.0929 & 0.2201\,$\pm$\,0.1774 & 0.5349\,$\pm$\,0.0364 & 0.1823\,$\pm$\,0.0596 & 0.2420\,$\pm$\,0.0681 \\
    & MI   & 0.5777±0.0443 & 0.9054±0.0439 & 0.8309\,$\pm$\,0.1032 & 0.2950\,$\pm$\,0.1155 & 0.6763\,$\pm$\,0.1729 & 0.5116\,$\pm$\,0.0675 & 0.1167\,$\pm$\,0.0317 & 0.1504\,$\pm$\,0.0446 \\
\midrule
DDU & Den. & 0.8660 ± 0.025 & 0.5660 ± 0.1660 & 0.9365\,$\pm$\,0.0343 & 0.5073\,$\pm$\,0.0609 & 0.3020\,$\pm$\,0.2299 & 0.5484\,$\pm$\,0.0088 & 0.1354\,$\pm$\,0.0146 & 0.1861\,$\pm$\,0.0127 \\
\midrule
\multirow{2}{*}{ST-LoRA (ours)} 
    & \cellcolor{gray!15}Ent. & \cellcolor{gray!15}0.8919±0.0871 & \cellcolor{gray!15}0.3871±0.1647 & \cellcolor{gray!15}\textbf{0.9752\,$\pm$\,0.0040} & \cellcolor{gray!15}0.5629\,$\pm$\,0.0374 & \cellcolor{gray!15}\textbf{0.0445\,$\pm$\,0.0015} & \cellcolor{gray!15}0.5693\,$\pm$\,0.0055 & \cellcolor{gray!15}\textbf{0.2946\,$\pm$\,0.0154} & \cellcolor{gray!15}\textbf{0.3568\,$\pm$\,0.0111} \\
    & \cellcolor{gray!15}MI   & \cellcolor{gray!15}\textbf{0.9594±0.0304} & \cellcolor{gray!15}\textbf{0.2043±0.1498} & \cellcolor{gray!15}0.9737\,$\pm$\,0.0045 & \cellcolor{gray!15}0.5995\,$\pm$\,0.0707 & \cellcolor{gray!15}0.0486\,$\pm$\,0.0032 & \cellcolor{gray!15}0.5704\,$\pm$\,0.0154 & \cellcolor{gray!15}0.2039\,$\pm$\,0.0205 & \cellcolor{gray!15}0.2780\,$\pm$\,0.0200 \\

\bottomrule
\end{tabular}}
\end{table*}

At the pixel level on the GrowliFlower-L dataset, no single winning approach emerges, as most methods perform close to the random model (AUROC=0.5) on several metrics. Specifically, on three out of six metrics, ST-LoRA performs the worst, while MC-Dropout achieves the best performance in terms of AUROC, AUPR, and FPR@95 compared to all other approaches. On sIOU, ST-LoRA ranks as the third-best approach; however, for PPV and MeanF1, ST-LoRA achieves the best performance. DDU achieves near-zero PPV and MeanF1, confirming that its density-based scores do not translate into reliable pixel-level anomaly localization in this setting.

For BUTom21, ST-LoRA outperforms FRE on four out of six metrics. Specifically, ST-LoRA achieves slightly better AUCROC, while ranking second after FRE on AUPR. Our approach performs best in terms of FPR@95. For sIoU, FRE takes the lead, albeit by a narrow margin. Finally, ST-LoRA delivers the best performance on both PPV and MeanF1. Overall, ST-LoRA provides the most consistent and reliable OoD detection profile: the best image-level separation, the best precision and F1 at the pixel level, and the lowest variance across seeds; making it the most dependable choice for uncertainty-aware deployment in agricultural environments where OoD inputs must be identified reliably.

\subsection{Statistical Significance and Results Discussion}

To draw more rigorous conclusions, we performed paired significance testing across all evaluation protocols used in this study. As shown in Table~\ref{tab:full_significance}, ST-LoRA emerges as a strong competitor to FRE, underscoring the favorable efficiency-performance trade-off it offers.

On in-domain performance, segmentation quality (mIoU) shows no significant difference between the two methods, while FRE achieves significantly better calibration. Under distribution shift (covariate shift only, in our setting), mIoU, Brier score, and NLL again show no significant difference, whereas FRE achieves significantly better ECE and ACE.

For image-level OoD detection, ST-LoRA proves highly competitive: performance is statistically tied on BUTom21 (near-OoD), while ST-LoRA significantly outperforms FRE on GrowliFlower-L (far-OoD). At the pixel level, ST-LoRA significantly outperforms FRE on three metrics for the near-OoD BUTom21 dataset, with no significant difference on the remaining metrics. On the far-OoD GrowliFlower-L dataset, however, this pattern reverses: FRE significantly outperforms ST-LoRA on three metrics, with no significant difference on the rest.

Taken together, these results demonstrate that ST-LoRA offers a compelling trade-off between efficiency and performance. It reduces training time compared to deep ensemble, I/O inference latency, and checkpoint size -and therefore storage requirements compared to FRE- while matching FRE on several evaluation protocols and outperforming it in specific scenarios, most notably far-OoD image-level detection and near-OoD pixel-level localization.

\begin{table*}[t]
\centering
\caption{\textbf{Paired significance comparison across OoD detection, 
calibration, and distribution shift, FRE vs.\ ST-LoRA for Mask2Former.} $d = 
\text{ST-LoRA}-\text{FRE}$; $\uparrow$: higher better, 
$\downarrow$: lower better (for a $\downarrow$-metric, negative $d$ 
favours ST-LoRA). Paired, $n{=}5$ seeds (42, 123, 456, 789, 1337). 
95\% CI via 20k-resample bootstrap. $p$ columns report paired-$t$ / 
Wilcoxon (two-sided; Wilcoxon floors at $0.0625$ at $n{=}5$). Significant iff $p{<}0.05$ \textbf{and} 95\% CI 
excludes 0. Scorer: entropy only shown, \colorbox{sigGreener!55}{ST-LoRA significantly 
better}\,; \colorbox{tieColor}{no significant difference}\,; 
\colorbox{sigWorse!55}{ST-LoRA significantly worse}.}
\label{tab:full_significance}
\setlength{\tabcolsep}{3pt}
\resizebox{0.90\textwidth}{!}{
\begin{tabular}{llcccccc}
\toprule
\textbf{Setting} & \textbf{Metric} & \textbf{FRE} & \textbf{ST-LoRA} & \textbf{$d$} & \textbf{$p$ ($t$/$W$)} & \textbf{$d_z$} & \textbf{95\% CI} \\
\midrule
\multicolumn{8}{c}{\cellcolor{gray!8}\textit{In-domain (clean)}} \\
\midrule
Clean & mIoU $\uparrow$  & 0.7023\,$\pm$\,0.0182 & \sigTie{0.6942\,$\pm$\,0.0176} & $-0.0081$ & 0.227/0.312 & $-0.64$ & $[-0.018,+0.002]$ \\
Clean & ECE $\downarrow$   & 0.0068\,$\pm$\,0.0035 & \sigUp{0.0187\,$\pm$\,0.0046} & $+0.0120$ & 0.019/0.062 & $+1.70$ & $[+0.007,+0.017]$ \\
Clean & ACE $\downarrow$   & 0.0067\,$\pm$\,0.0035 & \sigUp{0.0187\,$\pm$\,0.0047} & $+0.0120$ & 0.019/0.062 & $+1.70$ & $[+0.007,+0.017]$ \\
Clean & Brier $\downarrow$ & 0.0171\,$\pm$\,0.0004 & \sigUp{0.0186\,$\pm$\,0.0004} & $+0.0014$ & 0.003/0.062 & $+2.87$ & $[+0.001,+0.002]$ \\
Clean & NLL $\downarrow$   & 0.0389\,$\pm$\,0.0027 & \sigUp{0.0518\,$\pm$\,0.0050} & $+0.0130$ & 0.013/0.062 & $+1.92$ & $[+0.008,+0.018]$ \\
\midrule
\multicolumn{8}{c}{\cellcolor{gray!8}\textit{Distribution shift (overall mean across shifts, 35 conditions)}} \\
\midrule
Overall & mIoU $\uparrow$  & 0.4558\,$\pm$\,0.0090 & \sigTie{0.4641\,$\pm$\,0.0096} & $+0.0083$ & 0.291/0.312 & $+0.54$ & $[-0.003,+0.019]$ \\
Overall & ECE $\downarrow$   & 0.0196\,$\pm$\,0.0029 & \sigUp{0.0295\,$\pm$\,0.0050} & $+0.0099$ & 0.015/0.062 & $+1.85$ & $[+0.006,+0.015]$ \\
Overall & ACE $\downarrow$   & 0.0200\,$\pm$\,0.0023 & \sigUp{0.0295\,$\pm$\,0.0051} & $+0.0095$ & 0.014/0.062 & $+1.87$ & $[+0.006,+0.014]$ \\
Overall & Brier $\downarrow$ & 0.0567\,$\pm$\,0.0014 & \sigTie{0.0563\,$\pm$\,0.0017} & $-0.0003$ & 0.774/0.625 & $-0.14$ & $[-0.002,+0.002]$ \\
Overall & NLL $\downarrow$   & 0.1671\,$\pm$\,0.0050 & \sigTie{0.1621\,$\pm$\,0.0086} & $-0.0050$ & 0.404/0.438 & $-0.42$ & $[-0.013,+0.005]$ \\
\midrule
\multicolumn{8}{c}{\cellcolor{gray!8}\textit{Image-level OoD (entropy)}} \\
\midrule
BUTom21 (near) & AUROC $\uparrow$ & 0.8340\,$\pm$\,0.1381 & \sigTie{0.8919\,$\pm$\,0.0871} & $+0.0579$ & 0.111/0.125 & $+0.91$ & $[+0.013,+0.108]$ \\
BUTom21 (near) & AUPR $\uparrow$  & 0.9945\,$\pm$\,0.0064 & \sigTie{0.9965\,$\pm$\,0.0038} & $+0.0020$ & 0.188/0.188 & $+0.71$ & $[+0.000,+0.005]$ \\
BUTom21 (near) & FPR95 $\downarrow$ & 0.5441\,$\pm$\,0.1821 & \sigTie{0.3871\,$\pm$\,0.1647} & $-0.1570$ & 0.055/0.125 & $-1.20$ & $[-0.256,-0.056]$ \\
GrowliFlower-L (far) & AUROC $\uparrow$ & 0.9283\,$\pm$\,0.0555 & \sigDown{0.9981\,$\pm$\,0.0025} & $+0.0699$ & 0.049/0.062 & $+1.25$ & $[+0.030,+0.119]$ \\
GrowliFlower-L (far) & AUPR $\uparrow$  & 0.9962\,$\pm$\,0.0032 & \sigTie{0.9999\,$\pm$\,0.0001} & $+0.0037$ & 0.060/0.062 & $+1.17$ & $[+0.002,+0.007]$ \\
GrowliFlower-L (far) & FPR95 $\downarrow$ & 0.4129\,$\pm$\,0.2723 & \sigDown{0.0065\,$\pm$\,0.0096} & $-0.4065$ & 0.030/0.062 & $-1.48$ & $[-0.607,-0.170]$ \\
\midrule
\multicolumn{8}{c}{\cellcolor{gray!8}\textit{Pixel-level OoD (entropy)}} \\
\midrule
BUTom21 (near) & AUROC $\uparrow$ & 0.9734\,$\pm$\,0.0053 & \sigTie{0.9752\,$\pm$\,0.0040} & $+0.0019$ & 0.640/0.438 & $+0.23$ & $[-0.005,+0.009]$ \\
BUTom21 (near) & AUPR $\uparrow$  & 0.5826\,$\pm$\,0.0310 & \sigTie{0.5629\,$\pm$\,0.0374} & $-0.0197$ & 0.190/0.312 & $-0.70$ & $[-0.041,+0.005]$ \\
BUTom21 (near) & FPR95 $\downarrow$ & 0.0675\,$\pm$\,0.0165 & \sigDown{0.0445\,$\pm$\,0.0015} & $-0.0231$ & 0.038/0.062 & $-1.36$ & $[-0.038,-0.011]$ \\
BUTom21 (near) & sIoU $\uparrow$   & 0.5745\,$\pm$\,0.0096 & \sigTie{0.5693\,$\pm$\,0.0055} & $-0.0053$ & 0.336/0.625 & $-0.49$ & $[-0.014,+0.003]$ \\
BUTom21 (near) & PPV $\uparrow$    & 0.2514\,$\pm$\,0.0117 & \sigDown{0.2946\,$\pm$\,0.0154} & $+0.0432$ & 0.007/0.062 & $+2.27$ & $[+0.029,+0.058]$ \\
BUTom21 (near) & MeanF1 $\uparrow$ & 0.3157\,$\pm$\,0.0119 & \sigDown{0.3568\,$\pm$\,0.0111} & $+0.0411$ & 0.007/0.062 & $+2.24$ & $[+0.028,+0.055]$ \\
GrowliFlower-L (far) & AUROC $\uparrow$ & 0.6881\,$\pm$\,0.1250 & \sigUp{0.4335\,$\pm$\,0.1051} & $-0.2546$ & 0.037/0.062 & $-1.38$ & $[-0.420,-0.147]$ \\
GrowliFlower-L (far) & AUPR $\uparrow$  & 0.4148\,$\pm$\,0.1425 & \sigUp{0.1603\,$\pm$\,0.0269} & $-0.2545$ & 0.017/0.062 & $-1.77$ & $[-0.383,-0.152]$ \\
GrowliFlower-L (far) & FPR95 $\downarrow$ & 0.8567\,$\pm$\,0.1384 & \sigTie{0.9702\,$\pm$\,0.0261} & $+0.1135$ & 0.174/0.062 & $+0.74$ & $[+0.019,+0.251]$ \\
GrowliFlower-L (far) & sIoU $\uparrow$   & 0.4263\,$\pm$\,0.1333 & \sigUp{0.1889\,$\pm$\,0.0093} & $-0.2374$ & 0.016/0.062 & $-1.78$ & $[-0.342,-0.122]$ \\
GrowliFlower-L (far) & PPV $\uparrow$    & 0.0641\,$\pm$\,0.0533 & \sigTie{0.0722\,$\pm$\,0.0370} & $+0.0081$ & 0.736/1.000 & $+0.16$ & $[-0.030,+0.046]$ \\
GrowliFlower-L (far) & MeanF1 $\uparrow$ & 0.0920\,$\pm$\,0.0634 & \sigTie{0.0696\,$\pm$\,0.0301} & $-0.0224$ & 0.460/0.812 & $-0.36$ & $[-0.072,+0.024]$ \\

\bottomrule
\end{tabular}}
\end{table*}

\section{Final Remarks}
\label{fremark}
\subsection{Comparison with Related Work}
ST-LoRA establishes a strong, scalable ensembling framework that achieves a compelling balance between computational efficiency and 
uncertainty quantification quality, reducing training time compared to deep ensemble, inference I/O 
latency, trainable parameter count, and memory footprint simultaneously while 
remaining competitive compared to established SOTA methods across all evaluated tasks.

Table~\ref{tab:comparison} situates our work within the existing 
LoRA ensemble literature across four dimensions: uncertainty 
evaluation schemes, ablation scope, target components, and 
application domain. The closest related work is 
\cite{halbheer2024lora}\footnote{Concurrent with our submission.}, 
which is the only prior study to evaluate both distribution shift and 
OoD detection. However, their analysis stops short of investigating 
how LoRA hyperparameters interact with covariate shift, a gap our 
work directly addresses. They also do not report whether ensemble 
members were trained with augmentation, nor examine its effect on 
robustness, whereas we show that heterogeneous augmentation is 
critical for calibration stability under shift.

On the ablation side, \cite{halbheer2024lora} focus exclusively on 
rank and target component selection, reporting that FF layer 
adaptation improves accuracy but degrades calibration, a 
conclusion our results contradict, and which a growing body of 
evidence from both text and vision literature supports in our 
favour~\cite{hetowards,zhangadaptive,melas2021you}. Critically, 
they overlook the scaling factor $\delta_r = \nicefrac{\alpha}{r}$ 
entirely, which we identify as a key driver of the 
accuracy/calibration trade-off and could be further investigated in future work as a training-free deployment-time calibration knob. Their focus on attention-only adaptation - using up to 16 ensemble members - also makes FF adaptation appear costly, whereas in our setting of $M{=}4$ members, FF layers provide the strongest performance at the lowest parameter count.

Beyond \cite{halbheer2024lora}, remaining related works are either 
restricted to specific vision tasks such as depth estimation, or 
operate exclusively in the text domain, with no ablation studies and 
no evaluation under distribution shift. None address dense prediction, 
which imposes substantially different representational demands than 
image classification.

ST-LoRA is therefore the first LoRA ensemble framework to jointly 
address parameter efficiency, calibration robustness under 
distribution shift, near/far OoD detection, and dense agricultural prediction, providing both empirical evidence and principled guidance for 
practitioners deploying uncertainty-aware models in resource-
constrained real-world environments.

\subsection{Limitations of ST-LoRA and Future Work}

While ST-LoRA has demonstrated strong performance, several 
limitations and open directions remain.


\begin{enumerate}

    \item \textbf{Hyperparameter sensitivity.} ST-LoRA inherits the large hyperparameter search space characteristic of LoRA-based methods, making configuration time-consuming; a limitation shared by most uncertainty quantification approaches. 
    
    \noindent \textbf{Promising directions:} Broader studies across datasets, tasks, and architectures will help identify which  hyperparameters matter most and under what conditions. Neural Architecture Search (NAS)~\cite{white2023neural} is a natural complement, enabling principled adapter placement within the pre-trained encoder rather than searching for architectures from scratch.

    \item \textbf{Miscalibration.} Although ST-LoRA improves calibration over strong baselines, residual 
    miscalibration persists; a challenge shared across all uncertainty quantification methods evaluated.
    
    \noindent \textbf{Promising directions:} Post-hoc calibration methods~\cite{Wang2022OnCS} are a low-cost remedy that leave the trained model intact. More specifically, we propose for future work training with $\alpha{=}r$ and subsequently fine-tuning $\alpha$ as a post-hoc calibration knob could simultaneously reduce training complexity and improve deployment-time calibration.

    \item \textbf{Data, task, and model dependency.} LoRA adaptation remains sensitive to the specific model, dataset, and task,  meaning conclusions from one setting cannot be assumed to  transfer directly to another; a known challenge across the LoRA literature.
    
    \noindent \textbf{Promising directions:} Extending ablation studies to account for hyperparameter interactions across more diverse tasks would yield more transferable guidelines. Additionally, theoretical analysis of LoRA in the vision domain~\cite{xu2025understanding, kim2025lora} remains significantly underdeveloped relative to the text domain and represents an important open research direction.

    \item \textbf{Deployment efficiency.} Despite its parameter efficiency, The frozen backbone continues to dominate the overall computation, while the adapters introduce additional FLOPs and sometimes training speed reduction. Furthermore, maintaining multiple adapter sets in memory remains a practical bottleneck, particularly on memory-constrained edge devices.
    
    \noindent \textbf{Promising directions:} Post-training pruning~\cite{cheng2024survey} of LoRA modules~\cite{zhang2024loraprune} 
    can reduce adapter size further, while quantization~\cite{gholami2022survey} - as demonstrated by QLoRA~\cite{dettmers2023qlora} - 
    can substantially reduce memory footprint for edge deployment.



\end{enumerate}


\subsection{Ethical Implications}
The computational efficiency of ST-LoRA carries implications that 
extend beyond benchmark performance. By restricting fine-tuning to 
less model parameters and eliminating the need for 
$M$ independent training runs, ST-LoRA  reduces GPU hours compared to the deep ensemble required to train and deploy uncertainty-aware segmentation models. This reduction in compute demand has a direct environmental 
consequence: lower energy consumption and, by extension, lower CO$_2$ emissions and water usage during both training and inference. As large-scale 
deep learning increasingly contributes to the carbon footprint of the technology sector~\cite{fernandez2025energy}, parameter-efficient methods such as ST-LoRA represent a step toward more environmentally responsible machine learning practice.

We believe that the research community has a responsibility to account for these downstream 
consequences when designing and reporting computational methods, and that efficiency is not only a technical virtue but an ethical one. In addition, these efficiency gains align with broader global sustainability objectives, particularly the United Nations Sustainable Development Goals (SDGs), including SDG 2: Zero Hunger. Climate change remains a critical factor affecting agricultural productivity, land usability, and food security worldwide. By 
reducing energy consumption and associated emissions, more efficient machine learning methods may contribute, in aggregate, to mitigating climate-related risks that threaten food systems. While the impact of any single methodological improvement is necessarily limited, such efforts collectively support a transition toward more sustainable technological practices that are better aligned with long-term environmental stability and global food security.

\begin{table*}[t]
\centering
\caption{\textbf{Qualitative comparison of ST-LoRA-based uncertainty quantification 
methods}. \cmark~indicates the property is present, \xmark~indicates it 
is absent.}
\label{tab:comparison}
\setlength{\tabcolsep}{4pt}
\small
\begin{tabular}{l|cc|cc|cc|cc}
\toprule
& \multicolumn{2}{c|}{\textbf{Uncertainty Evaluation}}
& \multicolumn{2}{c|}{\textbf{Ablation Study}}
& \multicolumn{2}{c|}{\textbf{Component}}
& \multicolumn{2}{c}{\textbf{Domain}} \\
\textbf{Method}
& OoD 
& \makecell{Dist.\\ Shift} 
& \makecell{Rank\\ Only}
& \makecell{Layer\\ Selection}
& \makecell{Attn.\\ Only}
& \makecell{FF\\ Layers}
& Generic
& Agri. \\
\midrule
\cite{halbheer2024lora}     
& \cmark & \xmark  & \cmark & \xmark & \cmark & \xmark & \cmark & \xmark \\
\cite{Balabanov2024UncertaintyQI} 
& \xmark & \xmark  & \xmark & \xmark & \cmark & \xmark & \cmark & \xmark \\
\cite{paul2024parameter}       
& \xmark & \xmark  & \xmark & \xmark & \cmark & \xmark & \cmark & \xmark \\
\cite{onal2024gaussian}            
& \xmark & \xmark  & \xmark & \xmark & \cmark & \xmark & \cmark & \xmark \\
\midrule
\rowcolor{gray!15}
\textbf{ST-LoRA (Ours)}
& \cmark & \cmark  & \cmark & \cmark & \cmark & \cmark & \xmark & \cmark \\

\bottomrule
\end{tabular}
\end{table*}

\section{Conclusion}
\label{conc}

We introduced ST-LoRA, a parameter-efficient ensemble framework for uncertainty-aware semantic segmentation that combines Low-Rank Adaptation with snapshot ensemble. By sharing a frozen backbone across ensemble members and restricting each member to lightweight low-rank adapters, ST-LoRA reduces training time by $\nicefrac{1}{M}$, trainable parameters, and checkpoint size; while demonstrating competitive performance compared to full-rank ensembles in segmentation quality, calibration, robustness under covariate shift, and far/near OoD performance.

Two findings stand out from our ablation study with one that has been proved empirically with support from literature while the second is left for a future work. We have shown that feed-forward layers are the critical LoRA target for dense prediction, consistently outperforming attention-only adaptation in both accuracy and calibration - contrary to the convention inherited from language model fine-tuning. 

For future work, The scaling ratio $\delta_r = \nicefrac{\alpha}{r}$ governs the accuracy/calibration trade-off under distribution shift and could serve as a training-free deployment-time calibration knob, allowing practitioners to adjust the model's uncertainty behaviour after fine-tuning without retraining. This property has no equivalent in full-rank fine-tuning.

Across two crops, three architectures, and three evaluation schemes, ST-LoRA emerges as a competitive approach compared to Snapshot Ensemble, MC Dropout, and DDU in calibration stability under distribution shift, while achieving strong performance in image- and pixel-level near/far out-of-distribution detection. Its lowest cross-seed variance among all methods confirms robust, seed-independent performance.

More broadly, the efficiency gains of ST-LoRA reduce the computational and environmental cost of deploying uncertainty-aware models in agriculture; a domain where the sustainability of the technology should align with the sustainability goals it serves.

\section*{Acknowledgements}

This work has partially been funded by the Deutsche Forschungsgemeinschaft (DFG, German Research Foundation) under Germany’s Excellence Strategy, EXC-2070 - 390732324 - PhenoRob. Further, it has been funded by the Deutsche Forschungsgemeinschaft (DFG, German Research Foundation) - RO 4839/6-1 - 459376902 - AID4Crops. The authors gratefully acknowledge the access to the \textit{Marvin} cluster of the University of Bonn.







\printcredits

\bibliographystyle{cas-model2-names}

\bibliography{cas-refs}

\appendix

\section{Complementary Analysis}
\label{appen_a}

\subsection{BUP20 Dataset Per Class Analysis}

We evaluate per-class mIoU and calibration 
performance\footnote{For overall performance (Table~\ref{tab:id_performance_m2f}), 
we average over the total number of pixels; for the per-class 
analysis (Tables~\ref{tab:per_class_iou} and \ref{tab:nll_brier_comparison}), we 
first compute per-class performance and then average over the 
number of classes.} to determine whether ST-LoRA's improvements are 
concentrated in majority classes - background and green peppers 
- or distributed consistently across all classes. As shown in 
Table~\ref{tab:per_class_iou}, FRE achieves the best overall per-class 
performance, followed by ST-LoRA, while MC Dropout and DDU degrade 
performance relative to the uncalibrated model. Most post-hoc 
calibration methods leave per-class mIoU largely unchanged, as 
calibration is expected to preserve the predicted class while 
smoothing the output distribution; Dirichlet, Meta-Cal, and 
selective scaling are exceptions, each degrading performance.

For per-class calibration (Table~\ref{tab:nll_brier_comparison}), FRE again achieves 
the best performance on both NLL and Brier score, followed by 
ST-LoRA, with LTS ranking third and showing strong in-distribution 
calibration gains. These results confirm that ST-LoRA remains 
highly competitive with both FRE and the post-hoc calibration 
methods at the per-class level, indicating that its improvements are 
not limited to majority classes.

\begin{table*}[t]
\centering
\caption{\textbf{Per-class IoU and macro-mIoU comparison across methods for Mask2Former model on BUP20 dataset (test split) (mean\,$\pm$\,std over 5 seeds)}. 
$\uparrow$: higher is better. Best result per metric in \textbf{bold}.}
\label{tab:per_class_iou}
\resizebox{0.80\textwidth}{!}{
\begin{tabular}{lccccccc}
\toprule
\textbf{Method} & \textbf{bg} $\uparrow$ & \textbf{pepper red} $\uparrow$ & \textbf{pepper yellow} $\uparrow$ & \textbf{pepper green} $\uparrow$ & \textbf{mixed red} $\uparrow$ & \textbf{mixed yellow} $\uparrow$ & \textbf{macro-mIoU} $\uparrow$ \\
\midrule
\multicolumn{8}{l}{\textit{Uncertainty estimation methods}} \\
\midrule
DDU             & 0.990 & 0.605 & 0.494 & 0.750 & 0.550 & 0.544 & 0.655\,$\pm$\,0.018 \\
MC-Dropout      & 0.990 & 0.641 & 0.418 & 0.765 & 0.582 & 0.516 & 0.652\,$\pm$\,0.035 \\
FRE   & \textbf{0.991} & \textbf{0.672} & \textbf{0.545} & 0.774 & \textbf{0.611} & 0.620 & \textbf{0.702\,$\pm$\,0.016} \\
\rowcolor{gray!15} ST-LoRA (ours) & \textbf{0.991} & 0.645 & 0.536 & 0.769 & 0.602 & \textbf{0.627} & 0.695\,$\pm$\,0.015 \\
\midrule
\multicolumn{8}{l}{\textit{Post-hoc calibrators (applied to single model)}} \\
\midrule
Uncalibrated    & \textbf{0.991} & 0.617 & 0.499 & \textbf{0.780} & 0.545 & 0.583 & 0.669\,$\pm$\,0.029 \\
TS              & \textbf{0.991} & 0.617 & 0.499 & \textbf{0.780} & 0.545 & 0.583 & 0.669\,$\pm$\,0.029 \\
Logistic        & \textbf{0.991} & 0.623 & 0.498 & \textbf{0.780} & 0.551 & 0.588 & 0.672\,$\pm$\,0.028 \\
Dirichlet       & 0.989 & 0.509 & 0.503 & 0.760 & 0.510 & 0.588 & 0.643\,$\pm$\,0.020 \\
LTS             & \textbf{0.991} & 0.617 & 0.499 & \textbf{0.780} & 0.545 & 0.583 & 0.669\,$\pm$\,0.029 \\
Meta-Cal        & 0.977 & 0.155 & 0.130 & 0.567 & 0.276 & 0.247 & 0.392\,$\pm$\,0.069 \\
Selective       & 0.984 & 0.196 & 0.243 & 0.744 & 0.227 & 0.306 & 0.450\,$\pm$\,0.126 \\
\bottomrule
\end{tabular}}
\end{table*}

\begin{figure*}[t]
    \centering
    \includegraphics[width=0.85\linewidth]{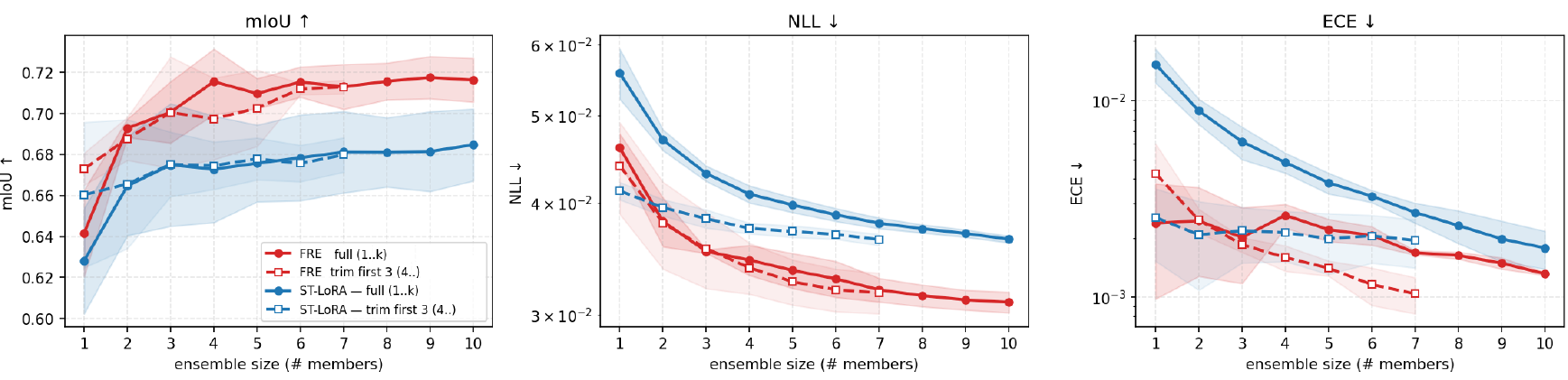}
    \caption{\textbf{Number of ensemble members ablation study for FRE Vs. ST-LoRA.} The performance is reported as mean and standard deviation of three seeds.}
    \label{ensemble_ablation}
\end{figure*}

\begin{table*}[t]
\centering
\caption{\textbf{Per-class NLL and Brier score comparison across methods for Mask2Former model on BUP20 dataset (test split) (mean\,$\pm$\,std over 5 seeds)}. 
$\downarrow$: lower is better. Overall best result per metric across all methods in \textbf{bold}.}
\label{tab:nll_brier_comparison}
\resizebox{0.80\textwidth}{!}{
\begin{tabular}{l|ccccccc|ccccccc}
\toprule
 & \multicolumn{7}{c|}{\textbf{Per-Class NLL} $\downarrow$} & \multicolumn{7}{c}{\textbf{Per-Class Brier} $\downarrow$} \\
\textbf{Method} & \textbf{bg} & \textbf{red} & \textbf{yel} & \textbf{grn} & \textbf{m-red} & \textbf{m-yel} & \textbf{macro} & \textbf{bg} & \textbf{red} & \textbf{yel} & \textbf{grn} & \textbf{m-red} & \textbf{m-yel} & \textbf{macro} \\
\midrule
\multicolumn{15}{l}{\textit{Uncertainty estimation methods}} \\
\midrule
DDU           & 0.013 & 0.877 & 0.644 & 0.537 & 1.494 & 1.002 & 0.761\,$\pm$\,0.053 & 0.007 & 0.407 & \textbf{0.357} & 0.238 & 0.560 & 0.471 & 0.340\,$\pm$\,0.015 \\
MC-Dropout    & 0.018 & 0.474 & 0.870 & 0.491 & 1.209 & 0.968 & 0.672\,$\pm$\,0.024 & 0.008 & \textbf{0.220} & 0.474 & 0.210 & 0.536 & 0.489 & 0.323\,$\pm$\,0.007 \\
FRE  & 0.016 & \textbf{0.467} & \textbf{0.661} & 0.481 & \textbf{0.873} & \textbf{0.744} & \textbf{0.540\,$\pm$\,0.019} & \textbf{0.006} & 0.225 & 0.381 & 0.227 & \textbf{0.459} & \textbf{0.395} & \textbf{0.282\,$\pm$\,0.011} \\
\rowcolor{gray!15} ST-LoRA (ours) & 0.030 & 0.583 & 0.709 & 0.458 & 0.943 & 0.874 & 0.600\,$\pm$\,0.015 & 0.007 & 0.272 & 0.379 & 0.220 & 0.495 & 0.465 & 0.306\,$\pm$\,0.009 \\
\midrule
\multicolumn{15}{l}{\textit{Post-hoc calibrators (applied to single model)}} \\
\midrule
Uncalibrated  & 0.014 & 0.528 & 1.078 & 0.525 & 1.983 & 1.548 & 0.946\,$\pm$\,0.132 & 0.007 & 0.241 & 0.494 & 0.212 & 0.572 & 0.483 & 0.335\,$\pm$\,0.043 \\
TS            & 0.015 & 0.521 & 1.041 & 0.501 & 1.866 & 1.464 & 0.901\,$\pm$\,0.120 & 0.007 & 0.241 & 0.491 & 0.211 & 0.567 & 0.479 & 0.333\,$\pm$\,0.042 \\
Logistic      & 0.014 & 0.585 & 1.105 & 0.485 & 1.835 & 1.448 & 0.912\,$\pm$\,0.118 & 0.007 & 0.264 & 0.503 & 0.206 & 0.559 & 0.470 & 0.335\,$\pm$\,0.045 \\
Dirichlet     & 0.018 & 1.388 & 1.131 & \textbf{0.387} & 1.007 & 1.206 & 0.856\,$\pm$\,0.076 & 0.010 & 0.639 & 0.502 & \textbf{0.165} & 0.385 & 0.430 & 0.355\,$\pm$\,0.040 \\
LTS           & \textbf{0.013} & 0.654 & 0.969 & 0.498 & 1.339 & 1.079 & 0.759\,$\pm$\,0.053 & \textbf{0.006} & 0.285 & 0.464 & 0.219 & 0.528 & 0.458 & 0.327\,$\pm$\,0.035 \\
Meta-Cal      & 0.075 & 1.824 & 1.973 & 1.070 & 2.057 & 2.060 & 1.510\,$\pm$\,0.143 & 0.031 & 0.748 & 0.817 & 0.416 & 0.727 & 0.729 & 0.578\,$\pm$\,0.046 \\
Selective     & 0.499 & 1.107 & 2.085 & 0.812 & 2.148 & 2.087 & 1.456\,$\pm$\,0.368 & 0.210 & 0.482 & 0.896 & 0.344 & 0.756 & 0.750 & 0.573\,$\pm$\,0.159 \\
\bottomrule
\end{tabular}}
\end{table*}

\begin{table*}[t]
\centering
\caption{\textbf{Member‑diversity metrics for Mask2Former under three training regimes (BUP20 test set, M=4 shots, 5 seeds, mean\,$\pm$\,std).
$\Delta = \text{ST‑LoRA} - \text{FRE}$.
``Members'' are within‑seed snapshots at identical epochs.}}
\label{tab:div_all_conditions}

\par\noindent\makebox[\textwidth]{\textbf{Constant LR, no--augmentation}}%
\medskip

\resizebox{0.85\textwidth}{!}{%
\begin{tabular}{l ccc ccc ccc}
\toprule
\multirow{2}{*}{\textbf{Metric}} & \multicolumn{3}{c}{\textbf{Foreground (gt>0)}} & \multicolumn{3}{c}{\textbf{All valid pixels}} & \multicolumn{3}{c}{\textbf{Macro (mean over classes)}} \\
\cmidrule(lr){2-4} \cmidrule(lr){5-7} \cmidrule(lr){8-10}
 & \textbf{FRE} & \textbf{ST‑LoRA} & \textbf{$\Delta$} & \textbf{FRE} & \textbf{ST‑LoRA} & \textbf{$\Delta$} & \textbf{FRE} & \textbf{ST‑LoRA} & \textbf{$\Delta$} \\
\midrule
Pairwise \textit{argmax} Disag. & $0.1176 \pm 0.0072$ & $0.0677 \pm 0.0088$ & $-0.0499$ & $0.0080 \pm 0.0005$ & $0.0049 \pm 0.0005$ & $-0.0031$ & $0.1467 \pm 0.0142$ & $0.0887 \pm 0.0147$ & $-0.0579$ \\
KL Symm.   & $0.3755 \pm 0.0307$ & $0.1308 \pm 0.0178$ & $-0.2447$ & $0.0245 \pm 0.0020$ & $0.0113 \pm 0.0014$ & $-0.0132$ & $0.4670 \pm 0.0654$ & $0.1988 \pm 0.0286$ & $-0.2682$ \\
Mutual Info.   & $0.0857 \pm 0.0077$ & $0.0370 \pm 0.0045$ & $-0.0487$ & $0.0058 \pm 0.0005$ & $0.0033 \pm 0.0004$ & $-0.0024$ & $0.1087 \pm 0.0126$ & $0.0560 \pm 0.0071$ & $-0.0527$ \\
Fraction of Pixels with Disag. & $0.2117 \pm 0.0145$ & $0.1225 \pm 0.0163$ & $-0.0892$ & $0.0144 \pm 0.0010$ & $0.0088 \pm 0.0009$ & $-0.0056$ & $0.2607 \pm 0.0256$ & $0.1614 \pm 0.0276$ & $-0.0993$ \\
Expected Entropy & $0.2894 \pm 0.0367$ & $0.2675 \pm 0.0111$ & $-0.0218$ & $0.0263 \pm 0.0028$ & $0.0378 \pm 0.0054$ & $+0.0115$ & $0.3235 \pm 0.0435$ & $0.2930 \pm 0.0090$ & $-0.0305$ \\
Shannon Entropy & $0.3751 \pm 0.0403$ & $0.3046 \pm 0.0135$ & $-0.0705$ & $0.0320 \pm 0.0030$ & $0.0411 \pm 0.0056$ & $+0.0091$ & $0.4322 \pm 0.0443$ & $0.3490 \pm 0.0103$ & $-0.0831$ \\
\bottomrule
\end{tabular}}

\vspace{2ex}

\par\noindent\makebox[\textwidth]{\textbf{Cosine LR, no--augmentation}}%
\medskip

\resizebox{0.85\textwidth}{!}{%
\begin{tabular}{l ccc ccc ccc}
\toprule
\multirow{2}{*}{\textbf{Metric}} & \multicolumn{3}{c}{\textbf{Foreground (gt>0)}} & \multicolumn{3}{c}{\textbf{All valid pixels}} & \multicolumn{3}{c}{\textbf{Macro (mean over classes)}} \\
\cmidrule(lr){2-4} \cmidrule(lr){5-7} \cmidrule(lr){8-10}
 & \textbf{FRE} & \textbf{ST‑LoRA} & \textbf{$\Delta$} & \textbf{FRE} & \textbf{ST‑LoRA} & \textbf{$\Delta$} & \textbf{FRE} & \textbf{ST‑LoRA} & \textbf{$\Delta$} \\
\midrule
Pairwise \textit{argmax} Disag. & $0.0889 \pm 0.0024$ & $0.0685 \pm 0.0047$ & $-0.0204$ & $0.0063 \pm 0.0001$ & $0.0050 \pm 0.0002$ & $-0.0013$ & $0.1059 \pm 0.0032$ & $0.0881 \pm 0.0131$ & $-0.0178$ \\
KL Symm. & $0.2598 \pm 0.0079$ & $0.1774 \pm 0.0277$ & $-0.0824$ & $0.0184 \pm 0.0007$ & $0.0175 \pm 0.0012$ & $-0.0009$ & $0.3501 \pm 0.0175$ & $0.2799 \pm 0.0558$ & $-0.0702$ \\
Mutual Info.   & $0.0561 \pm 0.0012$ & $0.0479 \pm 0.0072$ & $-0.0082$ & $0.0041 \pm 0.0001$ & $0.0050 \pm 0.0003$ & $+0.0009$ & $0.0745 \pm 0.0018$ & $0.0757 \pm 0.0143$ & $+0.0011$ \\
Fraction of Pixels with Disag. & $0.1611 \pm 0.0047$ & $0.1242 \pm 0.0079$ & $-0.0369$ & $0.0115 \pm 0.0001$ & $0.0091 \pm 0.0003$ & $-0.0023$ & $0.1901 \pm 0.0079$ & $0.1598 \pm 0.0230$ & $-0.0302$ \\
Expected Entropy & $0.1744 \pm 0.0101$ & $0.2633 \pm 0.0181$ & $+0.0889$ & $0.0192 \pm 0.0006$ & $0.0424 \pm 0.0031$ & $+0.0232$ & $0.1847 \pm 0.0172$ & $0.3197 \pm 0.0272$ & $+0.1350$ \\
Shannon Entropy & $0.2304 \pm 0.0093$ & $0.3111 \pm 0.0243$ & $+0.0807$ & $0.0233 \pm 0.0006$ & $0.0474 \pm 0.0031$ & $+0.0241$ & $0.2592 \pm 0.0166$ & $0.3954 \pm 0.0383$ & $+0.1361$ \\
\bottomrule
\end{tabular}}

\vspace{2ex}

\par\noindent\makebox[\textwidth]{\textbf{Cosine LR, with augmentation }}%
\medskip

\resizebox{0.85\textwidth}{!}{%
\begin{tabular}{l ccc ccc ccc}
\toprule
\multirow{2}{*}{\textbf{Metric}} & \multicolumn{3}{c}{\textbf{Foreground (gt>0)}} & \multicolumn{3}{c}{\textbf{All valid pixels}} & \multicolumn{3}{c}{\textbf{Macro (mean over classes)}} \\
\cmidrule(lr){2-4} \cmidrule(lr){5-7} \cmidrule(lr){8-10}
 & \textbf{FRE} & \textbf{ST‑LoRA} & \textbf{$\Delta$} & \textbf{FRE} & \textbf{ST‑LoRA} & \textbf{$\Delta$} & \textbf{FRE} & \textbf{ST‑LoRA} & \textbf{$\Delta$} \\
\midrule
Pairwise \textit{argmax} Disag. & $0.1355 \pm 0.0171$ & $0.1021 \pm 0.0059$ & $-0.0334$ & $0.0091 \pm 0.0009$ & $0.0076 \pm 0.0006$ & $-0.0015$ & $0.1876 \pm 0.0400$ & $0.1394 \pm 0.0130$ & $-0.0482$ \\
KL Symm. & $0.2557 \pm 0.0532$ & $0.2699 \pm 0.0921$ & $+0.0143$ & $0.0264 \pm 0.0109$ & $0.0238 \pm 0.0048$ & $-0.0025$ & $0.2846 \pm 0.0400$ & $0.3398 \pm 0.0916$ & $+0.0552$ \\
Mutual Info. & $0.0744 \pm 0.0144$ & $0.0791 \pm 0.0238$ & $+0.0048$ & $0.0077 \pm 0.0030$ & $0.0072 \pm 0.0014$ & $-0.0005$ & $0.0860 \pm 0.0119$ & $0.1020 \pm 0.0239$ & $+0.0159$ \\
Fraction of Pixels with Disag. & $0.2444 \pm 0.0330$ & $0.1845 \pm 0.0094$ & $-0.0599$ & $0.0164 \pm 0.0018$ & $0.0139 \pm 0.0012$ & $-0.0025$ & $0.3350 \pm 0.0742$ & $0.2520 \pm 0.0234$ & $-0.0830$ \\
Expected Entropy & $0.4400 \pm 0.0892$ & $0.4463 \pm 0.0290$ & $+0.0062$ & $0.0763 \pm 0.0160$ & $0.0690 \pm 0.0095$ & $-0.0073$ & $0.5243 \pm 0.0637$ & $0.5069 \pm 0.0312$ & $-0.0174$ \\
Shannon Entropy & $0.5144 \pm 0.1024$ & $0.5254 \pm 0.0520$ & $+0.0110$ & $0.0840 \pm 0.0187$ & $0.0762 \pm 0.0104$ & $-0.0078$ & $0.6103 \pm 0.0565$ & $0.6089 \pm 0.0490$ & $-0.0015$ \\
\bottomrule
\end{tabular}}

\end{table*}

\begin{table*}[t]
\centering
\caption{\textbf{Per‑class mutual information for Mask2Former}.
Mean\,$\pm$\,std over 5 seeds (M=4 shots 1‑4, BUP20 test set).
$\Delta = \text{ST‑LoRA} - \text{FRE}$.}
\label{tab:mutual_info_per_class}
\resizebox{0.65\textwidth}{!}{%
\begin{tabular}{l c c c}
\toprule
\textbf{Class} & \textbf{FRE} & \textbf{ST‑LoRA} & \textbf{$\Delta$} \\
\midrule
Background & $0.0047 \pm 0.0026$ & $0.0040 \pm 0.0009$ & $-0.0007$ \\

Pepper red & $0.0928 \pm 0.0345$ & $0.1272 \pm 0.0286$ & $+0.0344$ \\
Pepper yellow & $0.1111 \pm 0.0373$ & $0.1230 \pm 0.0431$ & $+0.0119$ \\
Pepper green & $0.0574 \pm 0.0274$ & $0.0530 \pm 0.0233$ & $-0.0044$ \\

Pepper mixed-red & $0.1462 \pm 0.0250$ & $0.1633 \pm 0.0419$ & $+0.0171$ \\
Pepper mixed-yellow & $0.1039 \pm 0.0251$ & $0.1413 \pm 0.0390$ & $+0.0373$ \\
\bottomrule
\end{tabular}}
\end{table*}

\subsection{Number of Ensemble Members}

We ablate across the number of ensemble members for FRE and ST-LoRA to understand how the number of members affect the performance of both approaches. We keep the same checkpoint (shots) sampling as it is with a shot every 20 epochs and we increase the number of the training epochs to 200. All of the training settings remain the same \footnote{We use three seeds to report mean and standard deviation instead of five seeds due to the requirements of the computational resources.}. As shown in Figure~\ref{ensemble_ablation}, with increasing the number of the ensemble members the performance improves for both approaches with a gap between FRE and ST-LoRA. However, with increasing the number of members beyond 5 we notice that on mIoU the change in performance is minimal while the calibration metrics keep improving with a linear trend.

Furthermore, in this paper we trim the first member as it represents the training region where the model has not converged or in other words the member at the first cycle is underfitting. Hence we want to check the effect of trimming the first three cycles/members from the ensemble on the performance. For mIoU, the performance of both methods degrades while using only the last seven shots, however we can see the effect is more obvious in the case of FRE compared to ST-LoRA. On calibrations metrics, the gains are more for ST-LoRA compared to FRE while the performance improves slightly for FRE relative to the case where we use the 10 members, the performance for ST-LoRA improves significantly with lower standard deviation compared to FRE on NLL while higher standard deviation on ECE compared to FRE.

\subsection{Diversity Analysis}

To compare the diversity between FRE and ST-LoRA\footnote{We use the same LoRA configuration from Experiment 4.} members -and thereby understand how ensembling low-rank matrices differs from full-rank ensembling-we isolate two specific effects: (1) the learning rate scheduler (cosine annealing) and (2) the augmentation pipeline. We quantify disagreement using pairwise hard-label disagreement, symmetric Kullback-Leibler (KL) divergence - \textit{Jeffreys divergence}, disagreement on fraction on all valid pixels,and mutual information. Furthermore, we report expected entropy (aleatoric component) and total entropy (total uncertainty) to gain insight into how members estimate these two components under different configurations.\footnote{For pairwise argmax disagreement, we measure the fraction of member pairs whose hard predictions differ at a given pixel, we ignore the confidence by focusing only on the hard predicted label.}

Table~\ref{tab:div_all_conditions} compares the disagreement between FRE and ST-LoRA under three experimental conditions: (a) when both the scheduler and augmentation are absent, (b) when only augmentation is absent, and (c) when both augmentation and the scheduler are present. The comparison is evaluated across three scopes: foreground pixels only, all pixels, and macro averaging. 

For condition (a), across all  diversity metrics, ST-LoRA consistently exhibits lower diversity than FRE. We hypothesize that this reduced diversity stems from the absence of the scheduler; since the underlying models already occupy a \textit{proper} position in the hypothesis space, a constant learning rate limits diversity, particularly given the constrained low-rank space explored for fine-tuning. In contrast, FRE operates in a full-rank space, which enlarges the hypothesis space and thus increases the likelihood of obtaining more diverse ensemble members.

When the scheduler is introduced, the effect differs across the two methods: for ST-LoRA, we observe a clear increase in diversity compared to the scheduler-free case, whereas for FRE, diversity decreases. We hypothesize that this reduction for FRE is not a failure of the scheduler itself, but rather a consequence of the specific learning rate values chosen for the cycles. Given the expanded search space associated with full-rank fine-tuning, we suspect that higher learning rates than those used for ST-LoRA may be required to achieve comparable diversity \footnote{We kept the learning rate the same for ST-LoRA and FRE to isolate the effect of other hyperparameters.}. We leave a systematic investigation of this hypothesis to future work.

When augmentation is introduced, we observe a clear boost in diversity for both methods, which further underscores the importance of heterogeneous ensemble members over homogeneous ones. Furthermore, ST-LoRA's diversity increases and approaches that of the full-rank ensemble; in particular, mutual information among ST-LoRA members is higher than for FRE, which may help explain its strong performance on OoD tasks. 

Finally, we measure the mutual information per class in Table~\ref{tab:mutual_info_per_class} and observe that, while both approaches agree on the major classes, the mutual information is higher for ST-LoRA across almost all remaining classes. This indicates higher disagreement among LoRA members compared to FRE members. Collectively, these findings underscore the critical roles of both the scheduler and augmentation in achieving effective diversity.

\subsection{Qualitative Results}

\begin{figure*}[t]
    \centering
    \includegraphics[width=1\linewidth]{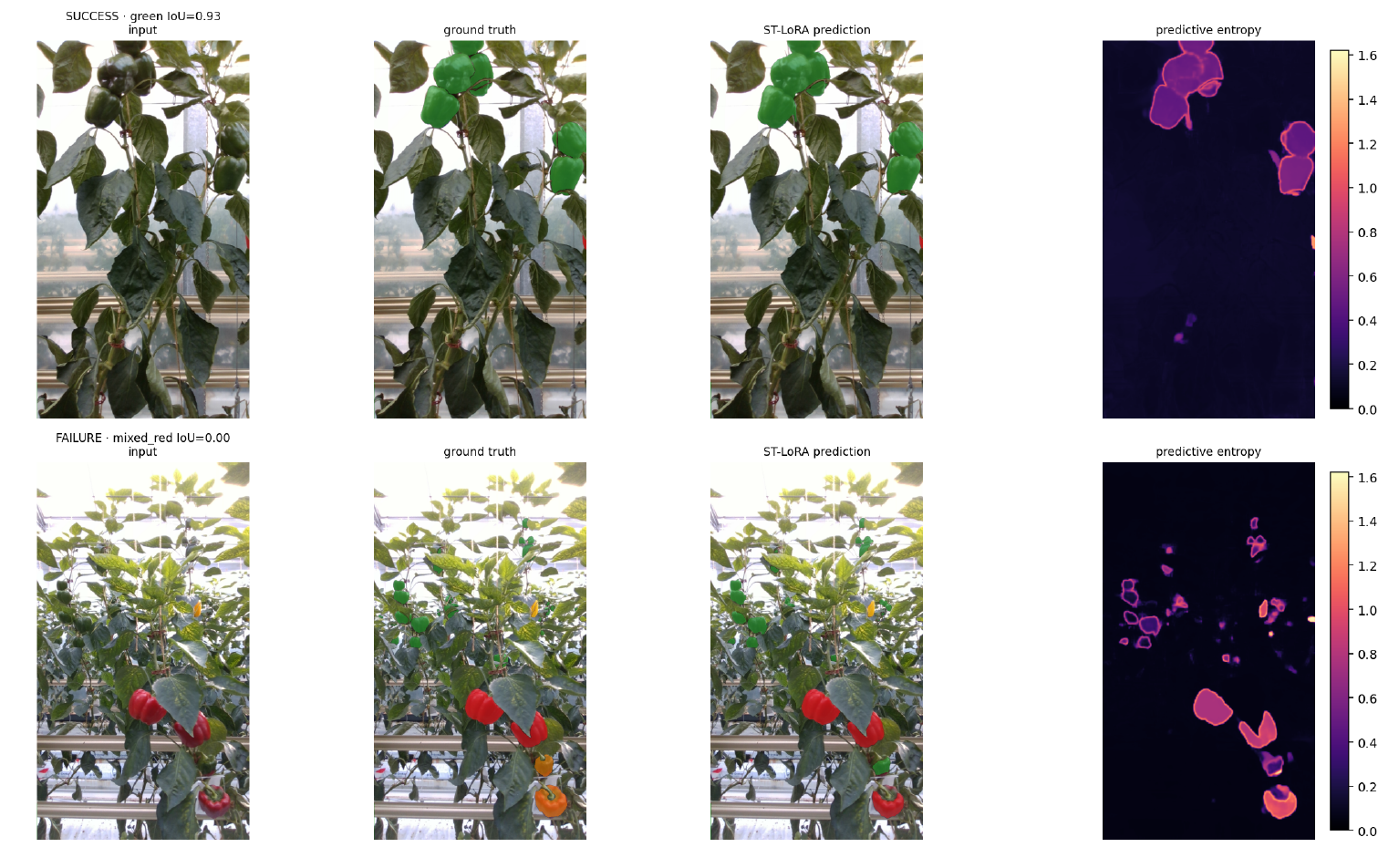}
    \caption{\textbf{Qualitative results for ST-LoRA approach on BUP20 dataset using Mask2Former model}. \textbf{(Top)}, demonstrates a success case where the approach is confident about the green peppers which are represented by low uncertainty. \textbf{(Bottom)}, represents a failure case, the approach is more uncertain regarding the mixed red class and completely confuse it with the pure red class.}
    \label{in_dist_scenario}
\end{figure*}

\textbf{Success and Failure Scenarios}. We present two qualitative cases the first represents ID analysis and OoD analysis where for both of the we show success and failure cases for our approach. As shown in Figure~\ref{in_dist_scenario} and according to the per class analysis our approach is segmenting green peppers correctly and confidently compared to the mixed red class which is second worst class the approach is performing on. 

\begin{figure*}[t]
    \centering
    \includegraphics[width=1\linewidth]{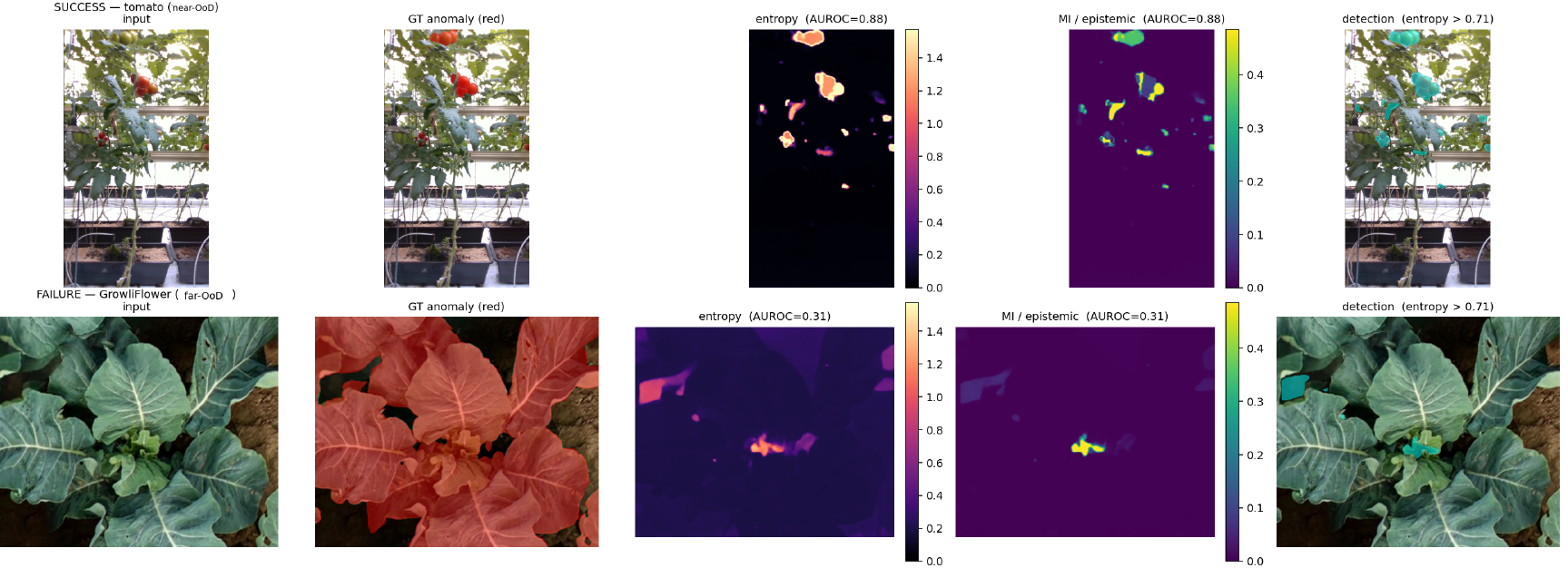}
    \caption{\textbf{Qualitative results for ST-LoRA approach on \textit{Near/Far} OoD detection using Mask2Former model}. \textbf{(Top)}, demonstrates a success case where the approach is able to localize the tomato plants using the uncertainty maps. \textbf{(Bottom)}, represents a failure case, the approach is challenged by the far OoD sample and fails to highlight the cauliflower plant.}
    \label{ood_scenario}
\end{figure*}

Figure~\ref{ood_scenario} shows that for near OoD tomato dataset the mutual information which is used to quantify epistemic uncertainty is useful to localize the regions that could be informative for the model for the case of Active Learning (AL) \cite{li2024survey} to enhance its performance on the new environment, compared to the cauliflower plants case where the mutual information and entropy fail to localize the plant.

\textbf{Aleatoric and Epistemic Decomposition}. Figure~\ref{fig:tu_au_eu_details} reports the three uncertainty components on the in-domain test set and on the near- and far-OoD datasets.\footnote{Averaged over 3 seeds, with 2000 and 1000 images sampled from BUTom21 and GrowliFlower-L respectively due to computational limitations.} ST-LoRA exhibits consistently higher aleatoric and total predictive uncertainty across all three domains, with a median increase in the aleatoric term of $+0.028$, $+0.046$ and $+0.147$ nats on BUP20, BUTom21 and GrowliFlower-L. The epistemic term behaves differently. In-domain, ST-LoRA members disagree slightly less than FRE members ($-0.0007$ nats), and this remains the case on near-OoD data ($-0.0068$ nats) although the gap narrows. On the far-OoD GrowliFlower-L set the ordering reverses: the disagreement between FRE members collapses, while ST-LoRA maintains a monotonic increase in mutual information with shift severity against FRE. Since only ST-LoRA's epistemic estimate tracks the severity of the shift, this could offer a direct explanation for its stronger image-level OoD detection performance. The same asymmetry holds for the total uncertainty, which increases only marginally for FRE as the shift grows but rises sharply for ST-LoRA as both components contribute.

We attribute the aleatoric gap to a difference in how the two adaptation schemes displace the pretrained representation rather than to a difference in fitted data noise. Under the entropy decomposition, the aleatoric term is the expected entropy of the individual members, which is a property of the fitted models and not of the underlying data. Full fine-tuning is free to move all members in the same direction away from the pretrained solution, so on inputs far from the training distribution they inherit the same distorted features and remain jointly confident; the low-rank constraint keeps each ST-LoRA member closer to the pretrained representation, where differences between members are still expressible on unfamiliar inputs.

\begin{figure*}[t]
    \centering
    \includegraphics[width=1\linewidth]{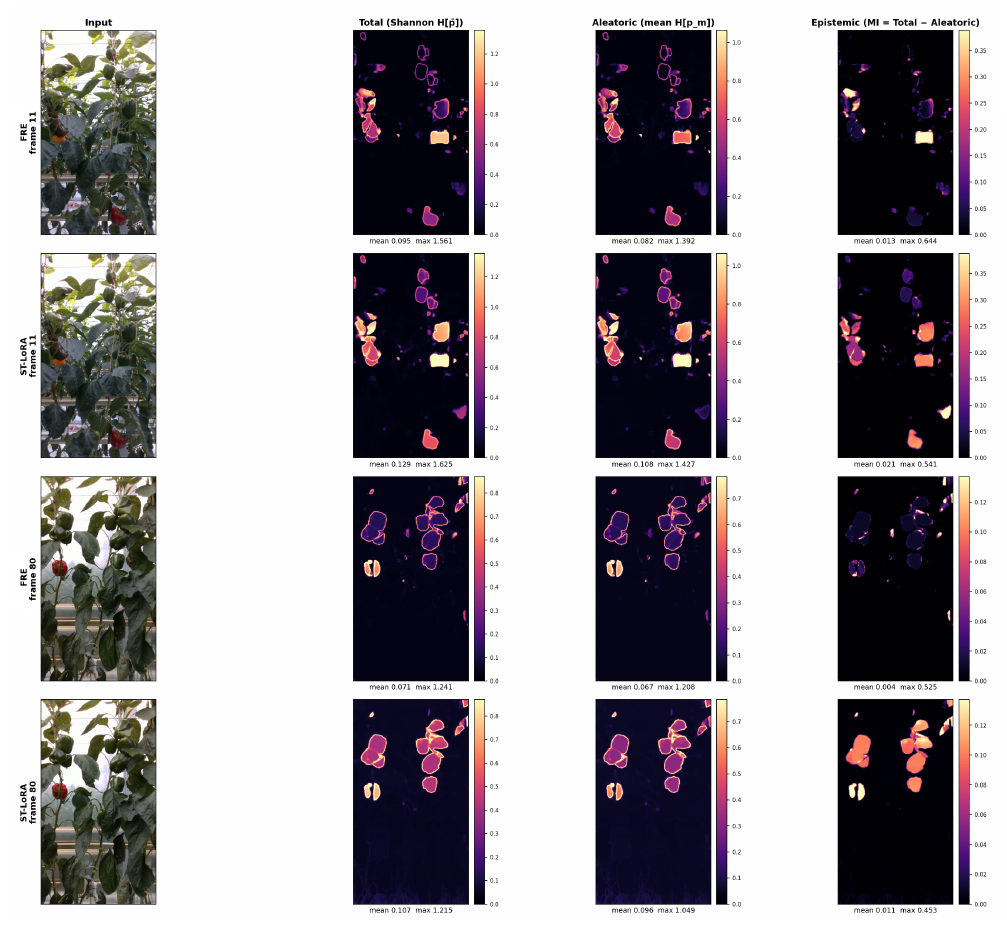}
    \caption{\textbf{Qaulitative results for pixel-wise uncertainty maps}. FRE shows estimates lower total and aleatoric uncertainties compared to ST-LoRA for the two frames, while we observe that for mutual information which is a measurement for epistemic uncertainty ST-LoRA has larger mean all over the image while the maximum remains higher for the FRE method.}
    \label{fig:tot_au_eu}
\end{figure*}

\begin{figure*}[t]
    \centering
    \includegraphics[width=1\textwidth]{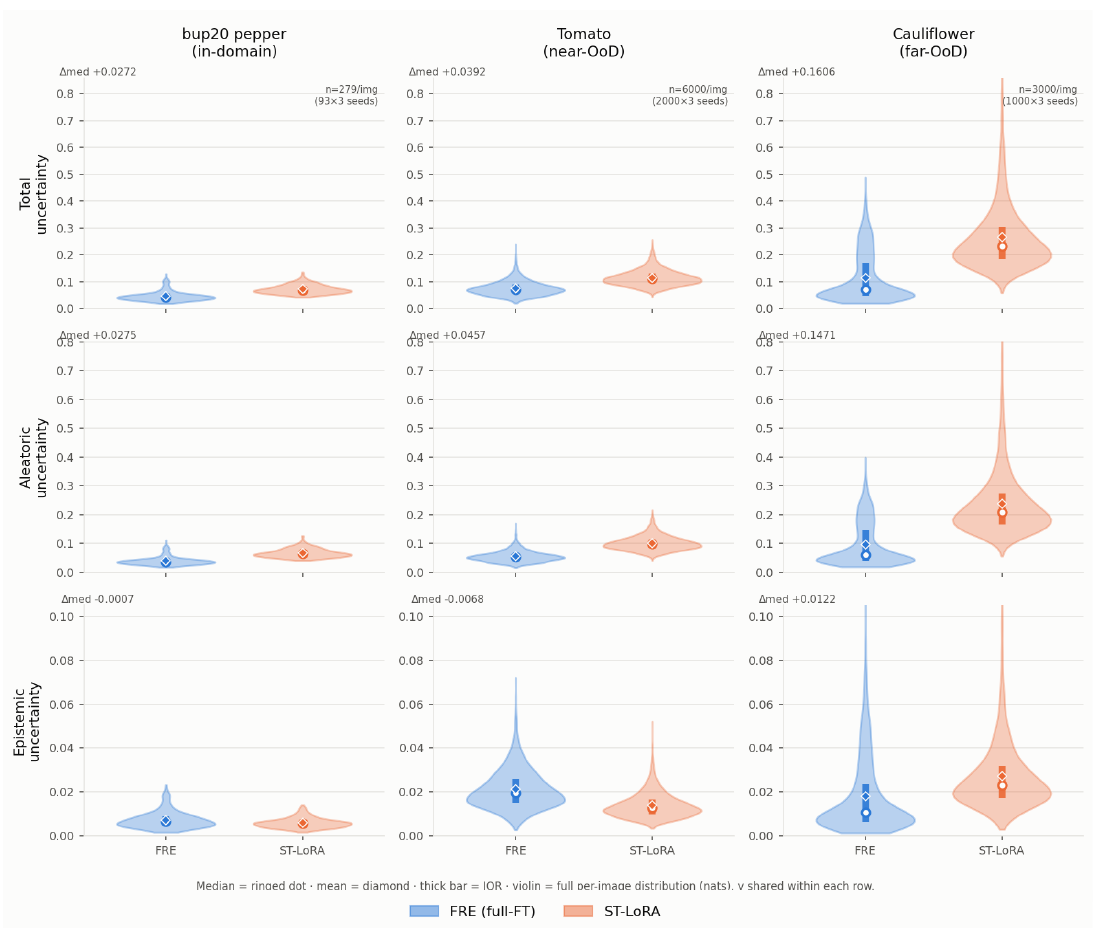}
    \caption{\textbf{Total predictive uncertainty decomposition to aleatoric and epistemic uncertainty for FRE and ST-LoRA.} }
    \label{fig:tu_au_eu_details}
\end{figure*}

Figure~\ref{fig:tot_au_eu} shows how these differences appear in the pixel-wise maps. For Frame~11 of the BUP20 test set, the ST-LoRA aleatoric map assigns elevated uncertainty over broader regions than FRE, while the epistemic maps of the two methods are comparable, yet ST-LoRA highlights more regions.

\subsection{LoRA Hyperparameters}

We ablate four additional hyperparameter configurations in this section, split into two parts: (A) rank and scaling ablations for SegFormer-B2 and EoMT, with SegFormer's decoder also adapted via LoRA; and (B) remaining hyperparameter ablations for SegFormer-B2 only.

\subsubsection{SegFormer and EoMT rank/$\alpha$ Ablation}

In Experiment~2, we used a heterogeneous setup between Mask2Former and SegFormer to assess whether LoRA adaptation could be performed using arbitrary hyperparameters without prior knowledge, comparing Mask2Former with an adapted decoder against SegFormer with a fully fine-tuned decoder across two datasets.

We extend this analysis by unifying the decoder adaptation strategy -applying LoRA to the decoder for both architectures- swhile also incorporating the EoMT architecture and restricting the evaluation to a single dataset (BUP20). We ablate rank ($r$) and scaling ($\alpha$) under a consistent experimental protocol: a baseline of $r{=}8$ and $\alpha{=}8$ is established, and each hyperparameter is varied independently while the other remains fixed. In all cases, the adapted modules include the feed-forward layers and attention mechanisms in the encoder.

For the rank ablation, as shown in Tables~\ref{tab:eomt_ablation}, \ref{tab:segformer_dh_ablation}, and~\ref{table15}, similar patterns emerge for SegFormer-B2 and Mask2Former. Both models achieve mIoU greater than $65\%$ with a rank as low as 2, and performance improves as the rank increases. In contrast, EoMT exhibits stable performance that remains largely unaffected by rank variation. Regarding calibration, SegFormer-B2 benefits from lower ranks (2--4), whereas EoMT and Mask2Former show better calibration with higher ranks (8--16).

For the $\alpha$ ablation, increasing the scaling value generally improves mIoU; SegFormer-B2 and Mask2Former perform best around $\alpha = 2r$, while EoMT continues to improve beyond this range. In terms of calibration, larger $\alpha$ values lead to better calibration across all three models. However, we observe a collapse for SegFormer-B2 at $\alpha = 64$, whereas Mask2Former and EoMT remain stable and exhibit similar performance without such degradation.\footnote{EoMT follows a distinct pattern, which we hypothesize is due to its underlying DINOv2 pretrained architecture that likely renders the model considerably less sensitive to hyperparameter variations..}

\subsubsection{ST-LoRA Initializer}

We also evaluate the impact of different ST-LoRA weight initialization strategies, including EVA, OST-LoRA, and PISSA \cite{paischer2024parameter, buyukakyuz2024olora, meng2024pissa}. As shown in Table~\ref{table15}, introducing these specialized initializers does not lead to consistent performance improvements. The baseline configuration without any dedicated ST-LoRA initializer achieves the best overall performance, with a mean IoU of $0.8178 \pm 0.0295$, outperforming EVA ($0.7795 \pm 0.0455$), OLoRA ($0.7872 \pm 0.0850$), and PISSA ($0.8097 \pm 0.0172$).

While some initializers yield competitive results, without specialized intializer option provide a clear advantage over the default initialization. This observation suggests that, for the considered segmentation task, explicit ST-LoRA initialization strategies may not be necessary. However, further experiments on different datasets and architectures would be required to confirm whether this behavior generalizes beyond the current setting.

\subsubsection{Bias Setting} 
We further investigate the impact of different bias training strategies, including fine-tuning all biases (Backbone and ST-LoRA), fine-tuning only ST-LoRA biases, and freezing all biases. The results indicate that the performance gap between the None and ST-LoRA-only bias settings is negligible. In contrast, fine-tuning all biases leads to a substantial degradation in performance, resulting in a mean IoU of $0.4348 \pm 0.3574$ as shown in Table~\ref{table15}.

This significant drop in performance may be related to the fact that the backbone has been pre-trained and already encodes useful representations through transfer learning. Allowing the backbone biases to be updated can perturb these learned representations, particularly given the depth of the network. Properly adapting such pre-trained components typically requires larger training datasets and longer training to avoid destabilizing previously learned features. Under the current experimental setup, where data and training duration are limited, fine-tuning all biases appears to introduce instability rather than improving adaptation.

\subsubsection {Batch Size and Learning Rate}
While varying the batch size does not appear to have a noticeable effect on performance, the learning rate plays a crucial role in model behavior and stability.

Ablation on the learning rate reveals two extreme regimes. When the learning rate is set to a large value ($2 \times 10^{-1}$), the model completely fails to learn meaningful representations as shown in Table~\ref{table15}, resulting in a mean IoU of $0.0$ and a False Negative Rate of $1.0$. This indicates that the model collapses to predicting only the background class. Such behavior is expected, as an excessively high learning rate can destabilize optimization, causing the model to overshoot minima and converge to degenerate solutions. Specifically in \cite{Biderman2024LoRALL} they mention that LoRA adapters need smaller learning rate compared to full rank fine tuning.

At the other extreme, smaller learning rates such as $2 \times 10^{-3}$ and $2 \times 10^{-4}$ yield the highest mean IoU scores in our experiments ($0.8860 \pm 0.0018$ and $0.8802 \pm 0.0023$, respectively). Although these results might initially suggest that very small learning rates are optimal. Very small learning rates might limit ST-LoRA diversity as we expect that with small learning rate the approach lose an aspect of diversification which leads to a degenerate solution reached by different checkpoints.

\subsubsection {LoRA Dropout and Frozen Encoder/Decoder}

As shown in Table~\ref{table16}, fine tuning the dropout value enhances the performance. The baseline with $p=0.10$ achieves mIoU of $68.09\%$ while the best result achieved is $69.0\%$ with $p=0.20$. However the calibration metrics for the former dropout value is smaller compared to the later one, which reflects the tension between optimizing for segmentation quality versus miscalibration reduction. Table~\ref{table16} shows that with a frozen encoder and increasing the rank monotonically, mIoU and calibration metrics have shown the worst performance compared to the case where only one component from the encoder to be adapted by LoRA.

For Decoder, we tried to compare the role of fine tuning the decoder compared to having it frozen during encoder fine-tuning. Table~\ref{table18} shows a clear trend, for nearly all cases the frozen decoder affects not only the segmentation quality but degrades the calibration heavily.

\begin{table*}[t]
\centering
\caption{\textbf{Effect of ST-LoRA hyperparameters and architectural choices on EoMT segmentation performance and calibration on BUP20 test set.} 
The baseline configuration is ($r{=}8,\alpha{=}8$). $\uparrow$: higher is better; $\downarrow$: lower is better. 
All results report mean\,$\pm$\,std over 5 seeds; best result per 
sweep in \textbf{bold}.}
\resizebox{0.95\textwidth}{!}{
\begin{tabular}{llccccc}
\toprule
\textbf{Category} & \textbf{Setting} & \textbf{mIoU} $\uparrow$ & \textbf{ECE} $\downarrow$ & \textbf{ACE} $\downarrow$ & \textbf{Brier} $\downarrow$ & \textbf{NLL} $\downarrow$ \\
\midrule
\multirow{6}{*}{LoRA rank -- $r$ ($\alpha{=}8$)}
& 2   & 0.6757\,$\pm$\,0.0020 & 0.0030\,$\pm$\,0.0005 & 0.0041\,$\pm$\,0.0005 & 0.0188\,$\pm$\,0.0002 & 0.0407\,$\pm$\,0.0015 \\
& 4   & \textbf{0.6780\,$\pm$\,0.0016} & 0.0029\,$\pm$\,0.0013 & \textbf{0.0034\,$\pm$\,0.0007} & \textbf{0.0185\,$\pm$\,0.0001} & 0.0413\,$\pm$\,0.0021  \\
& 8 (baseline)  & 0.6753\,$\pm$\,0.0024 & 0.0032\,$\pm$\,0.0007 & 0.0041\,$\pm$\,0.0005 & \textbf{0.0185\,$\pm$\,0.0004} & \textbf{0.0403\,$\pm$\,0.0006}  \\
& 16  & 0.6766\,$\pm$\,0.0024 & \textbf{0.0027\,$\pm$\,0.0010} & 0.0038\,$\pm$\,0.0008 & \textbf{0.0185\,$\pm$\,0.0003} & 0.0404\,$\pm$\,0.0013  \\
& 32  & 0.6762\,$\pm$\,0.0039 & 0.0032\,$\pm$\,0.0006 & 0.0040\,$\pm$\,0.0002 & 0.0187\,$\pm$\,0.0004 & 0.0411\,$\pm$\,0.0010 \\
& 64  & 0.6750\,$\pm$\,0.0032 & 0.0030\,$\pm$\,0.0006 & 0.0039\,$\pm$\,0.0005 & 0.0187\,$\pm$\,0.0005 & 0.0416\,$\pm$\,0.0017  \\
\midrule
\multirow{6}{*}{LoRA scaling -- $\alpha$ ($r{=}8$)}
& 2   & 0.6694\,$\pm$\,0.0016 & 0.0048\,$\pm$\,0.0031 & 0.0050\,$\pm$\,0.0012 & 0.0192\,$\pm$\,0.0003 & 0.0442\,$\pm$\,0.0030  \\
& 4   & 0.6710\,$\pm$\,0.0051 & \textbf{0.0023\,$\pm$\,0.0004} & 0.0038\,$\pm$\,0.0008 & 0.0188\,$\pm$\,0.0003 & 0.0418\,$\pm$\,0.0011 \\
& 8 (baseline)  & 0.6753\,$\pm$\,0.0024 & 0.0032\,$\pm$\,0.0007 & 0.0041\,$\pm$\,0.0005 & 0.0185\,$\pm$\,0.0004 & 0.0403\,$\pm$\,0.0006  \\
& 16  & 0.6813\,$\pm$\,0.0034 & 0.0029\,$\pm$\,0.0012 & \textbf{0.0033\,$\pm$\,0.0010} & 0.0180\,$\pm$\,0.0004 & 0.0389\,$\pm$\,0.0010 \\
& 32  & 0.6835\,$\pm$\,0.0034 & 0.0039\,$\pm$\,0.0007 & 0.0040\,$\pm$\,0.0005 & 0.0177\,$\pm$\,0.0005 & 0.0384\,$\pm$\,0.0014 \\
& 64  & \textbf{0.6882\,$\pm$\,0.0051} & 0.0033\,$\pm$\,0.0007 & 0.0034\,$\pm$\,0.0006 & \textbf{0.0172\,$\pm$\,0.0005} & \textbf{0.0366\,$\pm$\,0.0012} \\
\bottomrule
\end{tabular}}
\label{tab:eomt_ablation}
\end{table*}

\begin{table*}[t]
\centering
\caption{\textbf{Effect of ST-LoRA rank and scaling on SegFormer-B2 
decode-head adaptation performance and calibration on clean BUP20 
test set.} The baseline is ($r{=}8,\alpha{=}8$). $\uparrow$: higher is better; $\downarrow$: lower 
is better. All results report mean\,$\pm$\,std over 5 seeds; best 
result per sweep in \textbf{bold}.}
\resizebox{\textwidth}{!}{
\begin{tabular}{llccccccc}
\toprule
\textbf{Category} & \textbf{Setting} & \textbf{mIoU} $\uparrow$ & \textbf{ECE} $\downarrow$ & \textbf{ACE} $\downarrow$ & \textbf{MECE} $\downarrow$ & \textbf{MACE} $\downarrow$ & \textbf{Brier} $\downarrow$ & \textbf{NLL} $\downarrow$ \\
\midrule
\multirow{6}{*}{LoRA rank -- $r$ ($\alpha{=}8$)}
& 2   & 0.6678\,$\pm$\,0.0071 & 0.0025\,$\pm$\,0.0009 & \textbf{0.0030\,$\pm$\,0.0006} & \textbf{0.0477\,$\pm$\,0.0105} & \textbf{0.0243\,$\pm$\,0.0053} & 0.0190\,$\pm$\,0.0005 & 0.0411\,$\pm$\,0.0016 \\
& 4   & 0.6699\,$\pm$\,0.0035 & \textbf{0.0022\,$\pm$\,0.0007} & \textbf{0.0030\,$\pm$\,0.0006} & 0.0658\,$\pm$\,0.0307 & \textbf{0.0243\,$\pm$\,0.0053} & \textbf{0.0185\,$\pm$\,0.0003} & \textbf{0.0398\,$\pm$\,0.0008} \\
& 8 (baseline)  & 0.6731\,$\pm$\,0.0043 & 0.0023\,$\pm$\,0.0004 & 0.0036\,$\pm$\,0.0004 & 0.0800\,$\pm$\,0.0670 & 0.0289\,$\pm$\,0.0044 & 0.0186\,$\pm$\,0.0004 & 0.0421\,$\pm$\,0.0013 \\
& 16  & 0.6762\,$\pm$\,0.0037 & 0.0025\,$\pm$\,0.0005 & 0.0039\,$\pm$\,0.0004 & 0.0653\,$\pm$\,0.0193 & 0.0317\,$\pm$\,0.0041 & 0.0186\,$\pm$\,0.0002 & 0.0422\,$\pm$\,0.0010 \\
& 32  & \textbf{0.6752\,$\pm$\,0.0078} & 0.0025\,$\pm$\,0.0007 & 0.0038\,$\pm$\,0.0005 & 0.0639\,$\pm$\,0.0227 & 0.0309\,$\pm$\,0.0052 & 0.0185\,$\pm$\,0.0005 & 0.0425\,$\pm$\,0.0011 \\
& 64  & 0.6738\,$\pm$\,0.0036 & 0.0026\,$\pm$\,0.0003 & 0.0037\,$\pm$\,0.0001 & 0.0836\,$\pm$\,0.0604 & 0.0293\,$\pm$\,0.0017 & 0.0186\,$\pm$\,0.0002 & 0.0417\,$\pm$\,0.0010 \\
\midrule
\multirow{6}{*}{LoRA scaling -- $\alpha$ ($r{=}8$)}
& 2   & 0.6597\,$\pm$\,0.0045 & 0.0026\,$\pm$\,0.0005 & 0.0033\,$\pm$\,0.0004 & 0.0604\,$\pm$\,0.0261 & 0.0261\,$\pm$\,0.0040 & 0.0193\,$\pm$\,0.0004 & 0.0415\,$\pm$\,0.0013 \\
& 4   & 0.6658\,$\pm$\,0.0037 & 0.0028\,$\pm$\,0.0002 & 0.0037\,$\pm$\,0.0004 & 0.0804\,$\pm$\,0.0574 & 0.0306\,$\pm$\,0.0044 & 0.0190\,$\pm$\,0.0002 & 0.0417\,$\pm$\,0.0007 \\
& 8 (baseline)  & 0.6731\,$\pm$\,0.0043 & \textbf{0.0023\,$\pm$\,0.0004} & 0.0036\,$\pm$\,0.0004 & 0.0800\,$\pm$\,0.0670 & 0.0289\,$\pm$\,0.0044 & 0.0186\,$\pm$\,0.0004 & 0.0421\,$\pm$\,0.0013 \\
& 16  & \textbf{0.6808\,$\pm$\,0.0081} & 0.0020\,$\pm$\,0.0003 & \textbf{0.0033\,$\pm$\,0.0003} & \textbf{0.0503\,$\pm$\,0.0093} & \textbf{0.0264\,$\pm$\,0.0031} & \textbf{0.0182\,$\pm$\,0.0004} & \textbf{0.0410\,$\pm$\,0.0012} \\
& 32  & 0.6668\,$\pm$\,0.0373 & 0.0125\,$\pm$\,0.0155 & 0.0097\,$\pm$\,0.0095 & 0.1112\,$\pm$\,0.0778 & 0.0500\,$\pm$\,0.0402 & 0.0202\,$\pm$\,0.0043 & 0.0469\,$\pm$\,0.0112 \\
& 64  & 0.0166\,$\pm$\,0.0371 & 0.0336\,$\pm$\,0.0459 & 0.0215\,$\pm$\,0.0194 & 0.1950\,$\pm$\,0.2773 & 0.0778\,$\pm$\,0.1036 & 0.0742\,$\pm$\,0.0125 & 0.1813\,$\pm$\,0.0441 \\
\bottomrule
\end{tabular}}
\label{tab:segformer_dh_ablation}
\end{table*}

\begin{table*}[t]
\centering
\caption{\textbf{Effect of ST-LoRA hyperparameters and architectural choices on 
SegFormer-B2 performance on GrowliFlower-L}. $\uparrow$: higher is better; $\downarrow$: lower is better. 
All results report mean\,$\pm$\,std over 5 seeds, and best result per metric in \textbf{bold}.}
\resizebox{0.85\textwidth}{!}{

\begin{tabular}{llccccc}

\textbf{Category} & \textbf{Setting} & \textbf{mIoU} $\uparrow$ & \textbf{FNR} $\downarrow$ & \textbf{ECE} $\downarrow$ & \textbf{MECE} $\downarrow$ & \textbf{Trainable Parameters} (\%) $\downarrow$ \\
\hline
\multirow{6}{*}{LoRA rank - $r$}
& 2   & 0.3151 $\pm$ 0.3863 & 0.6574 $\pm$ 0.4205 & 0.0646 $\pm$ 0.0303 & 0.1367 $\pm$ 0.0863 & 4.64 \\
& 4   & 0.5553 $\pm$ 0.3191 & 0.4009 $\pm$ 0.3532 & 0.1021 $\pm$ 0.0338 & 0.2905 $\pm$ 0.1190 & 5.11 \\
& 8   & 0.7285 $\pm$ 0.1081 & 0.2151 $\pm$ 0.1460 & 0.0744 $\pm$ 0.0688 & 0.2864 $\pm$ 0.1483 & 6.03 \\
& 16  & 0.6277 $\pm$ 0.3154 & 0.3257 $\pm$ 0.3409 & 0.0529 $\pm$ 0.0350 & 0.1748 $\pm$ 0.0640 & 7.81 \\
& 32 (baseline) & 0.8178 $\pm$ 0.0295 & 0.1093 $\pm$ 0.0410 & 0.0126 $\pm$ 0.0166 & 0.1367 $\pm$ 0.0198 & 11.19 \\
& 64  & 0.8134 $\pm$ 0.0194 & 0.0973 $\pm$ 0.0442 & 0.0329 $\pm$ 0.0338 & 0.2165 $\pm$ 0.0835 & 17.26 \\
& 128 & \textbf{0.8231 $\pm$ 0.0110 }& \textbf{0.0879 $\pm$ 0.0180} & \textbf{0.0062 $\pm$ 0.0041} & \textbf{0.1607 $\pm$ 0.0750} & 27.20 \\

\hline

\multirow{5}{*}{Components}
& $q$, $kv$, $fc_1$, $fc_2$ (baseline) & 0.8178 $\pm$ 0.0295 & 0.1093 $\pm$ 0.0410 & 0.0126 $\pm$ 0.0166 & 0.1367 $\pm$ 0.0198 & 11.19 \\
& $q$, $fc_1$, $fc_2$  & 0.8469 $\pm$ 0.0021 & 0.0726 $\pm$ 0.0039 & 0.0041 $\pm$ 0.0008 & 0.1561 $\pm$ 0.0348 & 9.87 \\
& $q$, $kv$, $fc_1$  & 0.8429 $\pm$ 0.0095 & 0.0719 $\pm$ 0.0097 & 0.0038 $\pm$ 0.0018 & 0.1280 $\pm$ 0.0601 & 8.97 \\
& $q$, $kv$, $fc_2$ & 0.7505 $\pm$ 0.1439 & 0.1810 $\pm$ 0.1765 & 0.0358 $\pm$ 0.0604 & 0.1916 $\pm$ 0.1460 & 8.97 \\
& $q$, $kv$ & 0.8158 $\pm$ 0.0273 & 0.1300 $\pm$ 0.0394 & 0.0037 $\pm$ 0.0013 & 0.0999 $\pm$ 0.0290 & 6.63 \\
& $fc_1$, $fc_2$ & 0.8509 $\pm$ 0.0023 & 0.0718 $\pm$ 0.0045 & 0.0034 $\pm$ 0.0003 & 0.1231 $\pm$ 0.0124 & 8.97 \\
& $kv$ & 0.8311 $\pm$ 0.0128 & 0.1064 $\pm$ 0.0145 & \textbf{0.0025 $\pm$ 0.0018} & \textbf{0.0705 $\pm$ 0.0430} & 8.97 \\
& $q$ & 0.8551 $\pm$ 0.0012 & 0.0557 $\pm$ 0.0016 & 0.0035 $\pm$ 0.0002 & 0.1163 $\pm$ 0.0057 & 8.97 \\
& $fc_2$ & 0.8498 $\pm$ 0.0046 & 0.0715 $\pm$ 0.0053 & 0.0035 $\pm$ 0.0002 & 0.1367 $\pm$ 0.0075 & 6.63 \\
& $fc_1$ & \textbf{0.8613 $\pm$ 0.0018} & \textbf{0.0513 $\pm$ 0.0021} & 0.0042 $\pm$ 0.0003 & 0.1686 $\pm$ 0.0079 & 6.63 \\

\hline

\multirow{5}{*}{LoRA scaling factor - $\alpha$}
& 8   & \textbf{0.8622 $\pm$ 0.0080} & \textbf{ 0.0498 $\pm$ 0.0109} & 0.0132 $\pm$ 0.0173 & 0.1719 $\pm$ 0.0334 & 11.19 \\
& 16  & 0.8531 $\pm$ 0.0042 & 0.0551 $\pm$ 0.0034 & \textbf{0.0047 $\pm$ 0.0007} & 0.1760 $\pm$ 0.0190 & 11.19 \\
& 32  & 0.8290 $\pm$ 0.0163 & 0.0709 $\pm$ 0.0104 & 0.0091 $\pm$ 0.0085 & 0.2005 $\pm$ 0.0804 & 11.19 \\
& 64 (baseline) & 0.8178 $\pm$ 0.0295 & 0.1093 $\pm$ 0.0410 & 0.0126 $\pm$ 0.0166 & 0.1367 $\pm$ 0.0198 & 11.19 \\
& 128 & 0.4645 $\pm$ 0.3796 & 0.5072 $\pm$ 0.4031 & 0.0492 $\pm$ 0.0379 & \textbf{0.1184 $\pm$ 0.1015} & 11.19 \\

\hline
\multirow{4}{*}{Initializer}
& EVA   & 0.7795 $\pm$ 0.0455 & 0.1557 $\pm$ 0.0580 & 0.0130 $\pm$ 0.0164 & 0.1256 $\pm$ 0.0574 & 11.19 \\
& OLoRA & 0.7872 $\pm$ 0.0850 & 0.1244 $\pm$ 0.1202 & 0.0261 $\pm$ 0.0350 & 0.2331 $\pm$ 0.0929 & 11.19 \\
& PISSA & 0.8097 $\pm$ 0.0172 & 0.1224 $\pm$ 0.0291 & 0.0299 $\pm$ 0.0353 & 0.1527 $\pm$ 0.0888 & 11.19 \\
& None (baseline) & \textbf{0.8178 $\pm$ 0.029}5 & \textbf{0.1093 $\pm$ 0.0410} & \textbf{0.0126 $\pm$ 0.0166} & \textbf{0.1367 $\pm$ 0.0198} & 11.19 \\

\hline
\multirow{3}{*}{Bias Setting}
& All & 0.4348 $\pm$ 0.3574 & 0.5515 $\pm$ 0.3690 & 0.0703 $\pm$ 0.0377 & 0.2537 $\pm$ 0.0958 & 11.45 \\
& LoRA-only & 0.8103 $\pm$ 0.0232 & 0.1143 $\pm$ 0.0456 & 0.0142 $\pm$ 0.0160 & 0.1599 $\pm$ 0.0479 & 11.31 \\
& None (baseline) & 0.8178 $\pm$ 0.0295 & 0.1093 $\pm$ 0.0410 & 0.0126 $\pm$ 0.0166 & 0.1367 $\pm$ 0.0198 & 11.19 \\

\hline
\multirow{3}{*}{Batch Size}
& 8  & \textbf{0.8254 $\pm$ 0.0098} & \textbf{0.0899 $\pm$ 0.0162} & \textbf{0.0050 $\pm$ 0.0015} & 0.1405 $\pm$ 0.0518 & 11.19 \\
& 16 & 0.8106 $\pm$ 0.0212 & 0.1124 $\pm$ 0.0489 & 0.0408 $\pm$ 0.0421 & 0.2222 $\pm$ 0.0916 & 11.19 \\
& 32 (baseline) & 0.8178 $\pm$ 0.0295 & 0.1093 $\pm$ 0.0410 & 0.0126 $\pm$ 0.0166 & \textbf{0.1367 $\pm$ 0.0198} & 11.19 \\

\hline
\multirow{4}{*}{Learning Rate}
& $2\times10^{-1}$ & 0.0000 $\pm$ 0.0000 & 1.0000 $\pm$ 0.0000 & 0.0820 $\pm$ 0.0148 & \textbf{0.0820 $\pm$ 0.0148} & 11.19 \\
& $2\times10^{-2}$ (baseline) & 0.8178 $\pm$ 0.0295 & 0.1093 $\pm$ 0.0410 & 0.0126 $\pm$ 0.0166 & 0.1367 $\pm$ 0.0198 & 11.19 \\
& $2\times10^{-3}$ & \textbf{0.8860 $\pm$ 0.0018} & \textbf{0.0254 $\pm$ 0.0013} & \textbf{0.0050 $\pm$ 0.0003} & 0.2876 $\pm$ 0.0138 & 11.19 \\
& $2\times10^{-4}$ & 0.8802 $\pm$ 0.0023 & 0.0274 $\pm$ 0.0049 & 0.0057 $\pm$ 0.0005 & 0.2590 $\pm$ 0.0427 & 11.19 \\

\hline
\multirow{4}{*}{$\eta_{\min}$}
& $2\!\times\!10^{-3}$ & 0.8135 $\pm$ 0.0021 & 0.1011 $\pm$ 0.0161 & 0.0248 $\pm$ 0.0188 & 0.2021 $\pm$ 0.0764 & 11.19\\
& $2\!\times\!10^{-4}$ (baseline) & 0.8178 $\pm$ 0.0295 & 0.1093 $\pm$ 0.0410 & 0.0126 $\pm$ 0.0166 & 0.1367 $\pm$ 0.0198 & 11.19\\
& $2\!\times\!10^{-5}$ & 0.8096 $\pm$ 0.0212 & 0.1132 $\pm$ 0.0438 & 0.0310 $\pm$ 0.0330 & 0.2046 $\pm$ 0.0881 & 11.19\\
& $2\!\times\!10^{-6}$ & 0.8148 $\pm$ 0.0304 & 0.0983 $\pm$ 0.0543 & 0.0240 $\pm$ 0.0341 & 0.2168 $\pm$ 0.0735 & 11.19\\

\hline
\end{tabular}}
\label{table15}
\end{table*}

\begin{table*}[t]
\centering
\caption{\textbf{Effect of ST-LoRA hyperparameters and architectural choices on Mask2Former segmentation performance on BUP20.} $q_d$, $k_d$, and cls denote the transformer decoder's query projection, key projection, and class predictor, respectively. $q_e$, $k_e$, $v_e$, and dense$_e$ refer to the Swin-B encoder's query, key, value projections and dense (MLP) layers. The rank, $\alpha$, and dropout sweeps use $q_d$, $k_d$, cls, dense$_e$ as target modules. $\uparrow$: higher is better; $\downarrow$: lower is better. 
All results report mean\,$\pm$\,std over 5 seeds. 
and best result per metric in \textbf{bold}.}
\resizebox{0.85\textwidth}{!}{

\begin{tabular}{llccccc}

\textbf{Category} & \textbf{Setting} & \textbf{mIoU} $\uparrow$ & \textbf{ECE }$\downarrow$ & \textbf{MECE} $\downarrow$ & \textbf{ACE} $\downarrow$ & \textbf{MACE} $\downarrow$ \\
\hline
\multirow{6}{*}{ST-LoRA rank -- $r$}
& 2   & 0.6582 $\pm$ 0.0170 & 0.0143 $\pm$ 0.0058 & 0.1507 $\pm$ 0.0410 & 0.0646 $\pm$ 0.0331 & 0.0961 $\pm$ 0.0377 \\
& 4   & 0.6734 $\pm$ 0.0223 & 0.0089 $\pm$ 0.0049 & 0.0833 $\pm$ 0.0361 & 0.0306 $\pm$ 0.0222 & 0.0510 $\pm$ 0.0246 \\
& 8 (baseline)  & 0.6809 $\pm$ 0.0171 & \textbf{0.0028 $\pm$ 0.0017} & 0.0939 $\pm$ 0.0612 & 0.0041 $\pm$ 0.0020 & 0.0092 $\pm$ 0.0060 \\
& 16  & 0.6791 $\pm$ 0.0292 & 0.0036 $\pm$ 0.0019 & 0.1672 $\pm$ 0.0320 & 0.0024 $\pm$ 0.0009 & \textbf{0.0091 $\pm$ 0.0059} \\
& 32  & 0.6801 $\pm$ 0.0239 & 0.0030 $\pm$ 0.0003 & \textbf{0.0798 $\pm$ 0.0653} & \textbf{0.0018 $\pm$ 0.0008} & 0.0106 $\pm$ 0.0094 \\
& 64  & \textbf{0.6942 $\pm$ 0.0122} & 0.0045 $\pm$ 0.0039 & 0.1053 $\pm$ 0.0799 & 0.0019 $\pm$ 0.0025 & 0.0099 $\pm$ 0.0155 \\

\hline
\multirow{5}{*}{ST-LoRA scaling -- $\alpha$}
& 2   & 0.6763 $\pm$ 0.0097 & 0.0161 $\pm$ 0.0043 & 0.2020 $\pm$ 0.0945 & 0.0505 $\pm$ 0.0202 & 0.0759 $\pm$ 0.0260 \\
& 4   & 0.6773 $\pm$ 0.0230 & 0.0062 $\pm$ 0.0015 & 0.0678 $\pm$ 0.0147 & 0.0137 $\pm$ 0.0045 & 0.0245 $\pm$ 0.0060 \\
& 8 (baseline)  & 0.6809 $\pm$ 0.0171 & 0.0028 $\pm$ 0.0017 & 0.0939 $\pm$ 0.0612 & 0.0041 $\pm$ 0.0020 & 0.0092 $\pm$ 0.0060 \\
& 16  & \textbf{0.6831 $\pm$ 0.0140} & \textbf{0.0026 $\pm$ 0.0006} & 0.1033 $\pm$ 0.0968 & \textbf{0.0026 $\pm$ 0.0006} & \textbf{0.0074 $\pm$ 0.0044} \\
& 32  & 0.6766 $\pm$ 0.0145 & 0.0030 $\pm$ 0.0020 & \textbf{0.0746 $\pm$ 0.0402} & 0.0022 $\pm$ 0.0005 & 0.0087 $\pm$ 0.0029 \\

\hline
\multirow{4}{*}{ST-LoRA dropout}
& 0.10 (baseline) & 0.6809 $\pm$ 0.0171 & \textbf{0.0028 $\pm$ 0.0017} & 0.0939 $\pm$ 0.0612 & 0.0041 $\pm$ 0.0020 & 0.0092 $\pm$ 0.0060 \\
& 0.15 & 0.6734 $\pm$ 0.0165 & 0.0037 $\pm$ 0.0020 & 0.1277 $\pm$ 0.0551 & 0.0048 $\pm$ 0.0010 & 0.0094 $\pm$ 0.0039 \\
& 0.20 & \textbf{0.6908 $\pm$ 0.0153} & 0.0044 $\pm$ 0.0019 & 0.0752 $\pm$ 0.0342 & 0.0047 $\pm$ 0.0022 & 0.0098 $\pm$ 0.0060 \\
& 0.25 & 0.6825 $\pm$ 0.0209 & 0.0035 $\pm$ 0.0024 & 0.0711 $\pm$ 0.0301 & 0.0053 $\pm$ 0.0020 & 0.0105 $\pm$ 0.0054 \\
& 0.30 & 0.6752 $\pm$ 0.0144 & 0.0035 $\pm$ 0.0017 & \textbf{0.0622 $\pm$ 0.0428} & \textbf{0.0041 $\pm$ 0.0024} & \textbf{0.0081 $\pm$ 0.0071} \\

\hline
\multirow{7}{*}{Components}
& $q_d$, $k_d$, cls, dense$_e$, $q_e$                          & 0.6870 $\pm$ 0.0199 & \textbf{0.0030 $\pm$ 0.0013} & \textbf{0.0665 $\pm$ 0.0285} & 0.0042 $\pm$ 0.0018 & 0.0093 $\pm$ 0.0040 \\
& $q_d$, $k_d$, cls, dense$_e$, $k_e$                          & \textbf{0.6976 $\pm$ 0.0099} & 0.0038 $\pm$ 0.0013 & 0.0796 $\pm$ 0.0518 & 0.0049 $\pm$ 0.0031 & 0.0098 $\pm$ 0.0051 \\
& $q_d$, $k_d$, cls, dense$_e$, $v_e$                          & 0.6599 $\pm$ 0.0248 & 0.0030 $\pm$ 0.0011 & 0.0574 $\pm$ 0.0408 & \textbf{0.0040 $\pm$ 0.0019} & 0.0090 $\pm$ 0.0039 \\
& $q_d$, $k_d$, cls, dense$_e$, $q_e$, $k_e$                   & 0.6760 $\pm$ 0.0152 & 0.0027 $\pm$ 0.0013 & 0.0704 $\pm$ 0.0424 & 0.0032 $\pm$ 0.0010 & \textbf{0.0075 $\pm$ 0.0022} \\
& $q_d$, $k_d$, cls, dense$_e$, $q_e$, $v_e$                   & 0.6828 $\pm$ 0.0154 & 0.0041 $\pm$ 0.0026 & 0.0699 $\pm$ 0.0473 & 0.0043 $\pm$ 0.0017 & 0.0115 $\pm$ 0.0072 \\
& $q_d$, $k_d$, cls, dense$_e$, $k_e$, $v_e$                   & 0.6972 $\pm$ 0.0198 & 0.0040 $\pm$ 0.0012 & 0.0825 $\pm$ 0.0231 & 0.0036 $\pm$ 0.0015 & 0.0076 $\pm$ 0.0037 \\
& $q_d$, $k_d$, cls, dense$_e$, $q_e$, $k_e$, $v_e$            & 0.6767 $\pm$ 0.0313 & 0.0032 $\pm$ 0.0018 & 0.0990 $\pm$ 0.0296 & 0.0040 $\pm$ 0.0019 & 0.0098 $\pm$ 0.0049 \\

\hline
\end{tabular}}
\label{table16}
\end{table*}




\begin{table*}[t]
\centering
\caption{\textbf{Effect of ST-LoRA hyperparameters and architectural choices on Mask2Former segmentation performance on BUP20} $q_d$, $k_d$, and cls denote the transformer decoder's query projection, key projection, and class predictor, respectively. $q_e$, $k_e$, $v_e$, and dense$_e$ refer to the Swin-B encoder's query, key, value projections and dense (MLP) layers. The rank sweep uses $q_d$, $k_d$, cls as target modules. Best result per category is in \textbf{bold}. We ablate the rank here with the encoder completely frozen.}
\resizebox{0.85\textwidth}{!}{

\begin{tabular}{llccccc}

\textbf{Category} & \textbf{Setting} & \textbf{mIoU} $\uparrow$ & \textbf{ECE} $\downarrow$ & \textbf{MECE }$\downarrow$ & \textbf{ACE} $\downarrow$ & \textbf{MACE} $\downarrow$ \\
\hline
\multirow{6}{*}{ST-LoRA rank -- $r$}
& 2   & 0.2270 $\pm$ 0.0140 & 0.0720 $\pm$ 0.0253 & 0.3203 $\pm$ 0.0357 & 0.1619 $\pm$ 0.0347 & 0.2489 $\pm$ 0.0230 \\
& 4   & 0.2528 $\pm$ 0.0070 & 0.0480 $\pm$ 0.0043 & 0.2466 $\pm$ 0.0247 & 0.0933 $\pm$ 0.0199 & 0.1819 $\pm$ 0.0162 \\
& 8   & 0.2831 $\pm$ 0.0068 & 0.0487 $\pm$ 0.0014 & 0.2032 $\pm$ 0.0155 & 0.0533 $\pm$ 0.0160 & 0.1531 $\pm$ 0.0169 \\
& 16  & 0.3002 $\pm$ 0.0107 & 0.0467 $\pm$ 0.0038 & 0.1782 $\pm$ 0.0076 & 0.0386 $\pm$ 0.0096 & 0.1352 $\pm$ 0.0095 \\
& 32  & 0.3192 $\pm$ 0.0103 & \textbf{0.0445 $\pm$ 0.0035} & \textbf{0.1676 $\pm$ 0.0129} & 0.0363 $\pm$ 0.0066 & \textbf{0.1270 $\pm$ 0.0107} \\
& 64  & \textbf{0.3203 $\pm$ 0.0181} & 0.0478 $\pm$ 0.0040 & 0.1824 $\pm$ 0.0161 & \textbf{0.0316 $\pm$ 0.0093} & 0.1275 $\pm$ 0.0157 \\

\hline
\multirow{15}{*}{Components}
& $q_d$, $k_d$, cls, $q_e$                                      & 0.5818 $\pm$ 0.0073 & 0.0148 $\pm$ 0.0050 & 0.2164 $\pm$ 0.0900 & 0.0109 $\pm$ 0.0026 & 0.0416 $\pm$ 0.0126 \\
& $q_d$, $k_d$, cls, $k_e$                                      & 0.5594 $\pm$ 0.0256 & 0.0141 $\pm$ 0.0014 & 0.2528 $\pm$ 0.3065 & 0.0114 $\pm$ 0.0055 & 0.0407 $\pm$ 0.0076 \\
& $q_d$, $k_d$, cls, $v_e$                                      & 0.6310 $\pm$ 0.0235 & 0.0127 $\pm$ 0.0021 & 0.1133 $\pm$ 0.0115 & 0.0094 $\pm$ 0.0016 & 0.0383 $\pm$ 0.0033 \\
& $q_d$, $k_d$, cls, dense$_e$                                  & 0.6839 $\pm$ 0.0174 & 0.0032 $\pm$ 0.0012 & \textbf{0.0619 $\pm$ 0.0258} & \textbf{0.0032 $\pm$ 0.0008} & \textbf{0.0057 $\pm$ 0.0010} \\
& $q_d$, $k_d$, cls, $q_e$, $k_e$                               & 0.5982 $\pm$ 0.0086 & 0.0095 $\pm$ 0.0037 & 0.1035 $\pm$ 0.0154 & 0.0094 $\pm$ 0.0059 & 0.0271 $\pm$ 0.0133 \\
& $q_d$, $k_d$, cls, $q_e$, $v_e$                               & 0.6692 $\pm$ 0.0153 & 0.0102 $\pm$ 0.0035 & 0.1146 $\pm$ 0.0488 & 0.0088 $\pm$ 0.0048 & 0.0295 $\pm$ 0.0091 \\
& $q_d$, $k_d$, cls, $q_e$, dense$_e$                           & 0.6833 $\pm$ 0.0199 & 0.0035 $\pm$ 0.0014 & 0.0851 $\pm$ 0.0573 & 0.0043 $\pm$ 0.0018 & 0.0092 $\pm$ 0.0038 \\
& $q_d$, $k_d$, cls, $k_e$, $v_e$                               & 0.6647 $\pm$ 0.0142 & 0.0088 $\pm$ 0.0018 & 0.0921 $\pm$ 0.0135 & 0.0096 $\pm$ 0.0056 & 0.0258 $\pm$ 0.0059 \\
& $q_d$, $k_d$, cls, $k_e$, dense$_e$                           & 0.6873 $\pm$ 0.0114 & 0.0033 $\pm$ 0.0012 & 0.0920 $\pm$ 0.0729 & 0.0048 $\pm$ 0.0018 & 0.0091 $\pm$ 0.0041 \\
& $q_d$, $k_d$, cls, $v_e$, dense$_e$                           & 0.6833 $\pm$ 0.0248 & 0.0042 $\pm$ 0.0011 & 0.0774 $\pm$ 0.0610 & 0.0047 $\pm$ 0.0020 & 0.0106 $\pm$ 0.0036 \\
& $q_d$, $k_d$, cls, $q_e$, $k_e$, $v_e$                        & 0.6704 $\pm$ 0.0237 & 0.0073 $\pm$ 0.0022 & 0.1206 $\pm$ 0.0461 & 0.0072 $\pm$ 0.0022 & 0.0213 $\pm$ 0.0077 \\
& $q_d$, $k_d$, cls, $q_e$, $k_e$, dense$_e$                    & \textbf{0.6927 $\pm$ 0.0308} & 0.0036 $\pm$ 0.0006 & 0.0943 $\pm$ 0.0284 & 0.0043 $\pm$ 0.0009 & 0.0085 $\pm$ 0.0030 \\
& $q_d$, $k_d$, cls, $q_e$, $v_e$, dense$_e$                    & 0.6844 $\pm$ 0.0209 & 0.0041 $\pm$ 0.0013 & 0.0871 $\pm$ 0.0478 & 0.0036 $\pm$ 0.0024 & 0.0075 $\pm$ 0.0043 \\
& $q_d$, $k_d$, cls, $k_e$, $v_e$, dense$_e$                    & 0.6867 $\pm$ 0.0214 & 0.0043 $\pm$ 0.0015 & 0.1151 $\pm$ 0.0459 & 0.0037 $\pm$ 0.0010 & 0.0093 $\pm$ 0.0019 \\
& $q_d$, $k_d$, cls, $q_e$, $k_e$, $v_e$, dense$_e$             & 0.6859 $\pm$ 0.0175 & \textbf{0.0030 $\pm$ 0.0010} & 0.0741 $\pm$ 0.0430 & 0.0042 $\pm$ 0.0010 & 0.0076 $\pm$ 0.0021 \\

\hline
\end{tabular}}
\label{table17}
\end{table*}

\begin{table*}[t]
\centering
\caption{\textbf{Relative change (\%) when freezing the decoder versus 
fine-tuning performed on BUP20 by using Mask2Former architecture.}, computed as $\delta = (v_\text{frz} - 
v_\text{ft})\,/\,v_\text{ft} \times 100$. For mIoU: 
\textcolor{teal}{$\delta > 0$} indicates the frozen decoder yields 
higher accuracy (improvement); \textcolor{Plum}{$\delta < 0$} indicates 
degradation. For calibration metrics (ECE, MECE, ACE, MACE; lower is 
better): \textcolor{Plum}{$\delta > 0$} indicates higher error 
(degradation); \textcolor{teal}{$\delta < 0$} indicates improvement.}
\label{tab:ablation_delta}
\resizebox{0.85\textwidth}{!}{
\begin{tabular}{llccccc}

\textbf{Category} & \textbf{Setting} 
    & $\Delta$\,\textbf{mIoU}\,(\%)
    & $\Delta$\,\textbf{ECE}\,(\%)
    & $\Delta$\,\textbf{MECE}\,(\%)
    & $\Delta$\,\textbf{ACE}\,(\%)
    & $\Delta$\,\textbf{MACE}\,(\%)  \\
\hline
\multirow{6}{*}{Rank -- $r$}
    & 2
        & \textcolor{Plum}{$-1.7$}
        & \textcolor{Plum}{$+198.6$}
        & \textcolor{Plum}{$+97.3$}
        & \textcolor{Plum}{$+107.6$}
        & \textcolor{Plum}{$+86.4$} \\
    & 4
        & \textcolor{Plum}{$-0.6$}
        & \textcolor{Plum}{$+21.3$}
        & \textcolor{Plum}{$+44.8$}
        & \textcolor{Plum}{$+5.6$}
        & \textcolor{teal}{$-0.6$} \\
    & 8 (baseline)
        & \textcolor{teal}{$+1.0$}
        & \textcolor{Plum}{$+42.9$}
        & \textcolor{Plum}{$+12.7$}
        & \textcolor{Plum}{$+82.9$}
        & \textcolor{Plum}{$+42.4$} \\
    & 16
        & \textcolor{teal}{$+0.6$}
        & \textcolor{teal}{$-27.8$}
        & \textcolor{teal}{$-16.0$}
        & \textcolor{Plum}{$+62.5$}
        & \textcolor{teal}{$-35.2$} \\
    & 32
        & \textcolor{teal}{$+0.2$}
        & \textcolor{teal}{$-23.3$}
        & \textcolor{teal}{$-11.4$}
        & \textcolor{Plum}{$+77.8$}
        & \textcolor{teal}{$-25.5$} \\
    & 64
        & \textcolor{Plum}{$-1.9$}
        & \textcolor{Plum}{$+46.7$}
        & \textcolor{teal}{$-14.5$}
        & \textcolor{Plum}{$+100.0$}
        & \textcolor{Plum}{$+81.8$} \\
\hline
\multirow{5}{*}{Scaling -- $\alpha$}
    & 2
        & \textcolor{Plum}{$-3.3$}
        & \textcolor{Plum}{$+16.1$}
        & \textcolor{Plum}{$+75.2$}
        & \textcolor{Plum}{$+65.9$}
        & \textcolor{Plum}{$+48.7$} \\
    & 4
        & \textcolor{Plum}{$-2.5$}
        & \textcolor{Plum}{$+27.4$}
        & \textcolor{Plum}{$+78.3$}
        & \textcolor{Plum}{$+60.6$}
        & \textcolor{Plum}{$+51.0$} \\
    & 8 (baseline)
        & \textcolor{teal}{$+1.0$}
        & \textcolor{Plum}{$+42.9$}
        & \textcolor{Plum}{$+12.7$}
        & \textcolor{Plum}{$+82.9$}
        & \textcolor{Plum}{$+42.4$} \\
    & 16
        & \textcolor{Plum}{$-1.6$}
        & \textcolor{teal}{$-7.7$}
        & \textcolor{teal}{$-10.1$}
        & \textcolor{Plum}{$+30.8$}
        & \textcolor{Plum}{$+2.7$} \\
    & 32
        & \textcolor{teal}{$+0.2$}
        & \textcolor{teal}{$-6.7$}
        & \textcolor{Plum}{$+22.7$}
        & \textcolor{Plum}{$+118.2$}
        & \textcolor{Plum}{$+43.7$} \\
\hline
\multirow{5}{*}{Dropout -- $p$}
    & 0.10 (baseline)
        & \textcolor{teal}{$+1.0$}
        & \textcolor{Plum}{$+42.9$}
        & \textcolor{Plum}{$+12.7$}
        & \textcolor{Plum}{$+82.9$}
        & \textcolor{Plum}{$+42.4$} \\
    & 0.15
        & \textcolor{teal}{$+0.1$}
        & \textcolor{Plum}{$+35.1$}
        & \textcolor{teal}{$-52.2$}
        & \textcolor{Plum}{$+112.5$}
        & \textcolor{Plum}{$+115.9$} \\
    & 0.20
        & \textcolor{Plum}{$-1.2$}
        & \textcolor{Plum}{$+13.6$}
        & \textcolor{Plum}{$+9.8$}
        & \textcolor{Plum}{$+100.0$}
        & \textcolor{Plum}{$+60.2$} \\
    & 0.25
        & \textcolor{Plum}{$-0.7$}
        & \textcolor{Plum}{$+42.9$}
        & \textcolor{Plum}{$+53.3$}
        & \textcolor{Plum}{$+84.9$}
        & \textcolor{Plum}{$+63.8$} \\
    & 0.30
        & \textcolor{teal}{$+0.3$}
        & \textcolor{Plum}{$+37.1$}
        & \textcolor{teal}{$-9.3$}
        & \textcolor{Plum}{$+107.3$}
        & \textcolor{Plum}{$+83.9$} \\
\hline
\multirow{8}{*}{Components}
    & \texttt{dns}, \texttt{cls} (baseline)
        & \textcolor{teal}{$+1.0$}
        & \textcolor{Plum}{$+42.9$}
        & \textcolor{Plum}{$+12.7$}
        & \textcolor{Plum}{$+82.9$}
        & \textcolor{Plum}{$+42.4$} \\
    & \texttt{dns}, \texttt{cls}, $q_e$
        & \textcolor{Plum}{$-1.8$}
        & \textcolor{Plum}{$+53.3$}
        & \textcolor{Plum}{$+46.8$}
        & \textcolor{Plum}{$+123.8$}
        & \textcolor{Plum}{$+76.3$} \\
    & \texttt{dns}, \texttt{cls}, $k_e$
        & \textcolor{Plum}{$-0.6$}
        & \textcolor{Plum}{$+42.1$}
        & \textcolor{Plum}{$+32.9$}
        & \textcolor{Plum}{$+114.3$}
        & \textcolor{Plum}{$+83.7$} \\
    & \texttt{dns}, \texttt{cls}, $v_e$
        & \textcolor{teal}{$+1.6$}
        & \textcolor{Plum}{$+20.0$}
        & \textcolor{Plum}{$+38.3$}
        & \textcolor{Plum}{$+95.0$}
        & \textcolor{Plum}{$+60.0$} \\
    & \texttt{dns}, \texttt{cls}, $q_e$, $k_e$
        & \textcolor{teal}{$+2.5$}
        & \textcolor{Plum}{$+63.0$}
        & \textcolor{Plum}{$+25.7$}
        & \textcolor{Plum}{$+165.6$}
        & \textcolor{Plum}{$+96.0$} \\
    & \texttt{dns}, \texttt{cls}, $q_e$, $v_e$
        & \textcolor{Plum}{$-1.9$}
        & \textcolor{teal}{$-17.1$}
        & \textcolor{Plum}{$+10.9$}
        & \textcolor{Plum}{$+69.8$}
        & \textcolor{teal}{$-12.2$} \\
    & \texttt{dns}, \texttt{cls}, $k_e$, $v_e$
        & \textcolor{teal}{$+0.1$}
        & \textcolor{Plum}{$+60.0$}
        & \textcolor{Plum}{$+8.0$}
        & \textcolor{Plum}{$+130.6$}
        & \textcolor{Plum}{$+117.1$} \\
    & \texttt{dns}, \texttt{cls}, $q_e$, $k_e$, $v_e$
        & \textcolor{teal}{$+1.0$}
        & \textcolor{Plum}{$+18.8$}
        & \textcolor{teal}{$-15.7$}
        & \textcolor{Plum}{$+87.5$}
        & \textcolor{Plum}{$+38.8$} \\
\hline
\end{tabular}}
\label{table18}
\end{table*}

\begin{table*}[t]
\caption{\textbf{Default training settings for section~\ref{sec5}.} 
We use SegFormer B2/4 on GrowliFlower-L and Mask2Former (M2F)-Swin/B on BUP20.}
\centering
\resizebox{0.85\textwidth}{!}{%
\begin{tabular}{l l | c c | c}

 & & \multicolumn{2}{c|}{\textbf{ Pixel wise binary classification}} & \textbf{Pixel wise multiclass classification} \\
\cmidrule(lr){3-4}
\textbf{Category} & \textbf{Hyperparameter} & \textbf{Baseline} & \textbf{FRE / ST-LoRA} & \textbf{Baseline / FRE, and ST-LoRA} \\
\midrule
\multirow{5}{*}{\textit{Dataset}}
 & Name               & \multicolumn{2}{c|}{GrowliFlower-L}    & BUP20 \\
 & Train  & \multicolumn{2}{c|}{1381}           & 124 \\
 & Validation & \multicolumn{2}{c|}{293}        & 63 \\
 & Test   & \multicolumn{2}{c|}{296}            & 92 \\
\midrule
\multirow{3}{*}{\textit{Pre-processing}}
 & Image size         & \multicolumn{2}{c|}{$256 \times 256$}  & $1280 \times 720$ \\
 & Normalization      & \multicolumn{2}{c|}{$[0, 1]$}          & $\mu{=}[0.485, 0.456, 0.406]$, $\sigma{=}[0.229, 0.224, 0.225]$ \\
 & Augmentation       & \multicolumn{2}{c|}{None}              & None \\
\midrule
\multirow{3}{*}{\textit{Model}}
 & Architecture       & \multicolumn{2}{c|}{SegFormer (B2 / B4)} & Mask2Former (Swin-B)/EoMT-Large(DINOv2) \\
 & Pre-trained weights & \multicolumn{2}{c|}{ImageNet}           & ADE20K \\
 & LoRA rank          & ---           & $r{=}$8         & --- / $r{=}$8 \\
\midrule
\multirow{8}{*}{\textit{Training}}
 & Epochs             & 100           & 100                   & 100 \\
 & Optimizer          & Adam          & Adam                  & Adam \\
 & Learning rate      & $1{\times}10^{-4}$ & $[2{\times}10^{-4},\, 2{\times}10^{-2}]$ & $2{\times}10^{-4}$/ $[2{\times}10^{-5},\, 2{\times}10^{-3}]$ \\
 & LR schedule        & Constant      & Cosine annealing w/ warm restarts & Constant / Cosine annealing \\
 & Cycle length       & ---           & 20 epochs             & 20 epochs \\
 & Ensemble members   & ---           & 5                     & --- / 5 \\
 & Batch size         & 32            & 32                    & 4 \\
 & Early stopping     & Patience~$=100$ & ---                 & --- \\
 & Loss               & \multicolumn{2}{c|}{Cross-entropy}     & Cross-entropy \\
 & Weight decay       & \multicolumn{2}{c|}{0.0}               & 0.0 \\
 & Hardware           & \multicolumn{2}{c|}{NVIDIA A100 (80\,GB)}             & NVIDIA A100 (80\,GB) \\
\bottomrule
\end{tabular}%
 }
\label{tblsec5.1}
\end{table*}

\begin{table*}[t]
\centering
\caption{\textbf{Baseline configurations for the ablation study in section~\ref{sec6}}. Each experiment 
varies a single hyperparameter while keeping all others fixed.
Subscripts $e$ and $d$ denote encoder and decoder projections 
respectively. CLS: class predictor; MLP: dense layer.}
\label{tab:ablation_config}
\setlength{\tabcolsep}{4pt}
\small
\begin{tabularx}{0.85\textwidth}{lXX}

\textbf{Setting} & \textbf{SegFormer-B2} & \textbf{Mask2Former-Swin/B} \\
\midrule
\multicolumn{3}{l}{\textbf{Parameter Groups}} \\
ST-LoRA-adapted     
    & $q_e$, $k_e$, $v_e$, MLP (encoder)
    & $q_d$, $k_d$, CLS, MLP (encoder) \\
Fully fine-tuned 
    & Decoder + segmentation head 
    & — \\
Frozen           
    & Remaining encoder 
    & Remaining encoder parameters and pixel decoder \\
\midrule
\multicolumn{3}{l}{\textbf{ST-LoRA Configuration}} \\
Rank $r$                & 32     & 8 \\
Scaling $\alpha$        & 64     & 8 \\
$A$ init                & Random & Random \\
$B$ init                & Zero   & Zero \\
Bias                    & No     & No \\
\midrule
\multicolumn{3}{l}{\textbf{Training}} \\
Optimizer        & Adam   & Adam \\
Learning rate    & $2\times10^{-2}$--$2\times10^{-4}$ & $2\times10^{-3}$--$2\times10^{-5}$ \\
Batch size       & 32     & 4 \\
Epochs           & 100    & 110 \\
LR schedule      & \multicolumn{2}{X}{CosineAnnealingWarmRestarts 
                  ($T_0=20$, $T_\text{mult}=1$)} \\
Ensemble size    & \multicolumn{2}{X}{5 checkpoints} \\
\midrule
\multicolumn{3}{l}{\textbf{Evaluation}} \\
Seeds            & \multicolumn{2}{X}{5} \\
Reported stats   & \multicolumn{2}{X}{Mean $\pm$ std} \\
\bottomrule
\end{tabularx}
\end{table*}

\begin{table*}[t]
\centering
\caption{\textbf{Experimental setup for robustness evaluation under distribution 
shift in section~\ref{sec7}}. Two complementary analyses are conducted: ensemble diversity 
(SegFormer-B2 on GrowliFlower-L) and component sensitivity 
(Mask2Former-Swin/B on BUP20). Photometric corruptions alter pixel 
statistics only; geometric corruptions are applied to both image and 
mask. For GrowliFlower-L, severity increases continuously across 20 
levels; for BUP20, three discrete severity levels are used per 
corruption type.}
\label{tab:dist_shift_setup}
\setlength{\tabcolsep}{4pt}
\small
\begin{tabularx}{0.85\textwidth}{l l X X}

& & \textbf{Exp.\,1: Ensemble Diversity} 
  & \textbf{Exp.\,2: Component Sensitivity} \\
\cmidrule(lr){3-3}\cmidrule(lr){4-4}
& & \textbf{SegFormer-B2 / GrowliFlower-L} 
  & \textbf{Mask2Former-Swin/B / BUP20} \\
\midrule
\multirow{2}{*}{\textbf{Variants}}
  & \textbf{Details}\ & \textbf{Homogeneous ensemble (LR schedule only)} 
            & \multirow{2}{*}{\textbf{ST-LoRA hyperparameters robustness study}} \\
  &         & \textbf{Heterogeneous ensemble (LR + augmentation)} & \\
\midrule
\multirow{4}{*}{\rotatebox[origin=c]{90}{\textbf{Photometric}}}
  & Brightness   
    & Factor $\in [1.0, 2.0]$, 20 levels ($\Delta=0.05$)
    & Factor $\in \{0.50, 0.65, 0.80\}$ \\
  & Contrast     
    & Factor $\in [1.0, 2.0]$, 20 levels ($\Delta=0.05$)
    & Factor $\in \{0.50, 0.65, 0.80\}$ \\
  & Gaussian noise 
    & $\sigma \in [0, 0.285]$, 20 levels ($\Delta=0.015$)
    & $\sigma \in \{0.05, 0.10, 0.20\}$ \\
  & Gaussian blur  
    & ---
    & $\sigma \in \{1, 2, 3\}$\,px \\
\midrule
\multirow{4}{*}{\rotatebox[origin=c]{90}{\textbf{Geometric}}}
  & Rotation    
    & $\theta \in [0^{\circ}, 85.5^{\circ}]$, 20 levels ($\Delta=4.5^{\circ}$)
    & --- \\
  & Zoom        
    & Scale $\in [1.0\times, 1.475\times]$, 20 levels ($\Delta=0.025$)
    & Scale $\in \{1.10\times, 1.25\times, 1.50\times\}$ \\
  & Horizontal flip 
    & ---
    & Binary (single level) \\
  & Translation 
    & ---
    & $\{5\%, 10\%, 15\%\}$ of image width \\
\midrule
\multicolumn{2}{l}{\textbf{Total shift conditions}} 
  & $4 \times 20 = 80$ 
  & $7 \times 3 - 2 = 19$ \\
\multicolumn{2}{l}{\textbf{Mask handling}}          
  & Geometric: nearest-neighbour interp.\ on mask 
  & Geometric: nearest-neighbour interp.\ on mask \\
\multicolumn{2}{l}{\textbf{Metrics}}                
  & \multicolumn{2}{X}{mIoU, and ECE (mean $\pm$ std over 5 seeds)} \\
\bottomrule
\end{tabularx}
\end{table*}

\begin{table*}[t]
\centering
\caption{\textbf{Configuration summary for all uncertainty quantification 
methods evaluated on Mask2Former-Swin/B (BUP20) in section~\ref{sec8}}. SN: Spectral Normalization \cite{miyato2018spectral}. 
GMM: Gaussian Mixture Model fitted via EM~\cite{Dempster1977maximum}. 
ST-LoRA target modules follow the notation of 
Table~\ref{tab:ablation_config}}.
\label{tab:method_configs}
\setlength{\tabcolsep}{6pt}
\begin{tabularx}{0.85\textwidth}{l l l X}

\textbf{Method} 
    & \textbf{Ensemble} 
    & \textbf{Inference} 
    & \textbf{Settings} \\
\midrule
FRE                 
    & Yes   
    & ($M=4$)  forward passes    
    & Following the similar Full Rank Ensembles from previous experiments.  \\
\addlinespace
MC Dropout            
    & No  
    & $T=4$ stochastic passes 
    & We dropout layer before the class predictor using a dropout probability of \(p = 0.5\) following \cite{Gal2015DropoutAA,JMLR:v15:srivastava14a}. \\
\addlinespace
DDU                    
    & No  
    & Single forward pass   
    & SN has been applied to all linear and convolutional 
      layers. A Gaussian Mixture Model ($K{=}2$ 
      components per class, 8 classes total) is fitted to these 
      features using the EM algorithm for $1000$ iterations. Pixels 
      are treated as spatially independent data points, yielding 
      $1280 \times 720$ data points per image. \\
\addlinespace
ST-LoRA (ours)          
    & Yes  
    & ($M=4$)  forward passes    
    & LoRA adaptation with frozen decoder. Rank $r{=}32$, 
      scaling $\alpha{=}32$, dropout $p{=}0.1$. Target modules: 
      decoder $q_d$, $k_d$, \texttt{cls}; encoder \texttt{MLP}, 
      $k_e$, $v_e$.\\
\bottomrule
\end{tabularx}
\end{table*}

\section{Complementary Information: Experimental Setup Settings}
\label{appa_b}

\textbf{Hardware Settings  and Reproducibility}. All experiments are conducted on \textit{NVIDIA A100-PCIE-80GB} GPUs from the Marvin cluster, using up to four GPUs per node for training and a single GPU for inference. RAM allocation varies between 128\,GB and 512\,GB depending on dataset and image size. All results are reported across five random seeds (42, 123, 456, 789, 1337) for statistical significance.

\textbf{Uncertainty Methods Settings} For snapshot-based methods (ST-LoRA and FRE), we retain the last four checkpoints of each training trajectory, discarding the first 
snapshot as the model is still in an early underfitting phase at 
that point and including it would distort the ensemble evaluation. 
For DDU, we fit the GMM on features extracted from the last layer of 
the pixel decoder, using per-class pixel frequency in the training 
set as a prior.

All methods share the same augmentation policy to ensure a fair comparison. For both FRE and ST-LoRA, ensemble diversity is not 
enforced through fixed, member-specific augmentation schemes; 
instead, we apply highly randomized augmentation throughout the 
entire training trajectory, with most augmentations set to a 
probability close to 0.5. Diversity across members therefore emerges 
naturally from this stochasticity rather than from an explicitly 
assigned per-member policy.

\textbf{Post-hoc Calibration Methods Settings.} We keep the trained Mask2Former completely frozen and calibrate its outputs. For each image, the model emits two heads -\textit{class query logits} and \textit{ mask query logits}- which, by architectural design, collapse into a per-pixel semantic probability map. We then drop the no-object column, apply softmax to the class logits and sigmoid to the mask logits, and combine them via element-wise multiplication. Finally, we take the logarithm of the combined output to obtain per-pixel logits, which reside in a well-behaved logit space suitable for our six post-hoc calibrators.

Each calibrator is fit on a small, fixed held-out calibration split -specifically, a deterministic subset of 30 samples carved from the validation set, while the remaining 33 validation samples and the full 93 test samples remain untouched for tuning. Fitting is performed by minimizing pixel-wise NLL and cross-entropy against the ground-truth labels using AdamW \cite{loshchilov2018decoupled} for approximately 40 epochs, with the backbone kept frozen. Calibration is conducted at a low resolution (320x180),\footnote{We adapt the same approach from \url{https://github.com/dwang181/selectivecal}} and the learned transform is subsequently applied at native resolution (1270x720). At evaluation, we simply pass the frozen model's logits through the fitted calibrator, apply softmax, and recompute mIoU and accuracy alongside the calibration metrics (ECE, ACE, MECE, MACE, Brier, and NLL).

We acknowledge that the validation split used for calibrating the post-hoc methods differs from the validation set used for the four uncertainty baselines and the main experiments. However, we believe this does not compromise the fairness of our comparison, as we adopt the same training set and hyperparameters throughout. Crucially, we do not select model weights based on validation loss or accuracy; instead, we use the final checkpoint from the training trajectory, corresponding to the last snapshot. This choice effectively isolates the influence of the validation set and ensures a consistent basis for comparison. Had we instead used the checkpoint selected by best validation performance, we would have been forced to retrain all baseline methods under identical conditions to align with the post-hoc calibrators' underlying models, which would introduce additional complexity and potential variability.

\textbf{Architectures and LoRA software}. We use Hugging Face's (HF) tool PEFT (Parameter-Efficient Fine-Tuning) \footnote{\url{https://huggingface.co/docs/peft/en/index}} to adapt the LoRA modules to the different architectures in our study. Mask2Former \footnote{\url{https://huggingface.co/facebook/mask2former-swin-base-ade-semantic}}, and EoMT \footnote{\url{https://huggingface.co/tue-mps/ade20k_semantic_eomt_large_512}} architectures have been used from models library in HF. While for SegFormer we use the same model either from HF \footnote{\url{https://huggingface.co/docs/transformers/en/model_doc/segformer}} or \textit{segmentation models} \footnote{\url{https://github.com/qubvel-org/segmentation_models.pytorch}}.

\begin{figure*}[H]
    \centering
    \includegraphics[width=0.90\linewidth]{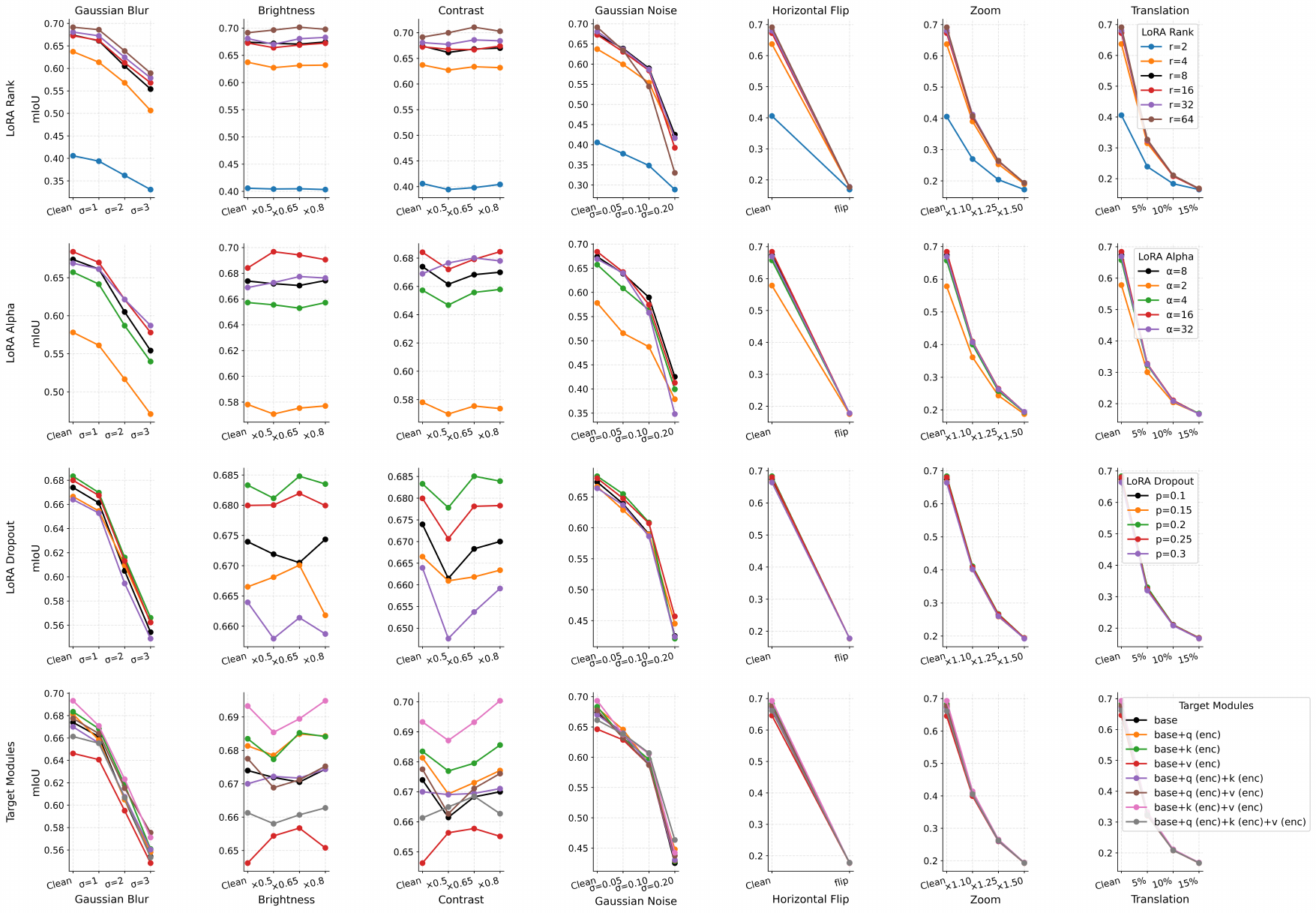}
    \caption{\textbf{Hyperparameters robustness against distribution shift measured by mIoU metric.}}
    \label{miouablat}
\end{figure*}

\begin{figure*}[H]
    \centering
    \includegraphics[width=0.90\linewidth]{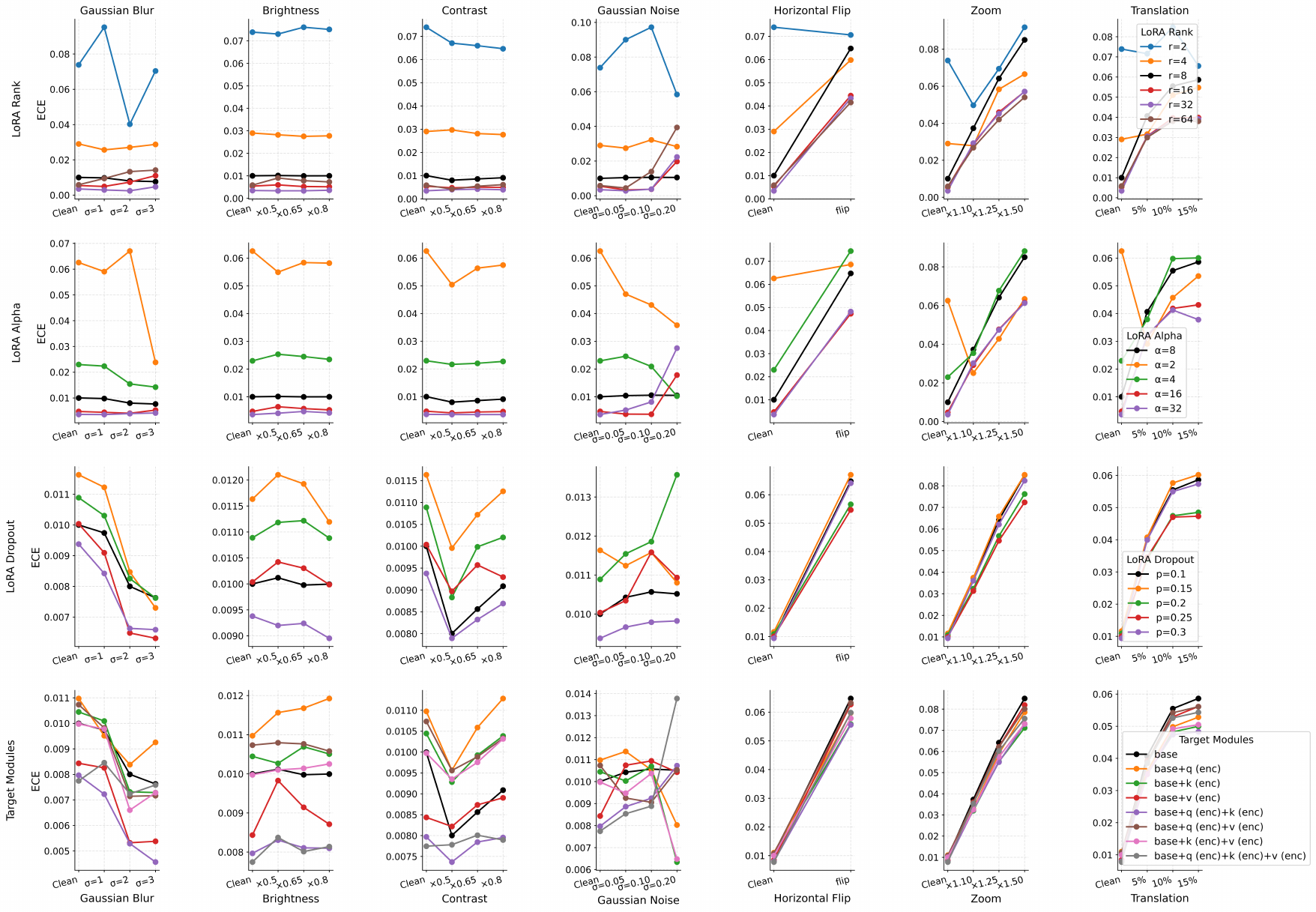}
    \caption{\textbf{Hyperparameters robustness against distribution shift measured by ECE metric.}}
    \label{eceablat}
\end{figure*}

\end{document}